\documentclass[nohyperref]{article}

\usepackage{microtype}
\usepackage{graphicx}
\usepackage{subcaption}
\usepackage{booktabs} 

\usepackage{hyperref}

\usepackage[accepted]{icml2022}

\usepackage{amsmath}
\usepackage{amssymb}
\usepackage{mathtools}
\usepackage{amsthm}
\usepackage{bm}
\usepackage{url}            
\usepackage{amsfonts}       
\usepackage{nicefrac}       
\usepackage{xcolor}         
\usepackage{enumitem}
\usepackage{bbm}
\usepackage{color}
\usepackage{multirow}
\usepackage{comment}
\usepackage{xr}  
\usepackage[figuresright]{rotating} 

\usepackage{listings}
\usepackage[capitalize,noabbrev]{cleveref}

\theoremstyle{plain}
\newtheorem{theorem}{Theorem}[section]

\theoremstyle{definition}

\theoremstyle{remark}

\newcommand{\bX}{\bm{X}}
\newcommand{\bx}{\bm{x}}

\newcommand{\btheta}{\bm{\theta}}
\newcommand{\Tau}{\mathcal{T}}

\usepackage[textsize=tiny]{todonotes}

\icmltitlerunning{Tree-based Model Averaging Approach for Heterogeneous Data Sources}

\begin{document}

\twocolumn[
\icmltitle{A Tree-based Model Averaging Approach for Personalized Treatment Effect Estimation from Heterogeneous Data Sources}



\icmlsetsymbol{equal}{*}

\begin{icmlauthorlist}
\icmlauthor{Xiaoqing Tan}{yyy}
\icmlauthor{Chung-Chou H. Chang}{yyy}
\icmlauthor{Ling Zhou}{comp}
\icmlauthor{Lu Tang}{yyy}
\end{icmlauthorlist}

\icmlaffiliation{yyy}{University of Pittsburgh, Pittsburgh, PA, USA}
\icmlaffiliation{comp}{Southwestern University of Finance and Economics, Chengdu, China}

\icmlcorrespondingauthor{Lu Tang}{lutang@pitt.edu}

\icmlkeywords{Machine Learning, ICML}

\vskip 0.3in
]



\printAffiliationsAndNotice{}  

\begin{abstract}
Accurately estimating personalized treatment effects within a study site (e.g., a hospital) has been challenging due to limited sample size. Furthermore, privacy considerations and lack of resources prevent a site from leveraging subject-level data from other sites. We propose a tree-based model averaging approach to improve the estimation accuracy of conditional average treatment effects (CATE) at a target site by leveraging models derived from other potentially heterogeneous sites, without them sharing subject-level data. To our best knowledge, there is no established model averaging approach for distributed data with a focus on improving the estimation of treatment effects. Specifically, under distributed data networks, our framework provides an interpretable tree-based ensemble of CATE estimators that joins models across study sites, while actively modeling the heterogeneity in data sources through site partitioning. The performance of this approach is demonstrated by a real-world study of the causal effects of oxygen therapy on hospital survival rate and backed up by comprehensive simulation results.
\end{abstract}

\section{Introduction} \label{sec:introduction}


Estimating individualized treatment effects has been a hot topic because of its wide applications, ranging from personalized medicine, policy research, to customized marketing advertisement. Treatment effects of certain subgroups within the population are often of interest. Recently, there has been an explosion of research devoted to improving estimation and inference of covariate-specific treatment effects, or conditional average treatment effects (CATE) at a target research site \citep{athey2016recursive,wager2018estimation,hahn2020bayesian,kunzel2019metalearners,nie2020quasioracle}. 
However, due to the limited sample size in a single study, improving the accuracy of the estimation of treatment effects remains challenging. 

Leveraging data and models from various research sites to conduct statistical analyses is becoming increasingly popular \citep{reynolds2020leveraging, cohen2020leveraging, berger2015optimizing}.
Distributed research networks have been established in many large scale studies \citep{fleurence2014launching,hripcsak2015observational,platt2018fda,donohue2021use}.  %
A question often being asked is whether additional data or models from other research sites could bring improvement to a local estimation task, especially when a single site does not have enough data to achieve a desired statistical precision. 
This concern is mostly noticeable in estimating treatment effects where sample size requirement is high yet observations are typically limited. 
Furthermore, information exchange between data sites is often highly restricted due to privacy, feasibility, or other concerns, prohibiting centralized analyses that pool data from multiple sources \citep{maro2009design,brown2010distributed,toh2011comparative,raghupathi2014big,deshazo2015comparison,donahue2018veterans,dayan2021federated}. 
One way to tackle this challenge is through model averaging \citep{raftery1997bayesian}, where multiple research sites collectively contribute to the tasks of statistical modeling without sharing sensitive subject-level data.
Although this idea has existed in supervised learning problems  \citep{dai2011greedy,mcmahan2017communication}, 
to our best knowledge, there are no established model averaging approach and theoretical results on estimating CATE in a distributed environment. The extension is non-trivial because CATE is unobserved in nature, as opposed to prediction problems where labels are given.



This paper focuses on improving the prediction accuracy of CATE concerning a target site by leveraging models derived from other sites 
where 
\textit{transportability}  \citep[to be formally defined in Section~\ref{sec:assumption},][]{pearl2011transportability,stuart2011use,pearl2014external,bareinboim2016causal,buchanan2018generalizing,dahabreh2019generalizing} may not hold. 
Specifically, there may exist heterogeneity in treatment effects. In the context of our multi-hospital example, these are: 
1) \textbf{local heterogeneity}: within a hospital, patients with different characteristics may have different treatment effects.
This is the traditional notion of CATE;
and 2) \textbf{global heterogeneity}: where the same patient may experience different treatment effects at different hospitals. The second type of heterogeneity is driven by site-level confounding, and hampers the transportability of models across hospital sites. 
We also note that these two types of heterogeneity may interact with each other in the sense that transportability is dependent on patient characteristics, which we will address.

We propose a model averaging framework that uses a flexible tree-based weighting scheme to combine learned models from sites that takes into account heterogeneity. The contribution of each learned model to the target site depends on subject characteristics. This is achieved by applying tree splittings \citep{breiman1984classification} at both the site and the subject levels. 
For example, effects of a treatment in two hospitals may be similar for female patients but not for male, suggesting us to consider borrowing information across sites only on selective subgroups. 
Our approach extends the classic model averaging framework \citep{raftery1997bayesian,wasserman2000bayesian,hansen2007least,yang2001adaptive} by allowing data-adaptive weights, which are interpretable in a sense that they can be used to lend credibility to transportability. 
For example, in the case of extreme heterogeneity where other sites merely contribute to the target, the weights can be used as a diagnostic tool to inform the decision against borrowing information.

\textbf{Main contributions.} 
{{1)}} We propose a model averaging scheme with interpretable weights that are adaptive to both local and global heterogeneity via tree-splitting dedicated to improving CATE estimation under distributed data networks. 
{{2)}} We generalize model averaging techniques to study the transportability of causal inference. Causal assumptions with practical implications are explored to warrant the use of our approach. 
{{3)}} We provide an extensive empirical evaluation of the proposed approach with a concrete real-data example on how to apply the method in practice. 
{{4)}} Compared to other distributed learning methods, the proposed framework enables causal analysis without sharing subject-level data, is easy to implement, offers ease of operations, and minimizes infrastructure, which facilitates practical collaboration within research networks. 



\section{Related Work} \label{sec:related}

There are two types of construct of a distributed database \citep{breitbart1986database}: \emph{homogeneous} versus \emph{heterogeneous}. 
For homogeneous data sources, data across sites are random samples of the global population. 
Recent modeling approaches \citep{lin2010relative, lee2017communication, mcmahan2017communication, battey2018distributed, jordan2018communication, tang2020distributed,wang2021tributarypca} all assume samples are randomly partitioned, which guarantees 
identical data distribution across sites. 
The goal of these works is to improve overall prediction by averaging results from homogeneous sample divisions. 
The classic random effects meta-analysis (see, e.g.,  \citet{whitehead2002meta,sutton2000methods,borenstein2011introduction} describes heterogeneity using modeling assumptions, but its focus mostly is still on global patterns. 


\textbf{Heterogeneous models.} In practice, however, there is often too much global heterogeneity 
in a distributed data network 
to warrant direct aggregation of models obtained from local sites. The focus shifts to improving the estimation of a target site by selectively leveraging information from other data sources.
There are two main classes of approaches.
The first class is based on comparison of the learned model parameters $\{\widehat\btheta_1,\dots,\widehat\btheta_K\}$ from $K$ different sites where for site $k$ we adopt model $f_k(\bx) = f(\bx; \btheta_k)$ with subject features $\bx$ to approximate the outcome of interest $Y$. 
Clustering and shrinkage approaches are then used
by merging data or models that are similar \citep{ke2015homogeneity,tang2016fused,smith2017federated,ma2017concave,wang2020sylvester,tang2020individualized,tang2021poststratification}. 
Most of these require the pooling of subject-level data.
The second class of approaches falls in the \textit{model averaging} framework \citep{raftery1997bayesian} with weights directly associated with the local prediction. 
Let site 1 be our target site, and the goal is to improve $f_{1}$ using a weighted estimator $f^*(\bx) = \sum_{k=1}^K {\omega}_{k} f_k(\bx)$ with weights $\omega_k$ to balance the contribution of each model and $\sum_k \omega_{k} = 1$. 
It provides an immediate interpretation of usefulness of each data source. 
When the weights are proportional to the prediction performance of $f_k$ on site 1, for example, 
$${\omega}_{k} =\frac{\exp\{- \sum_{i \in \mathcal{I}_1}(f_k(\bx_i) - y_i)^2\}}{ \sum_{\ell=1}^{K} \exp\{- \sum_{i \in \mathcal{I}_1}(f_\ell(\bx_i) - y_i)^2\} },$$
with $y_i$ being the observed outcome of subject $i$ in site 1, indexed by $\mathcal{I}_1$, the method is termed as the exponential weighted model averaging (EWMA).
Several variations of ${\omega}_{k}$ can be found in
\citet{yang2001adaptive, dai2011greedy,yao2018using,dai2018bayesian}. 
In general, separate samples are used to obtain the estimates of $\omega_k$'s and $f_k$'s, respectively. 

Here we focus on the literature review of model averaging. 
We note that our framework is also related to federated learning \citep{mcmahan2017communication}. But the latter often
involves iterative updating  rather than a one-shot procedure, and could be hard to apply to nonautomated distributed
research networks. Besides, it has been developed mainly to estimate a global prediction model by leveraging distributed data, and is not designed to target any specific site. We further discuss these approaches and other related research topics and their distinctions with model averaging in Appendix~\ref{suppl-related}.


\textbf{Transportability.}
In causal inference, there is a lot of interest in identifying subgroups with enhanced treatment effects, targeting  at  the  feasibility  of  customizing  estimates  for  individuals \citep{athey2016recursive,wager2018estimation,hahn2020bayesian,kunzel2019metalearners,nie2020quasioracle}. These methods aim to estimate the CATE function $\tau(\bx)$, denoting the difference in potential outcomes between treatment and control, conditional on subject characteristics $\bx$. 
To reduce uncertainty in estimation of personalized treatment effects, incorporating additional data or models are sought after. 
\citet{pearl2011transportability,pearl2014external,bareinboim2016causal} introduced the notion of transportability to warrant causal inference models be generalized to a new population. 
The issue of generalizability is common in practice due to the non-representative sampling of participants in randomized controlled trials \citep{cook2002experimental,druckman2011cambridge,allcott2015site,stuart2015assessing,egami2020elements}. 
Progress on bridging the findings from an experimental study with observational data can be found in, e.g., \citet{stuart2015assessing,kern2016assessing,stuart2018generalizability,ackerman2019implementing,yang2020elastic,harton2021combining}. 
See \citet{tipton2018review,colnet2020causal,degtiar2021review} and references therein for a comprehensive review. 
However, most methods require fully centralized data. In contrast, we leverage the distributed nature of model averaging to derive an integrative CATE estimator.

\section{A Tree-based Model Averaging Framework} \label{sec:method}


We first formally define the {conditional average treatment effect} (CATE). 
Let $Y$ denote the outcome of interest, $Z \in \{0,1\}$ denote a binary treatment indicator, and $\bX$ denote subject features. Correspondingly, let $y$, $z$ and $\bx$ denote their realizations. 
Using the potential outcome framework \citep{neyman1923applications,rubin1974estimating}, we define
CATE as $\tau(\bx)=E[Y^{(Z=1)}-Y^{(Z=0)} |\bX=\bx],$
where $Y^{(Z=1)}$ and $Y^{(Z=0)}$ are the potential outcomes under treatment arms $Z=1$ and $Z=0$, respectively. 
The expected difference of the potential outcomes is dependent on subject features $\bX$. By the causal consistency assumption, the observed outcome is $Y = ZY^{(Z=1)} + (1 - Z)Y^{(Z=0)}$.

Now suppose the distributed data network 
consists of $K$ sites, each with sample size of $n_k$. 
Site $k$ contains data $\mathcal{D}_k = \{y_i, z_i, \bx_i\}_{i \in \mathcal{I}_k}$, where $\mathcal{I}_k$ denotes its index set.
Its CATE function is given by $\tau_k (\bx) = E_k[Y^{(Z=1)}-Y^{(Z=0)} |\bX=\bx],$ where the expectation is taken over the data distribution in site $k$. 
Without loss of generality, we assume the goal is to estimate the CATE function in site 1, $\tau_1$.


\subsection{Causal Assumptions} \label{sec:assumption}

To ensure information can be properly borrowed across sites, we first impose the following idealistic assumptions, and then present relaxed version of Assumption 2. 
Let $S$ be the site indicator taking values in $\mathcal{S} = \{1, \dots, K\}$ such that $S_i = k$ if $i \in \mathcal{I}_k$. 

\textit{Assumption 1:} $\{Y^{(Z=0)}, Y^{(Z=1)}\} \perp Z | \bX, S$;

\textit{Assumption 2 (Transportability):} $$
\{Y^{(Z=0)}, Y^{(Z=1)}\} \perp S | \bX;$$
\textit{Assumption 3:} $	0< P(S = 1| \bX) < 1 \mbox{ and } 0< P(Z = 1| \bX, S) < 1   \quad \mbox{for all } \bX \mbox{ and } S$.

Assumption 1 
ensures treatment effects are unconfounded within sites so that $\tau_k(\bx)$ can be consistently identified. 
It holds by design when data are randomized controlled trials or when treatment assignment depends on $\bX$. 
By this assumption, we have $\tau_{k}(\bx) =
E[Y |\bX=\bx, S = k, Z = 1] - E[Y |\bX=\bx, S = k, Z = 0]$.
The equality directly results from the assumption. 
Assumption 2 
essentially states that the CATE functions are transportable, i.e., $\tau_k(\bx) = \tau_{k'}(\bx)$ for $k, k'\in \{1,\dots, K\}$. See also \citet{stuart2011use}, \citet{buchanan2018generalizing} and \citet{yang2020elastic} for similar consideration.
This assumption may not be satisfied due to heterogeneity across sites. In other words, site can be a confounder which prevents transporting of CATE functions across sites.
Our method allows Assumption 2 
to be violated and use model averaging weights to determine transportability. Explicitly, we consider a relaxed Assumption 2a to hold for a subset of sites that contains site 1.

\textit{Assumption 2a (Partial Transportability):} $$
\{Y^{(Z=0)}, Y^{(Z=1)}\} \perp S_1 | \bX.$$ 
Here, $S_1$ takes values in $\mathcal{S}_1 = \{k: \tau_k(\bx) = \tau_1(\bx)\}$ and $\{1\} \subset \mathcal{S}_1 \subset \mathcal{S}$. 
We denote $\mathcal{S}_1$ as the set of transportable sites with regard to site 1. 
Hence, transportability holds across some sites and specific subjects.
In a special case in Section~\ref{sec:simulation} where $\mathcal{S}_1 = \{1\}$, bias may be introduced to by model averaging. However, our approach is still able to exploits the bias and variance trade off to improve estimation. 
Assumption 3 
ensures that all subjects are possible to be observed in site 1 and all subjects in all sites are possible to receive either arm of treatment. 
The former ensures a balance of covariates between site 1 population and the population of other sites. 
Violation of either one may result in extrapolation and introduce unwanted bias to the ensemble estimates for site 1. This assumption is also used, e.g., in \citet{stuart2011use}.


\subsection{Model Ensemble} \label{sec:adaptiveMA}

We consider an adaptive weighting of $\{\tau_1, \dots, \tau_K\}$ by
\begin{equation} \label{eq:agg}
   {\tau}^*(\bx) = \sum_{k=1}^K \omega_{k}(\bx) \tau_k(\bx)
\end{equation}
where ${\tau}^*$ is the weighted model averaging estimator. The weight functions $\omega_{k}(\bx)$'s are not only site-specific, but also depend on $\bx$, and follow $\sum_{k=1}^{K} \omega_{k}(\bx) = 1$.
It measures the importance of $\tau_k$ in assisting site 1 when subjects with characteristics $\bx$ are of interest. 
We rely on each of the sites to derive their respective $\widehat\tau_k$ from $\mathcal{D}_k$ so that $\mathcal{D}_1, \dots, \mathcal{D}_K$ do not need to be pooled. Only the estimated functions $\{\widehat \tau_2, \dots, \widehat \tau_K\}$ are passed to site 1. 
We will describe the approaches to estimate $\widehat{\tau}_k$ in Section \ref{sec:local}.

\begin{figure}[tb]
\centering
 \begin{subfigure}{0.5\textwidth}
  \centerline{\includegraphics[width=\linewidth]{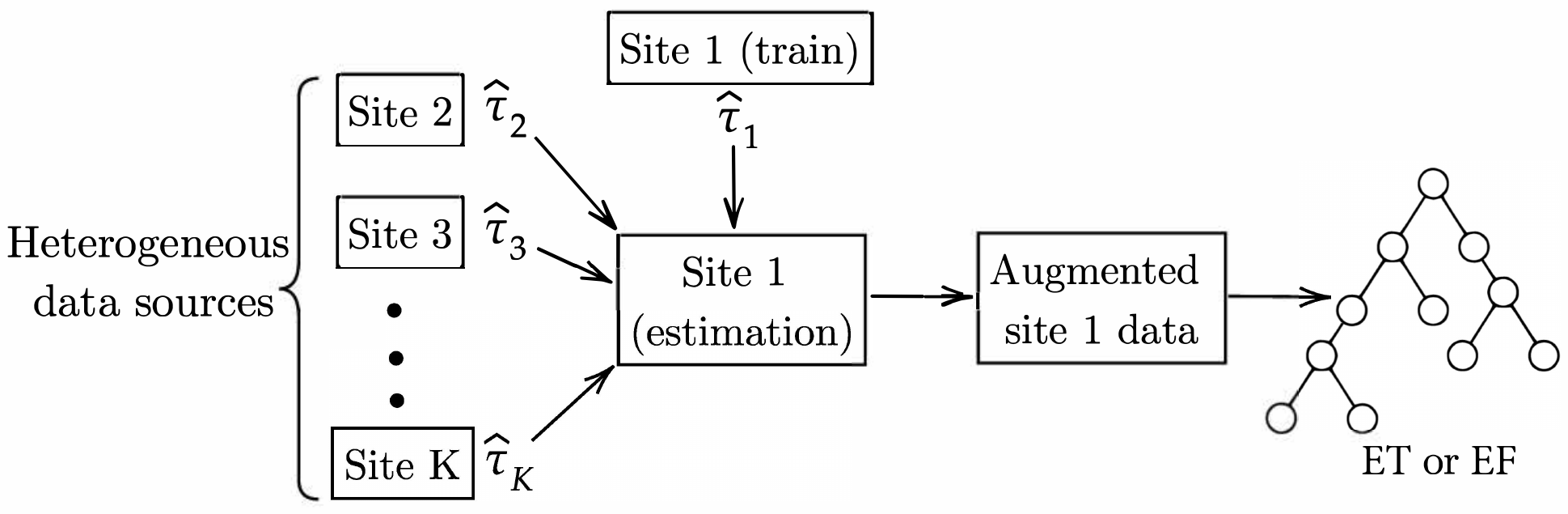}}
  \vspace{-0.3cm}
  \caption{}
  \vspace{0.2pc}
 \end{subfigure}
 \begin{subfigure}{.25\textwidth}
  \centerline{\includegraphics[width=\linewidth]{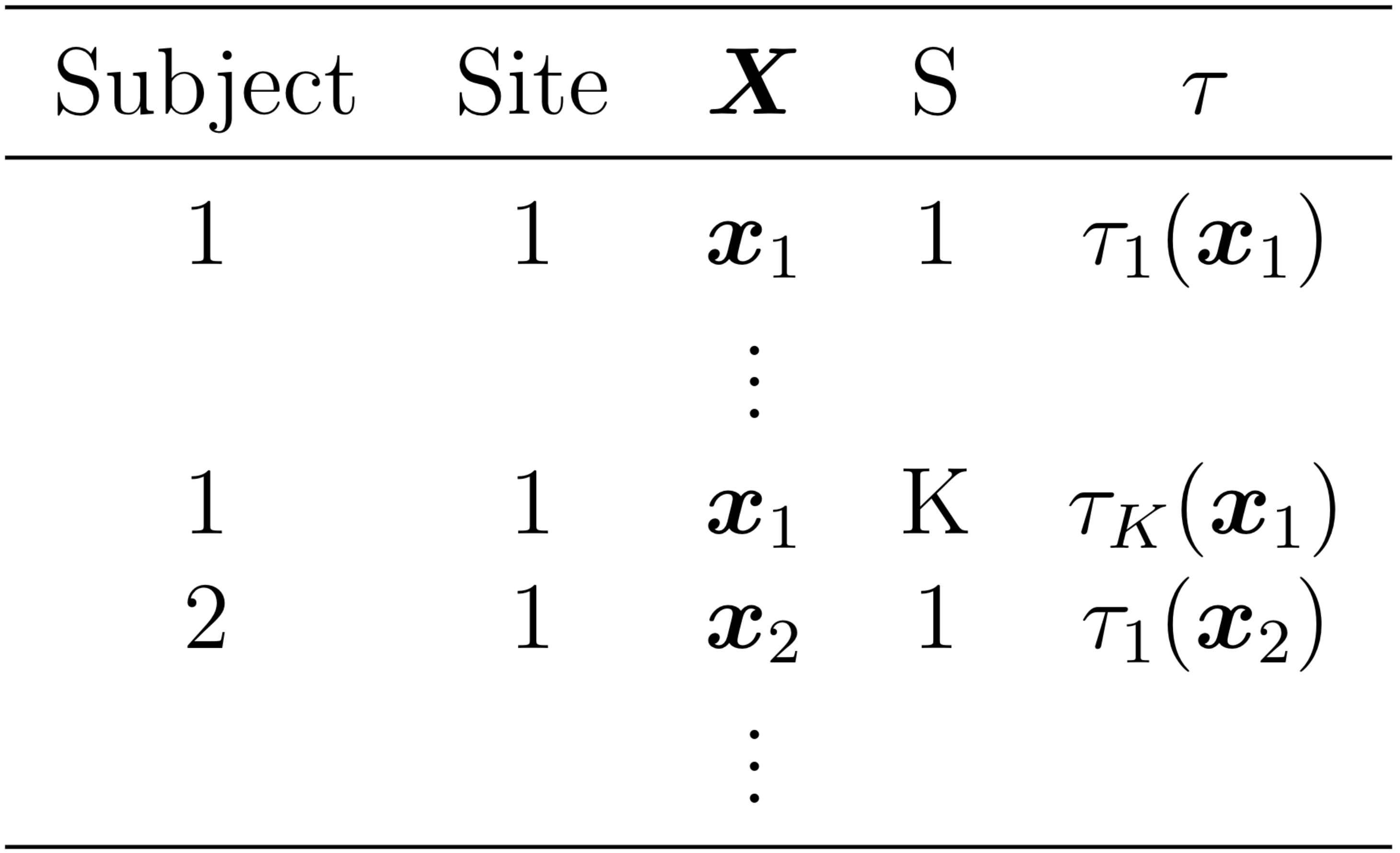}}
  \caption{}
 \end{subfigure}
 \vspace{-0.3cm}
 \caption{(a) Schema of the proposed algorithm. (b) Illustration of the augmented data constructed from the estimation set of site 1.} 
 \label{fig:framework}
\end{figure}

A two-stage model averaging approach is proposed.
We first split $\mathcal{D}_1$, the data in the target site, into a training set and an estimation set indexed by $\{i \in \mathcal{I}_1^{(1)}\}$ and $\{i \in \mathcal{I}_1^{(2)}\}$, respectively.
\emph{1) Local stage:} 
Obtain $\widehat{\tau}_1$ from subjects in $\mathcal{I}_1^{(1)}$. 
Obtain $\widehat{\tau}_k$ from local subjects in $\mathcal{I}_k$, $k = 2, \dots, K$. These $\{\widehat{\tau}_k\}_{k=1}^K$ are then passed to site 1 to get $K$ predicted treatment effects for each subject in $\mathcal{I}_1^{(2)}$, resulting in an augmented data set as shown in Figure~\ref{fig:framework}(b).
\emph{2) Ensemble stage:} A tree-based ensemble model is trained on the augmented data by either an ensemble tree (ET) or an ensemble random forest (EF), with the predicted treatment effects from the previous stage, i.e., $\widehat{\tau}_k(\bx_i)$ as the \emph{outcome}. The site indicator $S$ of which local model is used as well as the subject features $\bx_i$ are fed into the ensemble model as \emph{predictors}. The resulting model will be used to compute our proposed model averaging estimator.
Figure~\ref{fig:framework}(a) illustrates a conceptual diagram of the proposed model averaging framework and structure of the augmented data. 
Note the idea of data augmentation has been used in, e.g., computer vision \citep{wang2017effectiveness}, statistical computing \citep{van2001art}, and imbalanced classification \citep{chawla2002smote}. 
Here the technique is being used to construct weights for model averaging, which will be discussed in the following paragraph. 
Algorithm~\ref{algo:code} provides an algorithmic overview. 
Our method has been implemented as an R package \texttt{ifedtree} available on GitHub (\url{https://github.com/ellenxtan/ifedtree}). 

\setlength{\textfloatsep}{0.4cm} 
\begin{algorithm}[tb]
\small
  \caption{Tree-based model averaging for heterogeneous data sources}
  \label{algo:code}
\begin{algorithmic}
    \FOR{$k=1$ {\bfseries to} $K$}
    \STATE \hfill\COMMENT{Loop through $K$ sites. Can be run in parallel.}
    \STATE Build a local model using site $k$ data. Site 1 model uses its training set only.
    \ENDFOR
    \FOR{$i \in \mathcal{I}_1^{(2)}$} 
    \STATE \hfill\COMMENT{Loop through subjects in site 1 estimation set.}
        \FOR{$k=1$ {\bfseries to} $K$} 
        \STATE \hfill\COMMENT{Loop through $K$ local models.}
            \STATE Predict $\widehat{\tau}_k(\bx_i)$ using local model $k$.
            \STATE $D_{i, k} = [\bx_i, k, \widehat{\tau}_k(\bx_i)]$.
        \ENDFOR
    \ENDFOR
    \STATE Create augmented site 1 data $\mathfrak{D}_{aug, 1}$ by concatenating $D_{i, k}$ vectors.
    \STATE $\widehat{\Tau}_{\text{EF}} (\bx, s) =$ {\sc EnsembleForest}($\mathfrak{D}_{aug, 1}$)
    \STATE \hfill\COMMENT{Or {\sc EnsembleTree} when $B = 1$.}
\end{algorithmic}
\end{algorithm}
\setlength{\floatsep}{0.4cm} 

\textbf{Construction of weights.} A tree-based ensemble is constructed to estimate the weighting functions $\{\omega_{k}\}_{k=1}^K$.
Heterogeneity across sites is explained by including the site index into an augmented training set when building trees.
An intuition of our approach is that sites that are split away from site 1 (by tree nodes) are ignored and the sites that fall into the same leaf node are considered homogeneous to site 1 hence contribute to the estimation of $\tau_1(\bx)$. A splitting by site may occur in any branches of a tree, resulting in an information sharing scheme across sites that is dependent on $\bx$. 
We construct the ensemble by first creating an augmented data 
    $\mathfrak{D}_{aug, 1} = \{\bx_i, k, \widehat{\tau}_k(\bx_i)\}_{i \in \mathcal{I}_1^{(2)}, k \in \mathcal{S}}$,
for subjects in $\mathcal{I}_1^{(2)}$.
The illustration of this augmented site 1 data is given in Figure~\ref{fig:framework}(b). An ensemble is then trained on this data by either a tree or a random forest, 
with the estimated treatment effects $\widehat{\tau}_k(\bx_i)$ as the outcome, and a categorical site indicator of which local model is used along with all subject-level features as predictors, i.e., $(\bx_i, k)$. 
We denote the resulting function as $\Tau(\bx, s)$ which depends on both $\bx$ and site $s$, specifically, 
$\Tau_{\text{ET}} (\bx, s)$ and $ \Tau_{\text{EF}} (\bx, s)$ for ensemble tree (ET) and ensemble forest (EF), respectively. Let $\mathcal{L}(\bx,s)$ denote the final partition of the feature space by the tree to which the pair $(\bx, s)$ belongs. The ET estimate based on the augmented site 1 data can be derived by
\begin{align}\label{eq:ET}
    \widehat \Tau_{\text{ET}} (\bx, s)
    &=  \{|\{(i, k): (\bx_i, k) \in \mathcal{L}(\bx, s)\}_{i \in \mathcal{I}_1^{(2)}, k \in \mathcal{S}}|\}^{-1}  \nonumber\\
    &\sum_{\{(i, k): (\bx_i, k) \in \mathcal{L}(\bx, s)\}_{i \in \mathcal{I}_1^{(2)}, k \in \mathcal{S}}} \widehat{\tau}_k(\bx_i) \nonumber\\
    &=  \sum_{i\in\mathcal{I}_1^{(2)}} \sum_{k=1}^{K} \frac{\mathbbm{1}\{(\bx_i, k)\in \mathcal{L}(\bx, s)\}}{|\mathcal{L}(\bx, s)|} \widehat{\tau}_k(\bx_i).
\end{align}
Intuitively, observations with similar characteristics ($\bx$ and $\bx'$) and from similar sites ($s$ and $s'$) are more likely to fall in the same partition region in the ensemble tree, i.e., $(\bx, s) \in \mathcal{L}(\bx', s')$ or $(\bx', s') \in \mathcal{L}(\bx, s)$. This resembles a \emph{non-smooth kernel} where weights are $1/|\mathcal{L}(\bx, s)|$ for observations that are within the neighborhood of $(\bx, s)$, and 0 otherwise.
The estimator borrows information from neighbors in the space of $\bX$ and $S$. 
The splits of the tree are based on minimizing in-sample MSE of $\widehat\tau$ within each leaf
and pruned by cross-validation over choices of the complexity parameter. 
Since a single tree is prone to be unstable,
in practice, we use random forest to reduce variance and smooth the partitioning boundaries.
By aggregating $B$ ET estimates each based on a subsample of the augmented data, $\{\widehat{\Tau}^{(b)}\}_{b=1}^B$, an EF estimate can be constructed by
\begin{align}
       \widehat \Tau_{\text{EF}} (\bx, s) 
& = \frac{1}{B} \sum_{b=1}^{B} \widehat \Tau^{(b)} (\bx, s) \nonumber \\
& = \sum_{i\in \mathcal{I}_1^{(2)}} \sum_{k=1}^{K} \lambda_{i,k}(\bx, s) \widehat{\tau}_k(\bx_i), \label{eq:EF}\\
    \text{where~} &  \lambda_{i,k}(\bx, s) = \frac{1}{B} \sum_{b=1}^{B} \frac{\mathbbm{1}\{(\bx_i, k)\in \mathcal{L}_b(\bx, s)\}}{|\mathcal{L}_b(\bx, s)|}.\nonumber
\end{align}
The form of $\widehat \Tau^{(b)} (\bx, s)$ closely follows \eqref{eq:ET} but is based on a subsample of $\mathfrak{D}_{aug, 1}$. The weights, $\lambda_{i,k}(\bx, s)$, are similar to that in \eqref{eq:ET}, and can be viewed as kernel weighting that defines an adaptive neighborhood of $\bx$ and $s$. 
We then obtain the model averaging estimates defined in \eqref{eq:agg} by fixing $s = 1$ such that $\widehat \tau_{\text{ET}}^*(\bx) = \widehat\Tau_{\text{ET}} (\bx, s=1)$ or $\widehat \tau_{\text{EF}}^*(\bx) = \widehat\Tau_{\text{EF}} (\bx, s=1)$.
The weight functions $\{\omega_{k}(\bx)\}_{k=1}^K$ for $\widehat\tau^*(\bx)$ can be immediately obtained from the ET or EF by
\begin{align*}
    \widehat\tau_{\text{ET}}^*(\bx) = \ &  \widehat \Tau_{\text{ET}} (\bx, 1) = \sum_{k=1}^K \widehat \omega_{k}(\bx) \widehat \tau_k(\bx),  \\
    \text{where~~} & \widehat \omega_{k}(\bx) = \sum_{i\in\mathcal{I}_1^{(2)}} \frac{\mathbbm{1}\{(\bx_i, k)\in \mathcal{L}(\bx, 1)\}}{|\mathcal{L}(\bx, 1)|}; \nonumber \\
   \widehat\tau_{\text{EF}}^*(\bx) = \ &  \widehat \Tau_{\text{EF}} (\bx, 1) = \sum_{k=1}^K \widehat \omega_{k}(\bx) \widehat \tau_k(\bx),  \\
    \text{where~~} & \widehat \omega_{k}(\bx) = \sum_{i\in\mathcal{I}_1^{(2)}} \lambda_{i,k}(\bx, 1). 
\end{align*}
It can be verified that $\sum_{k=1}^{K} \widehat \omega_k(\bx) = 1$ for all $\bx$.
As our simulations in Section \ref{sec:simulation} show, $\widehat \tau^*$ improves the local functional estimate $\widehat\tau_1$.
We set $B=2,000$ throughout the paper.
Tree and forest estimates are obtained by R packages \verb|rpart| and \verb|grf|, respectively.

\textbf{Interpretability of weights.}
The choice of tree-based models naturally results in such kernel weighting $w_k(\bx)$ \citep{athey2019generalized}, which are not accessible by other ensemble techniques.
Such explicit and interpretable weight functions could deliver meaningful rationales for data integration.
For example, under scenarios where there exists extreme global heterogeneity (as shown in Section~\ref{sec:simulation} when $c$ is large), $w_k(\bx)$ can be used as a diagnostic tool to decide which external data sources should be co-used. 
Weights close to $0$ inform against model transportability, and they are adaptive to subject-level features $\bx$ so that decisions can be made based on the subpopulations of interest.




\subsection{Local Models: Obtaining $\widehat\tau_k$} \label{sec:local}

Estimate of $\tau_k(\bx)$ at each local site must be obtained separately before the ensemble. 
Our proposed ensemble framework can be applied to a general estimator of $\tau_k(\bx)$. For each site, the local estimate could be obtained using different methods. 
Recently, there has been many work dedicated to the estimation of individualized treatment effects \citep{athey2016recursive,wager2018estimation,hahn2020bayesian,kunzel2019metalearners,nie2020quasioracle}. 
As an example, we consider using the causal tree (CT) \citep{athey2016recursive} to estimate the local model at each site. CT is a non-linear learner that \textit{(i)} allows different types of outcome such as discrete and continuous, and can be applied to a broad range of real data scenarios; \textit{(ii)} can manage hundreds of features and high order interactions by construction; \textit{(iii)} can be applied to both experimental studies and observational studies by propensity score weighting or doubly robust methods. 
CT is implemented in the R package \verb|causalTree|. We also explore another estimating option for local models in Appendix~\ref{suppl-sec:sim}. 



\subsection{Asymptotic Properties} \label{sec:consist}

We provide consistency guarantee of the proposed estimator $\widehat \Tau_{\text{EF}}$ for the true target $\tau_1$. 
Assuming point-wise consistent local estimators are used for $\{\tau_k\}_{k=1}^{K}$, EF with subsampling procedure described in Appendix~\ref{suppl-sec:consist} is consistent. 
\begin{theorem}\label{Theorem}
Suppose the subsample used to build each tree in an ensemble forest is drawn from different subjects of the augmented data and the following conditions hold:

  (i) Bounded covariates: Features $\bX_i$ and the site indicator $S_i$ are independent and have a density that is bounded away from 0 and infinity.
  
  (ii) Lipschitz response: the conditional mean function $\mathbb{E}[ \Tau|\bX=\bx,S=1]$ is Lipschitz-continuous.
  
  (iii) Honest trees: trees in the random forest use different data for placing splits and estimating leaf-wise responses.

Then $\widehat \Tau_{\text{EF}}(\bx,1) \overset{p}{\to} \tau_{1}(\bx)$, for all $\bx$, as $\min_k n_k \to \infty$. Hence, $\widehat \tau_{\text{EF}}^*(\bx) \overset{p}{\to} \tau_1(\bx).$
\end{theorem}
The conditions and a proof of Theorem \ref{Theorem} is given in Appendix~\ref{suppl-sec:consist}.
To demonstrate the consistency properties of our methods, we add in Appendix~\ref{suppl-sec:sim} oracle versions of ET and EF estimators, denoted as ET-oracle and EF-oracle, which use the ground truth of local models $\{\tau_k\}_{k=1}^K$ in estimating $\{\widehat \omega_k\}_{k=1}^K$.
This removes the uncertainty in local models. 
The remaining uncertainty only results from the estimation of the ensemble weights, and we see both oracle estimators achieve minimal MSE.
Section~\ref{sec:simulation} gives a detailed evaluation of the finite sample performance.

\begin{figure*}[tb]
\centering
 \begin{subfigure}{0.49\textwidth}
  \centerline{\includegraphics[width=\linewidth]{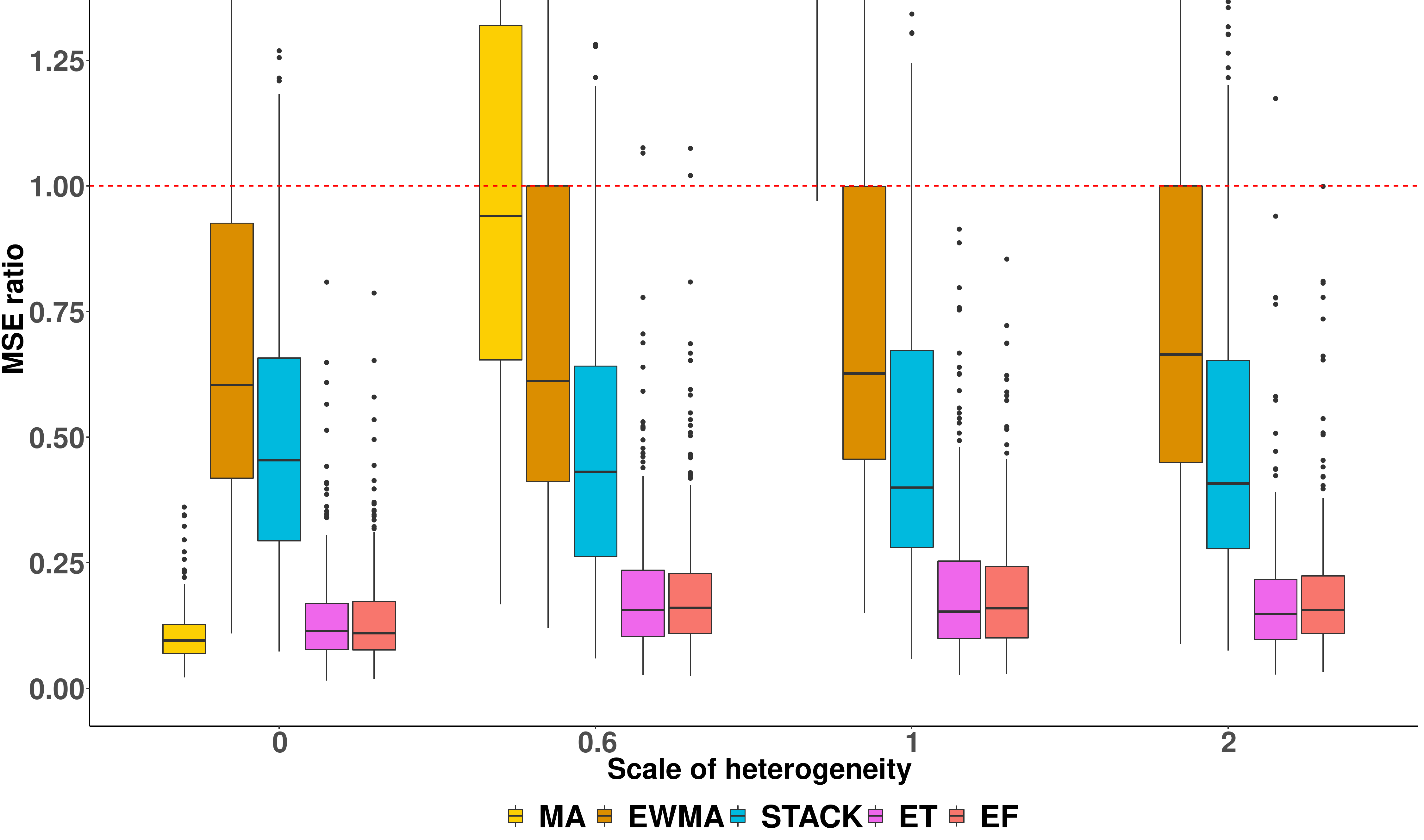}}
  \caption{}
 \end{subfigure}
 \begin{subfigure}{0.49\textwidth}
  \centerline{\includegraphics[width=\linewidth]{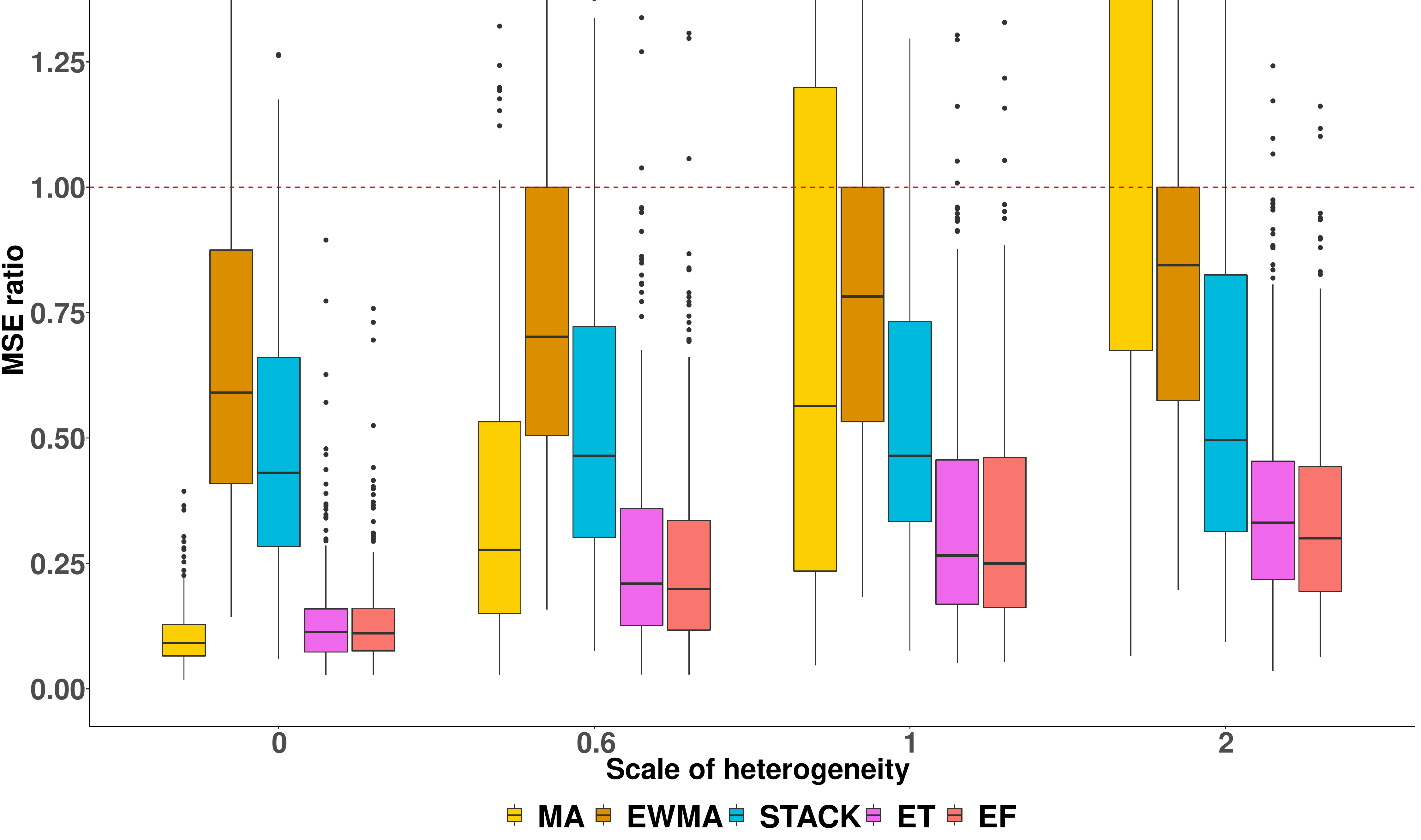}}
  \caption{}
 \end{subfigure}
 \vspace{-0.3cm}
 \caption{
 Box plots of MSE ratios of CATE estimators, respectively, over LOC, for \textbf{(a) discrete grouping} and \textbf{(b) continuous grouping} across site. 
  Different colors imply different estimators, and x-axis, i.e., the value of $c$, differentiates the scale of global heterogeneity. The red dotted line denotes an MSE ratio of 1. 
  MA performance is truncated due to large MSE ratios. 
The proposed ET and EF achieve smaller MSE ratios compared to standard model averaging or ensemble methods and are robust to heterogeneity across settings. 
}
\vspace{-0.3cm}
\label{fig:sim_box}
\end{figure*}

\section{Simulation Study}
\label{sec:simulation}

Monte Carlo simulations are conducted to assess the proposed methods. We specify $m(\bx, k)$ as the conditional outcome surface and $\tau(\bx, k)$ as the conditional treatment effect for individuals  with features $\bx$ in site $k$. The treatment propensity is specified as $e(\bx) = \Pr(Z=1|\bX=\bx)$. 
The potential outcomes can be written as $Y_i = m(\bX_i, S_i) + \{ Z_i - e(\bX_i) \} \tau(\bX_i, S_i) + \epsilon_i$, following notations in \citet{robinson1988root,athey2016recursive,wager2018estimation,nie2020quasioracle}. 
The mean function is $m(\bx, k) = \frac{1}{2} x_1 + \sum_{d=2}^4 x_d +(x_1 - 3) \cdot c \cdot U_k$, and the treatment effect function is specified as 
$$ \tau(\bx, k) = \mathbbm{1}\{x_1 > 0\} \cdot x_1 +(x_1 - 3) \cdot c \cdot U_k,$$
where $z=0,1$, $U_k$ denotes the global heterogeneity due to site-level confounding, controlled by a scaling factor $c$, and $\epsilon_i \sim N(0, 1)$. 
Features follow $\bX_i \sim {N}(\bm{0},\bm{I}_D)$, where $D=5$, and are independent of $\epsilon_i$.
The simulation setting within each site (with $k$ fixed) is motivated by designs in \citet{athey2016recursive}. Features in $\tau$ are determinants of treatment effect while those in $m$ but not in $\tau$ are prognostic only. 
The data are generated under a distributed data networks. 
We assume there are $K=20$ sites in total, each with a sample size $n=500$. 
In our main exposition, we consider an experimental study design where treatment propensity is $e(\bx) = 0.5$, i.e., individuals are randomly assigned to treatment and control.
Variations of the settings above are discussed, with results presented in Appendix~\ref{suppl-sec:sim}. 

\textbf{Global heterogeneity: discrete and continuous.} 
Two types for global heterogeneity are considered by the choice of $U_k$. 
For \textbf{\emph{discrete grouping}}, we assume there are two underlying groups among the $K$ sites $U_k \sim Bernoulli(0.5)$. Specifically, we assume odd-index sites and even-index sites form two distinct groups $\mathcal{G}_1 = 
\{1, 3, \dots, K-1\}$; $\mathcal{G}_2 = 
\{2, 4\dots, K\}$ such that $U_{k\in\mathcal{G}_1} = 0$ and $U_{k\in\mathcal{G}_2} = 1$. 
Sites from similar underlying groupings have similar treatment effects and mean effects, while sites from different underlying groupings have different treatment effects and mean effects. 
For \textbf{\emph{continuous grouping}}, we consider $U_k \sim Unif[0,1]$. 
We vary the scales of the global heterogeneity under the discrete and continuous cases, respectively, with $c$ taking values $c \in \{0,0.6,1,2\}$. A $c=0$ implies all data sources are homogeneous. 
In other words, Assumption~2 is satisfied when $c = 0$ but not when $c > 0$. 

\textbf{Compared estimators and evaluation.} The proposed approaches ET and EF are compared with several competing methods. 
\textbf{LOC:} A local CT estimator
that does not utilize external information. 
It is trained on $\mathcal{I}_1$ only, combining training and estimation sets.
\textbf{MA:} A naive model averaging method
with weights ${\omega}_{k}^{\text{MA}} = 1/k$. This approach assumes models are homogeneous. 
\textbf{EWMA:}
We consider a modified version of EWMA that can be used for CATE.
We obtain an approximation of $\tau_1(\bx)$ by fitting another local model using the estimation set of site 1, denoted by $\widetilde\tau_1(\bx)$. 
Its weights are given by $${\omega}_{k}^{\text{EWMA}} = \frac{\exp\{- \sum_{i \in \mathcal{I}_1^{(2)}}(\widehat\tau_k(\bx_i) - \widetilde\tau_1(\bx_i))^2\} }{ \sum_{\ell=1}^{K} \exp\{- \sum_{i \in \mathcal{I}_1^{(2)}}(\widehat\tau_\ell(\bx_i) - \widetilde\tau_1(\bx_i))^2\} }.$$ 
\textbf{STACK:} A stacking ensemble, which is a linear ensemble of predictions of several models \citep{breiman1996stacked}.
To our end, we regress $\widetilde\tau_1(\bx)$ on the predictions of the estimation set in site 1 from each local model, $\{\widehat\tau_1(\bx), \dots, \widehat\tau_k(\bx) \}$. The stacking weights are not probabilistic hence not directly interpretable.
We report the empirical mean squared error (MSE) of these methods over an independent testing set of sample size $n_{te}=2000$ from site 1. 
$\mbox{MSE}(\widehat{\tau}) = n_{te}^{-1}\sum_{i=1}^{n_{te}} \{ \widehat{\tau}(\bx_i) - \tau_1(\bx_i) \}^2. $
Each simulation scenario is repeated for 1000 times.  
Experiments are performed on a 6-core Intel Xeon CPU E5-2620 v3 2.40GHz equipped with 64GB RAM.

\begin{table}[tb]
\centering
\caption{MSE ratios of EF over LOC. As $n$ increases, model averaging becomes more powerful due to better estimation of $\tau_k$, and is more pronounced when $c$ is small.}
\label{fig:ratio}
\resizebox{0.9\columnwidth}{!}{
\begin{tabular}{@{}cccccc@{}}
\toprule
\multicolumn{1}{l}{}  &  & $c=0$  & $c=0.6$ & $c=1$  & $c=2$  \\ \midrule
\multirow{3}{*}{\shortstack{Discrete\\ grouping}} & $n=100$  & 0.57 & 0.59  & 0.61 & 0.59 \\
& $n=500$  & 0.12 & 0.17  & 0.17 & 0.16 \\
& $n=1000$ & 0.07 & 0.12  & 0.12 & 0.13 \\
\multirow{3}{*}{\shortstack{Continuous\\ grouping}} & $n=100$  & 0.54 & 0.59  & 0.63 & 0.69 \\
& $n=500$  & 0.11 & 0.24  & 0.31 & 0.34 \\
& $n=1000$ & 0.08 & 0.17  & 0.21 & 0.26 \\ \bottomrule
\end{tabular}
}
\vspace{-0.1cm}
\end{table}

\begin{figure*}[tb]
  \begin{subfigure}{0.3\textwidth}
    \centering
    \includegraphics[width=\linewidth]{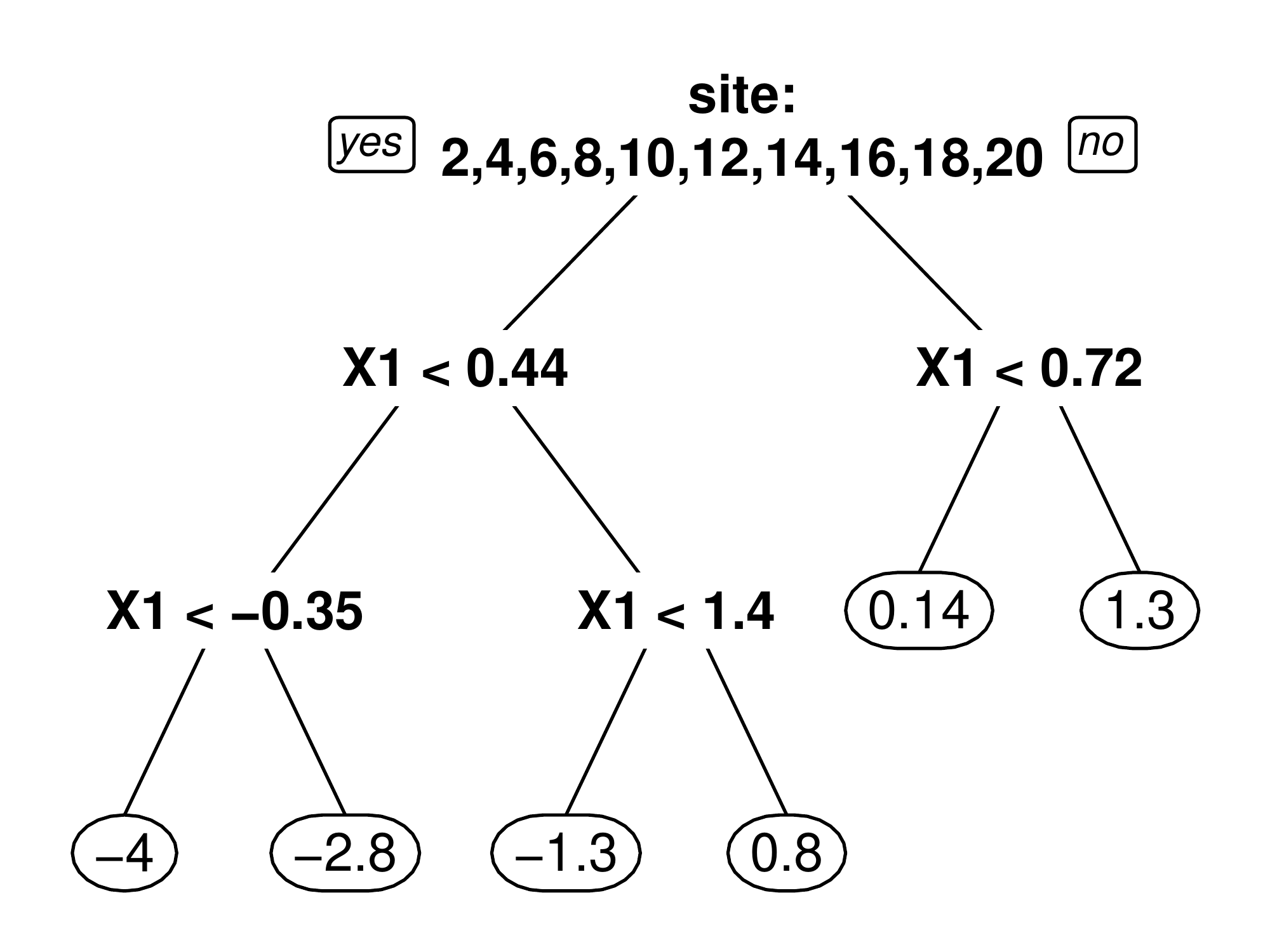}
    \caption{}
  \end{subfigure}
  \begin{subfigure}{0.325\textwidth}
    \centering
    \includegraphics[width=\linewidth]{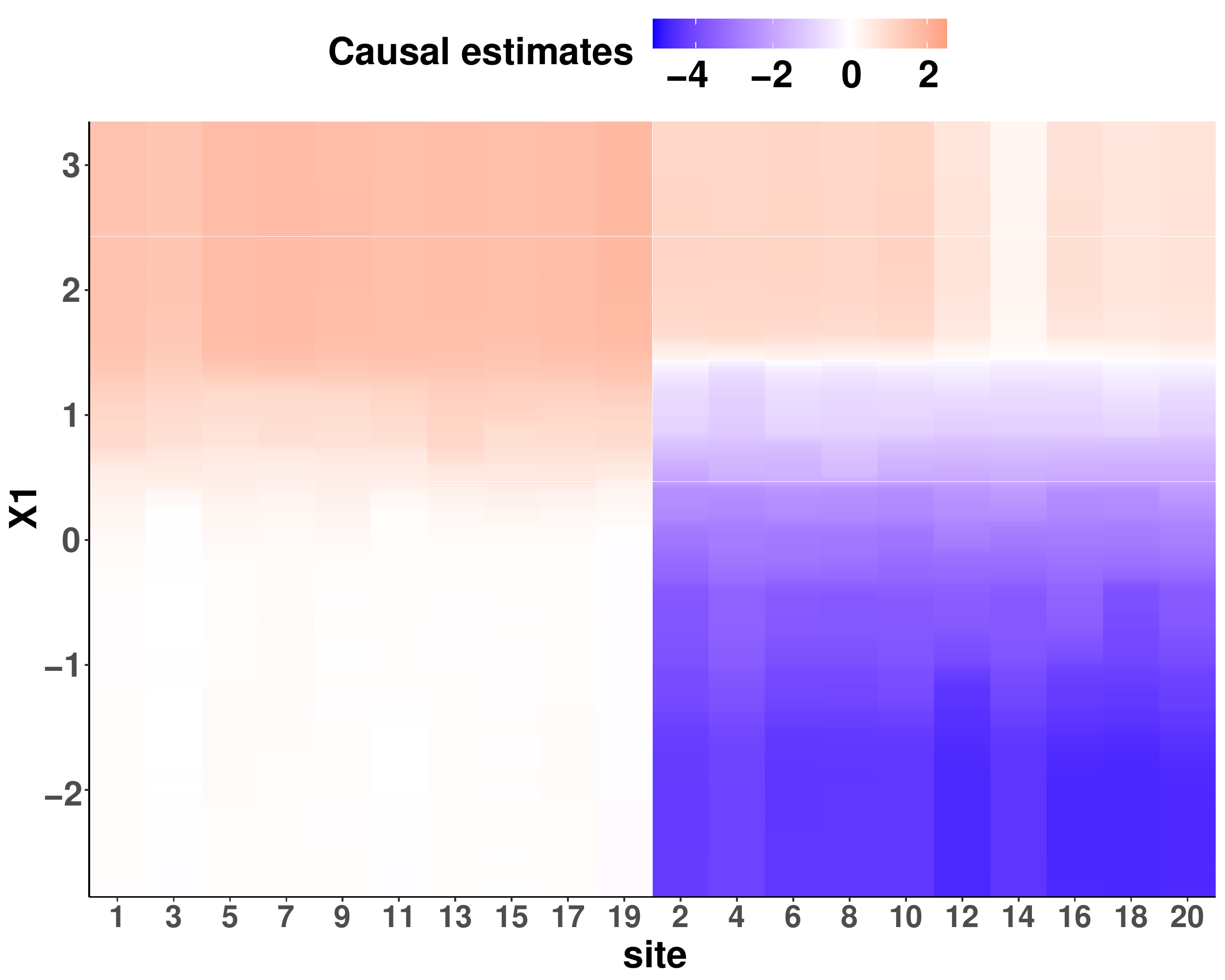}
    \caption{}
  \end{subfigure}
  \begin{subfigure}{0.37\textwidth}
    \centering
    \includegraphics[width=\linewidth]{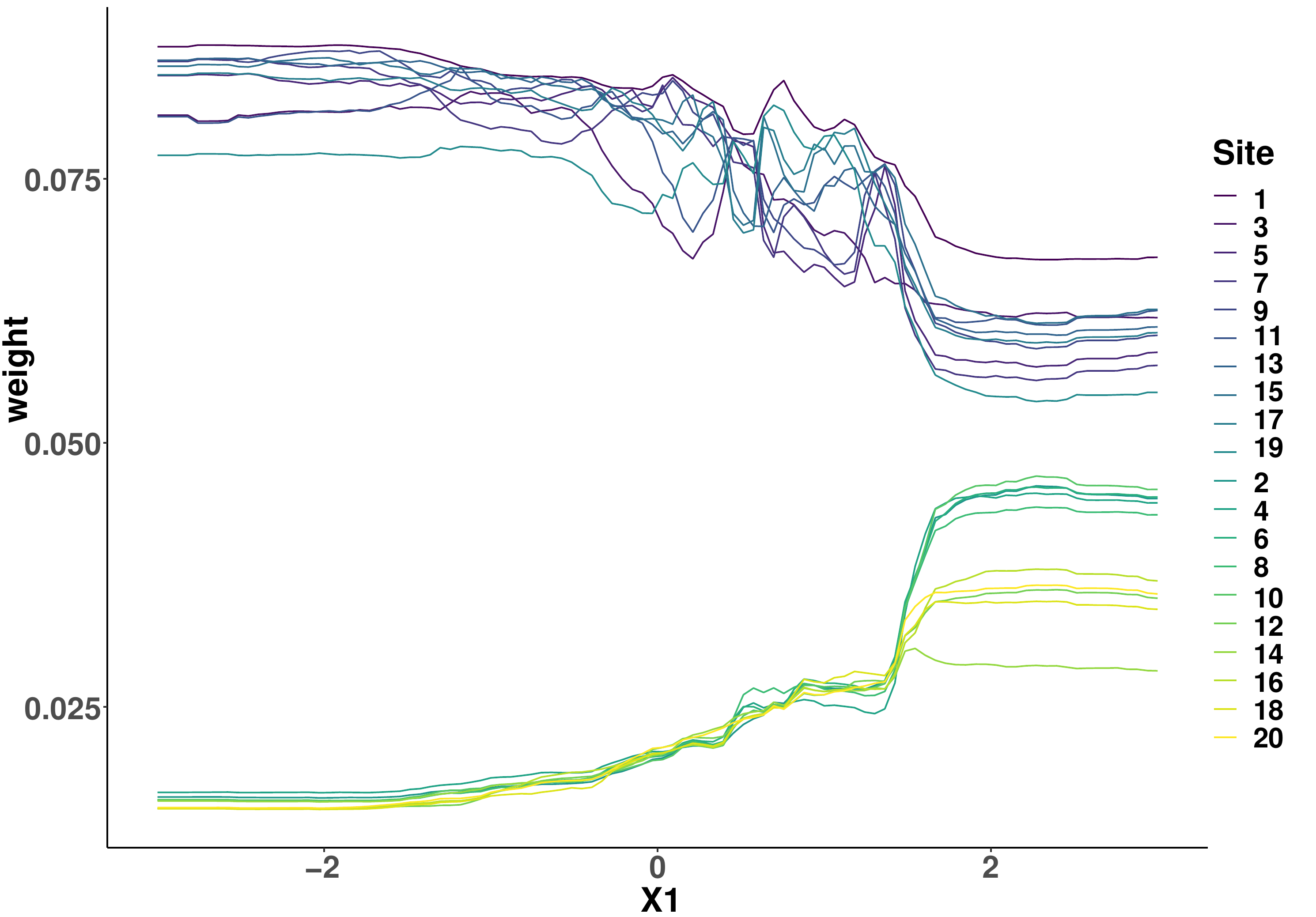}
    \caption{}
  \end{subfigure}
  \medskip

  \begin{subfigure}{0.3\textwidth}
    \centering
    \includegraphics[width=\linewidth]{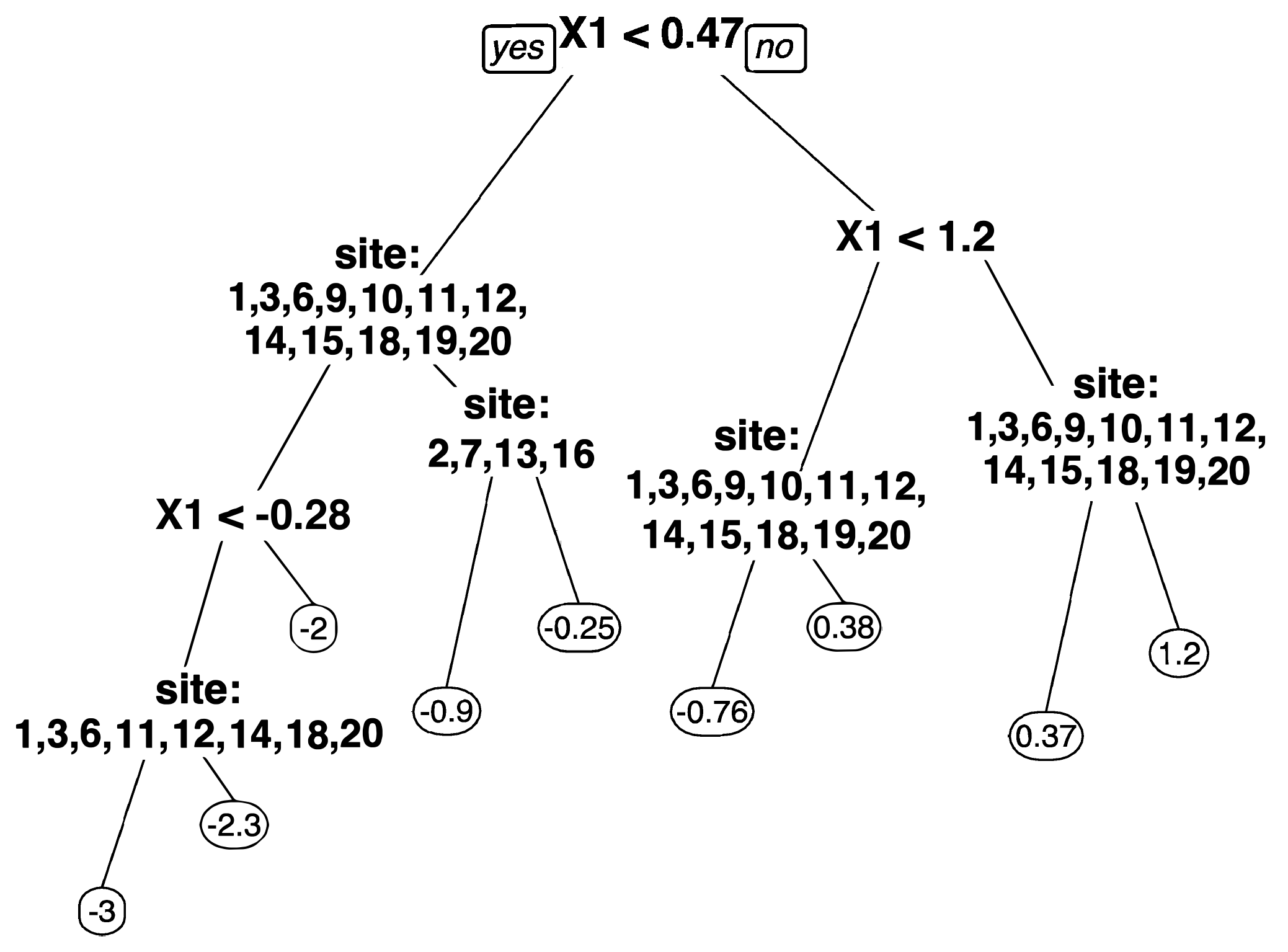}
    \caption{}
  \end{subfigure}
  \begin{subfigure}{0.325\textwidth}
    \centering
    \includegraphics[width=\linewidth]{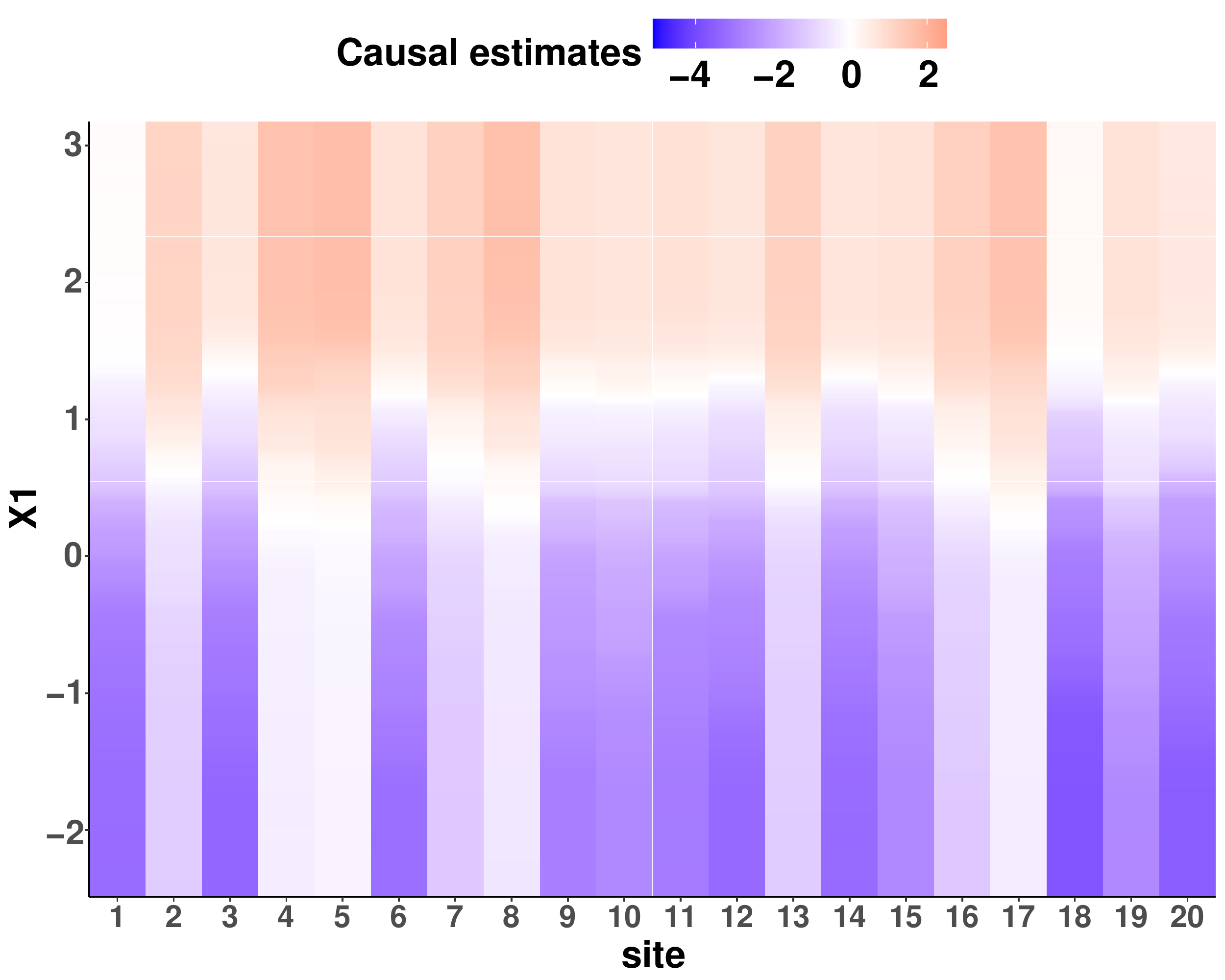}
    \caption{}
  \end{subfigure}
  \begin{subfigure}{0.37\textwidth}
    \centering
    \includegraphics[width=\linewidth]{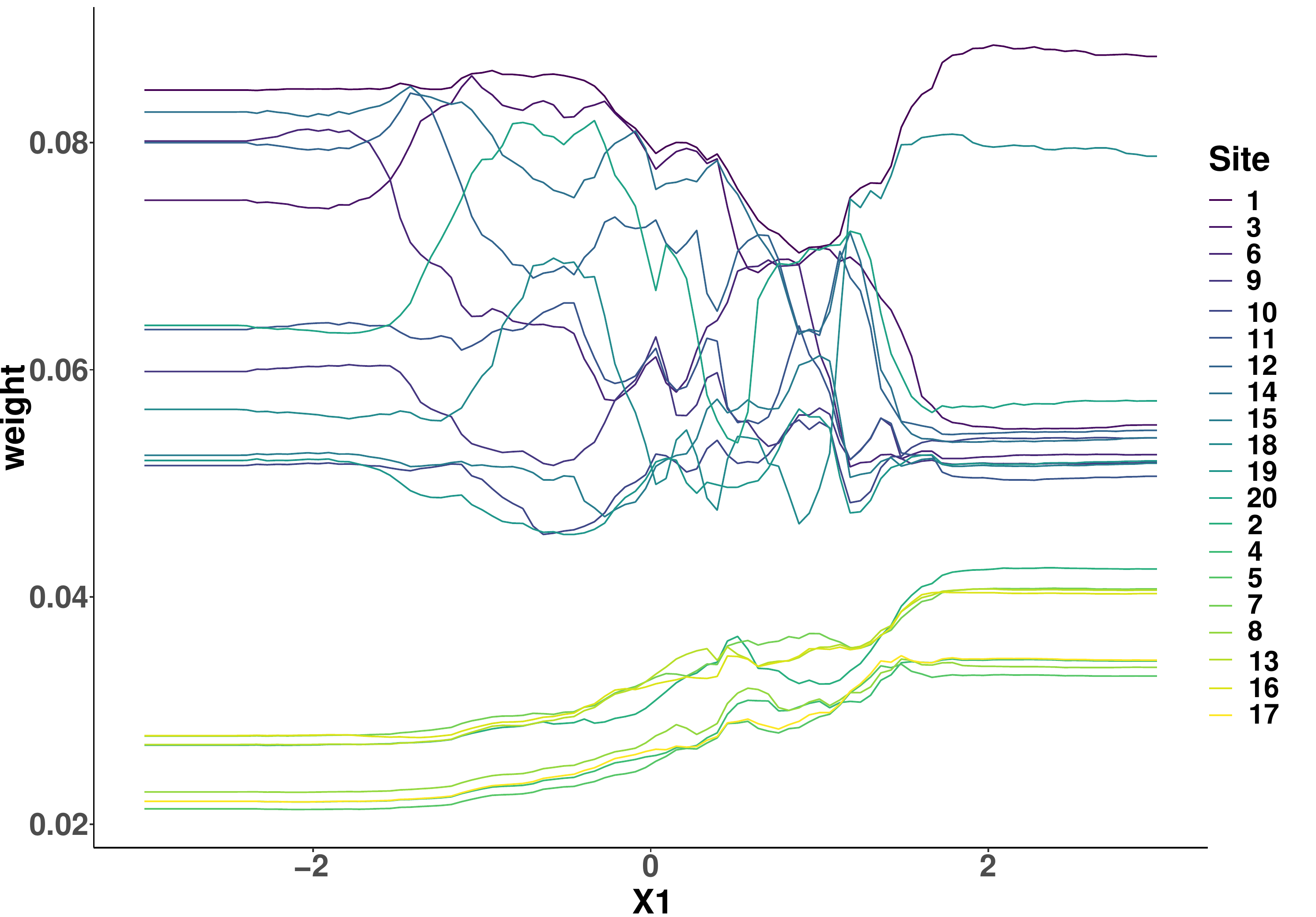}
    \caption{}
  \end{subfigure}
  \vspace{-0.3cm}
  \caption{
 Visualization of simulation 
 results under \textbf{discrete grouping (a,b,c)} and \textbf{continuous grouping (d,e,f)} when  $c=1$. (a) and (d) visualize the proposed ETs where the site indicator and $X_1$ are selected as splitting variables, which is consistent with the underlying data generation process. 
 (b) and (e) show the predicted treatment effects of the proposed EFs varying $X_1$ in each site, marginalized over all other features. (b) is arranged according to the true grouping, odd sites versus even sites. 
 The plot recovers the pattern of local and global heterogeneity. 
 (c) and (f) plot the interpretable model averaging weights in EFs over $X_1$. 
 The weights of site 1 have a relatively large contribution to the weighted estimator while models from other sites have different contributions for different $X_1$ depending on their similarity in $\tau(\bx,k)$ to that in site 1. 
 Corresponding ET and EF show consistent patterns and recover the true grouping.
  }
  \label{fig:sim_visual}
  \vspace{-0.3cm}
\end{figure*}


\textbf{Estimation performance}. 
Figure~\ref{fig:sim_box} shows the performance of the proposed estimators and the competing estimators, using LOC as the benchmark.
The proposed ET and EF show the best performance in terms of the mean and variation of MSE among other estimators when $c > 0$, and comparable to equal weighting MA when $c = 0$.
Although, a forest is more stable than a tree in practice,
both ET and EF give similar results because the true model is relatively simple and can be accurately estimated by a single ensemble tree under the given sample size. 
Although asymptotically consistency, under finite sample, bias exists in local models and leads to biased model averaging estimates. 
While explicit quantification of bias and variance remains challenging due to extra uncertainty carried forward from the local estimates, we demonstrated that the proposed estimators can improve upon the local models under small sample size via Table~\ref{fig:ratio}.
It shows the MSE ratio of EF over LOC as a measure of gain resulting from model averaging by varying $n =100, 500, 1000$. 
The decrease in MSE ratio as $n$ increases, regardless of the choice of $c$, is consistent with our asymptotic results in Theorem~\ref{Theorem}.
This is due to a bias-and-variance trade-off in the ensemble that ensures a small MSE, which remains smaller than that in LOC despite varying $n$.
It also shows our method is robust to the existence of local uncertainty.

\textbf{Visualization of information borrowing.} Figure~\ref{fig:sim_visual} visualizes the proposed ET and EF. 
In (a) and (d), the site indicator and $X_1$ appear as splitting variables in the ETs, which is consistent with the data generation process. 
The estimated treatment effect (b) and (e) reveals the pattern of transportability across sites and with respect to $X_1$. 
Panels (c) and (f) plot the model averaging weights in EFs over $X_1$. Site 1 has a relatively large contribution to the weighted estimator while models from other sites have different contributions at different values of $X_1$ depending on their similarity in $\tau(\bx,k)$ to that in site 1. 
Corresponding ET and EF show consistent patterns.

\begin{figure*}[htp]
\centering
 \begin{subfigure}{0.2\textwidth}
  \centerline{\includegraphics[width=\linewidth]{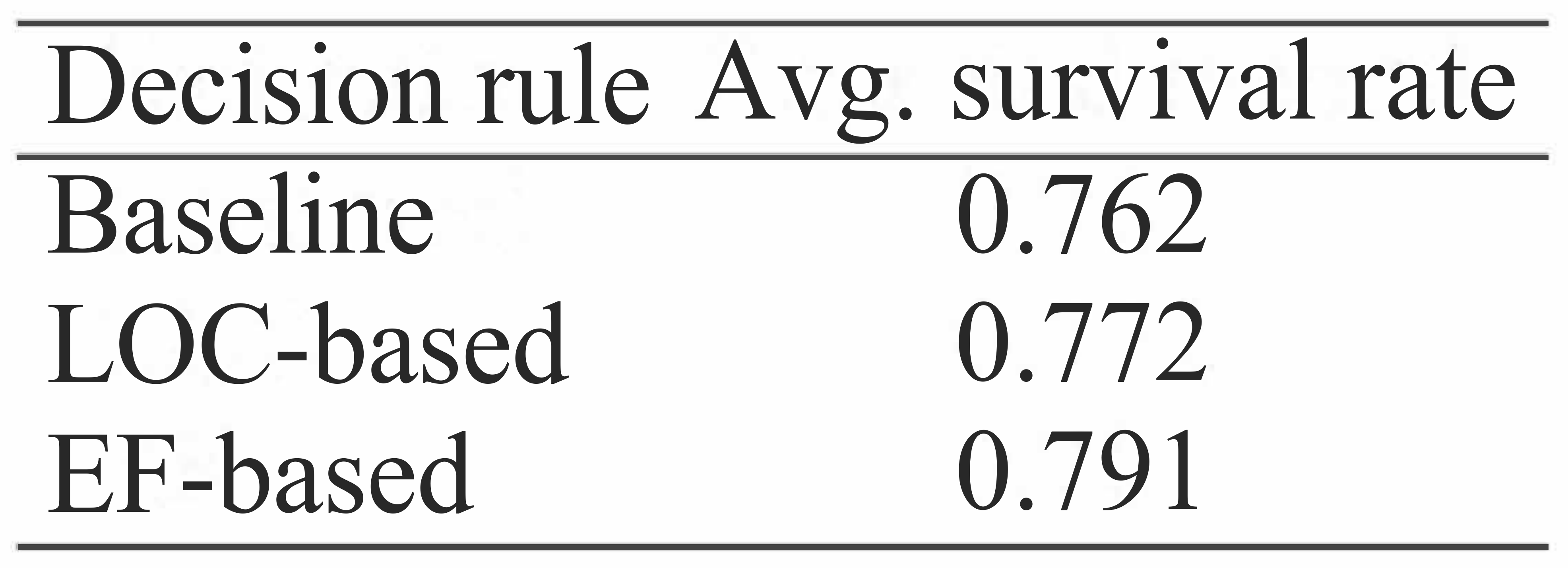}}
  \caption{}
 \end{subfigure}
 \begin{subfigure}{0.39\textwidth}
  \centerline{\includegraphics[width=\linewidth]{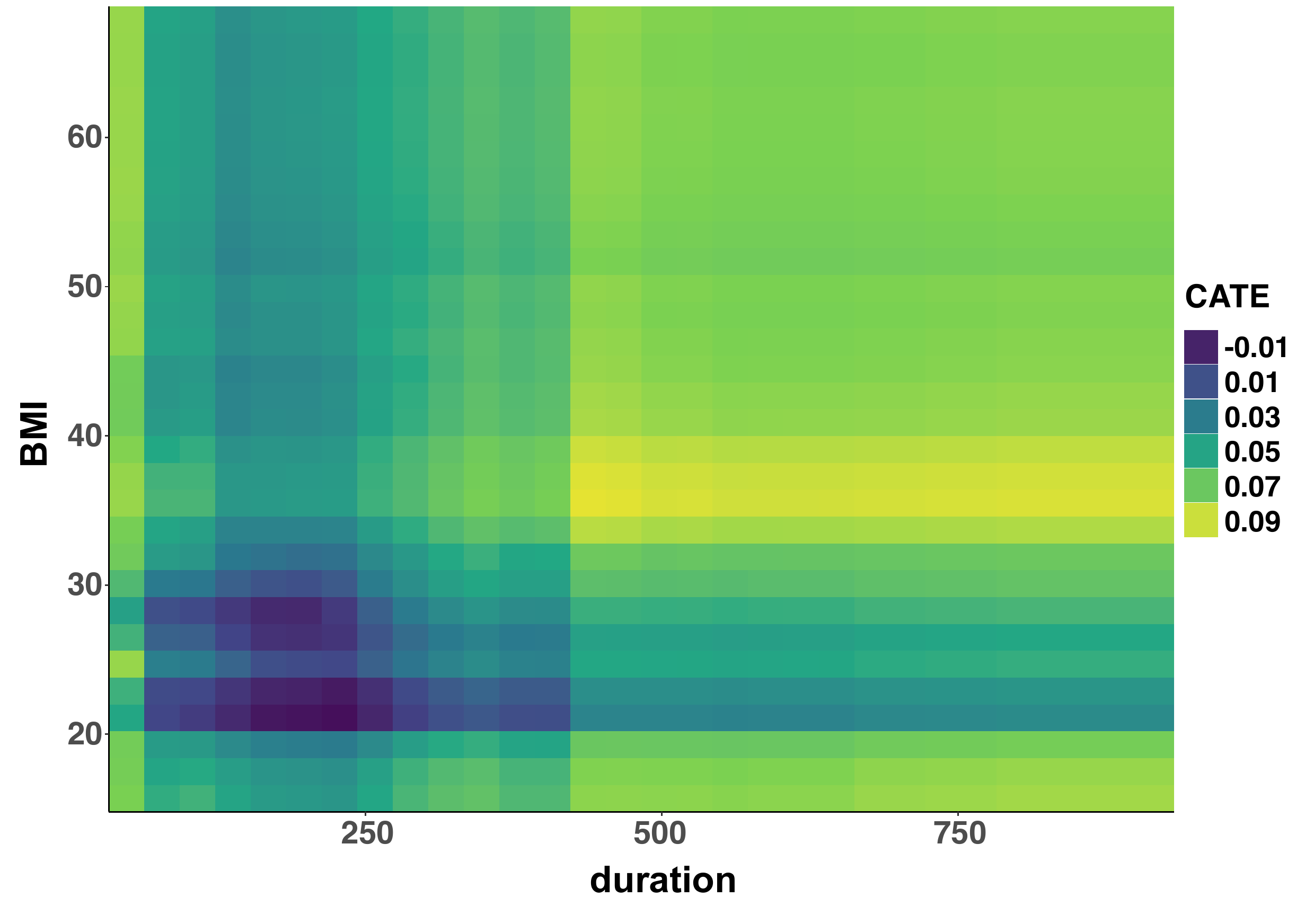}}
  \caption{}
 \end{subfigure}
 \begin{subfigure}{0.39\textwidth}
  \centerline{\includegraphics[width=\linewidth]{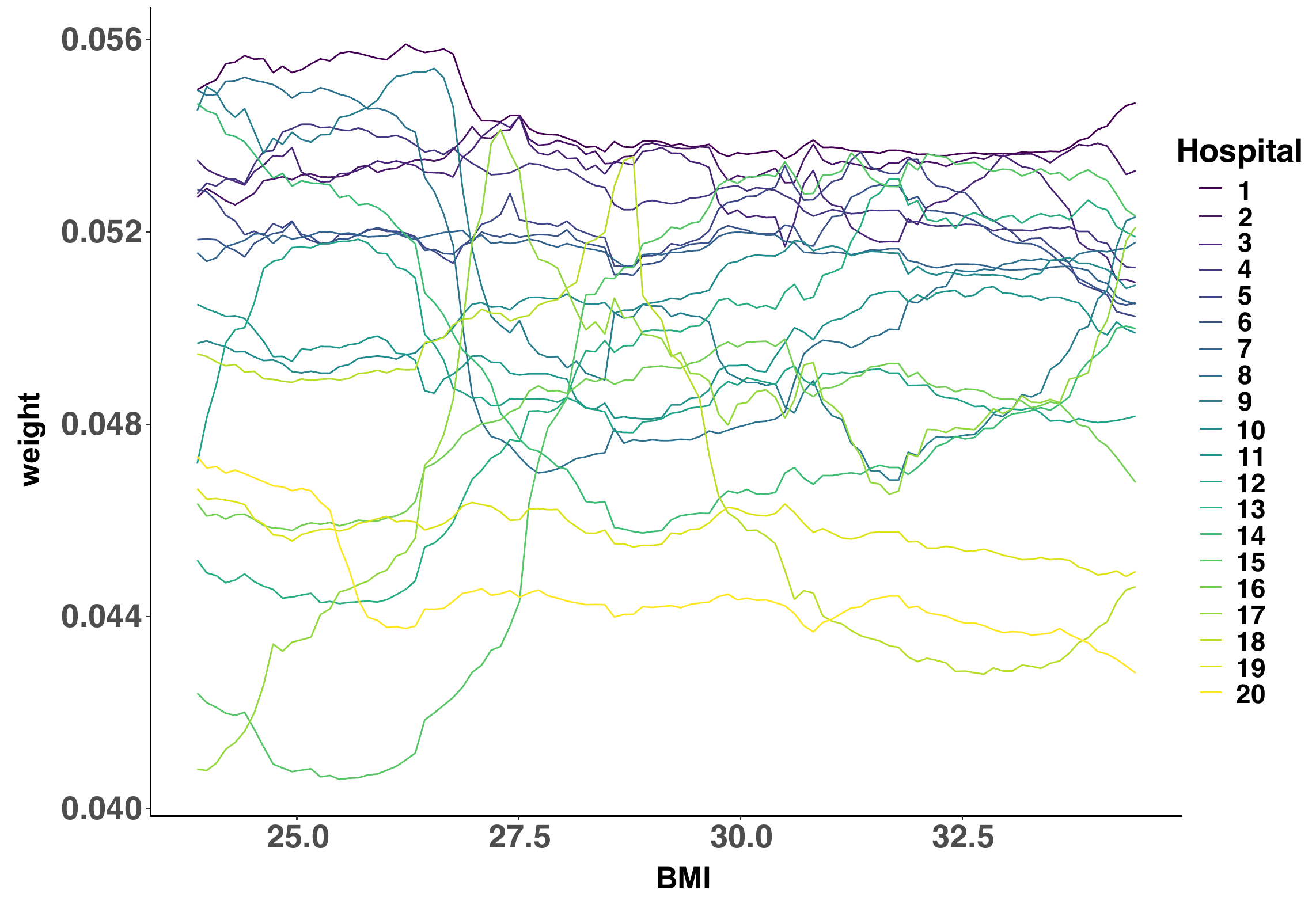}}
  \caption{}
 \end{subfigure}
 \vspace{-0.3cm}
 \caption{
 Application to estimating treatment effects of oxygen therapy on survival. 
 (a) Expected survival of treatment decision following different estimators. 
 The proposed EF shows the largest gain in improving survival rate, more promising than LOC and baseline. 
 (b) Estimated treatment effects varying duration and BMI, two important features in the fitted EF. 
 Patients with a BMI around 35, and a duration above 400 benefited the most. 
 (c) Visualization of data-adaptive weights in the estimated EF varying BMI. 
 Hospitals with a larger bed capacity tend to contribute more, the data of which might be more similar to hospital 1.
}
\vspace{-0.2cm}
\label{fig:real}
\end{figure*}

\textbf{Additional simulations.}
The detailed results of these additional simulations are included in Appendix~\ref{suppl-sec:sim}.
\emph{\textbf{1) Connection to supervised learning.}} The uniqueness of averaging $\tau_k(\bx)$ as opposed to supervised learning that averages prediction models $f_k(\bx)$ is that the outcome of $f_k(\bx)$ is immediately available. 
In our case, an additional estimation step is needed to construct the model averaging weights. 
We provide a comparison among estimators that utilize the ground truth $\{\tau_k(\bx)\}_{k=1}^K$ (denoted as ``-oracle'') when computing ensemble weights.
This mimics the case of supervised learning where weights are based on observed outcomes. 
Oracle methods achieve smaller MSE ratios; the pattern is consistent with Table~\ref{fig:ratio}.
\emph{\textbf{2) Simulation under observational studies.}} 
We also consider the treatment generation mechanism under an observational design. 
Specifically, the propensity is given as $e(\bx) = \text{expit}(0.6x_1)$. We consider both a correctly specified propensity model using a logistic regression of $Z$ on $X_1$ and a misspecified propensity model with a logistic regression of $Z$ on all $\bX$. In general, the proposed estimators obtain the best performance with similar results as in Figure~\ref{fig:sim_box}. With the correctly specified propensity score model, the local estimator is consistent in estimating $\tau_k(\bx)$, the proposed framework is valid. When the propensity model is misspecified, extra uncertainty is carried forward from the local estimates, but the proposed estimators can still improve upon LOC. This is due to a bias-and-variance trade-off that leads to small MSE, which remains smaller than the local models. 
\emph{\textbf{3) Covariate dimensions.}} 
Besides $D=5$, we consider other choices of covariate dimension including $D=20, 50$. With a higher dimension, the MSE ratio between the proposed estimates and LOC estimates increases but the same pattern across methods persists.
\emph{\textbf{4) Unequal sample size at each site.}} 
In the distributed date network, different sites may have a different sample size $n_k$. Those with a smaller sample size may not be representative of their population, leading to an uneven level of precision for local causal estimates. We consider a simulation setting where site 1 has a sample size of $n_1=500$ while other site $n_2,\ldots,n_K$ has a sample size of 200. Results show that the MSE ratio between the proposed estimates and LOC estimates increases compared to the scenario where the sample size in all sites are 500. However, the proposed estimators still enjoy the most robust performance. 
This also shows our method is robust to the existence of local uncertainty. 
\emph{\textbf{5) Different local estimators.}} We stress that other consistent estimators could be used as the local model. Options such as causal forest \citep{wager2018estimation} are explored varying the sample size at local sites. Similar performance is observed as in Figure~\ref{fig:sim_box}.
\emph{\textbf{6) Further comparisons to non-adaptive ensemble.}} Here we provide a brief discussion of the implications of the proposed method and how it differs from non-adaptive methods such as stacking. 
Although unrealistic, when the true weights are non-adaptive, the performance may be similar. Plus, our learned weights can be used to examine adaptivity, as shown in Figure~\ref{fig:sim_visual}(c,f) and Figure~\ref{fig:real}(c). 
Stacking is shown to be more robust than non-adaptive model averaging in case of model misspecification. See discussion in \citet{clarke2003comparing}. 
Our additional simulation results show that in case of a large global heterogeneity, as $c$ increases, the heterogeneity across sites gets larger, reducing the influence of important covariates on heterogeneity, hence the weights become more non-adaptive. However, the proposed methods still enjoy a comparable performance to STACK, which further indicates the robustness of the proposed methods. 

\section{Example: a Multi-Hospital Data Network}
\label{sec:application}




Application with contextual insights is provided based on an analysis of the eICU Collaborative Research Database, a multi-hospital database published by Philips Healthcare \citep{pollard2018eicu}. 
The analysis is motivated by a recent retrospective study that there is a higher survival rate when SpO$_2$ is maintained at 94-98\% among patients requiring oxygen therapy \citep{van2020search}, not ``the higher the better''. 
We use the same data extraction code to create our data. 
We consider SpO$_2$ within this range as treatment ($Z=1$) and outside of this range as control ($Z=0$). 
A total of 7,022 patients from 20 hospitals, each with at least 50 patients in each treatment arm, are included with a randomly selected target (hospital 1).
Hospital-level summary information is provided in Appendix~\ref{suppl-sec:real}.
Patient-level features include age, BMI, sex, Sequential Organ Failure Assessment (SOFA) score, and duration of oxygen therapy.
The outcome is hospital survival ($Y=1$) or death ($Y=0$).




Figure~\ref{fig:real} visualizes the performance of EF-based estimated effect of oxygen therapy setting on in-hospital survival. 
CT is used as the local model with propensity score modeled by a logistic regression. 
Figure~\ref{fig:real}(a) shows the propensity score-weighted \textbf{average survival} for those whose received treatment is consistent with the estimated decision. 
Specifically, the expected reward is given by $$\frac{\sum_i Y_i 1(Z_i = Z^{est}_i) / \pi(Z_i,\bX_i)}{\sum_i 1(Z_i = Z^{est}_i) / \pi(Z_i,\bX_i)},$$ where $Z^{est}_i = 1(\widehat\tau>0)$ denotes the estimated treatment rule and $\pi(Z_i,\bX_i)$ is the probability of receiving the actual treatment.
We provide expected reward for the 1) observed treatment assignment (baseline), 2) LOC-based rule, and 3) EF-based rule. 
The treatment rule based on our method can increase mean survival by 3\% points compared to baseline, and is more promising than LOC. 

In the fitted EF, 
the hospital indicator is the most important, explaining about 50\% of the decrease in training error. 
Figure~\ref{fig:real}(b) shows the estimated CATE varying two important features, BMI and oxygen therapy duration. 
Patients with BMI around 36 and duration above 400 show the most benefit from oxygen therapy in the target SpO$_2$ range. 
Patients with BMI between 20 and 30 and duration around 200 may not benefit from such alteration. 
Figure~\ref{fig:real}(c) visualizes 
the 
data-adaptive weights $\omega_{k}(\bx)$ in the fitted EF with respect to BMI for different models, while holding other variables constant. The weights of hospital 1 are quite stable while models from other sites may have different contribution to the weighted estimator for different values of BMI. 
Judging from hospital information 
in Appendix~\ref{suppl-sec:real}, hospitals with a larger bed capacity tend to be similar to hospital 1, and are shown to provide larger contributions.



In this distributed research network, different hospitals have a different sample size. 
For sensitivity analysis, we consider a weighting strategy to adjust for the sample size of site $k$. 
Results show similar patterns as in Figure \ref{fig:real}. Detailed results 
are provided in Appendix~\ref{suppl-sec:real}. 
The real-data access is provided in Appendix~\ref{suppl-sec:code}.

\section{Discussion} 
\label{sec:disc}


We have proposed an efficient and interpretable tree-based model averaging framework for enhancing treatment effect estimation at a target site by borrowing information from potentially heterogeneous data sources. We generalize standard model averaging scheme in a data-adaptive way such that the generated weights depend on subject-level features. 
This work makes multi-site collaborations and especially treatment effect estimation more practical by avoiding the need to share subject-level data. 
Our approach extends beyond causal inference to estimating a general $f(\bx)$ from heterogeneous data. 


Unlike in classic model averaging where prediction performance can be assessed against observed outcomes or labels, treatment effects are not directly observed. 
While our approach is guaranteed to be consistent under randomized studies, 
the weights are estimated based on expected treatment effects, hence relying on  Assumption 1 (unconfoundedness) to hold. It may be a strong assumption in observational studies with unmeasured confounding. 

\section*{Acknowledgments}
This research was supported in part by the Competitive Medical Research Fund of the UPMC Health System and the University of Pittsburgh Center for Research Computing through the resources provided. 
The authors thank Gong Tang and Yu Wang for insightful discussion.


\bibliography{reference}
\bibliographystyle{icml2022}

\newpage
\appendix
\onecolumn


\section{Related Topics and Distinctions}
\label{suppl-related}

In Section~\ref{sec:related}, we focused on the literature review of model averaging for ease of exposition, because the most innovated part of our method is motivated directly from this class of work. 
Here we clarify the main differences among model averaging, meta-analysis, federated learning, as well as super learner. 

Model averaging: a convex averaging of models via model-specific weights \citep{raftery1997bayesian,yang2001adaptive, dai2011greedy,yao2018using,dai2018bayesian}. The extension of the weights from scalars to functions provides the best motivation for our approach. 

Meta-analysis: classic in the way that it describes the site-level heterogeneity using modeling assumptions \citep{whitehead2002meta,sutton2000methods}, rather than a more data-driven approach such as tree models. It can be either frequentist or Bayesian, the latter of which tends to be more useful under limited sample sizes. However, its main interest is typically the overall effect rather than the site-level heterogeneity, which is usually modeled by a nuisance parameter \citep{borenstein2011introduction,riley2011interpretation,rover2020dynamically}. 

Federated learning: originated from the field of computer science \citep{mcmahan2017communication}, federated learning is a collaborative learning procedure that ensures data privacy by exchanging model parameters only. Federated learning methods often involves iterative updating \citep{fallah2020personalized,cho2021personalized,smith2017federated,yang2019federated}, rather than a one-shot procedure, which could be hard to apply to nonautomated distributed research networks.  
It has been developed mainly to estimate a global prediction model by leveraging distributed data \citep{li2020federated,kairouz2019advances,zhao2018federated,hard2018federated}, 
and is not designed to target any specific site. 

Super learner: an ensemble of multiple statistical and machine learning models \citep{van2007super}. It learns an optimal weighted average of those candidate models by minimizing the cross-validated risk, and assigns higher weights to more accurate models \citep{polley2010super}.  The final prediction on an independent testing data is the weighted combination of the predictions of those models. 
Super learner has been showed empirically to improve treatment effect estimation via the modeling of propensity score in observational studies \citep{pirracchio2015improving,wyss2018using,shortreed2019challenges,ju2019propensity,tan2022doubly}. 

Mixture of experts: an ensemble learning technique that decomposes a task into multiple subtasks with domain knowledge, followed by using multiple expert models to handle each subtask. A gating model is then used to decide which expert to use to make future prediction \citep{masoudnia2014mixture}. It differs from other ensemble methods typically in that often only a few experts will be selected for predictions, rather than combining results from all experts \citep{masoudnia2014mixture}. 



\section{Proof of Theorem~\ref{Theorem}} 
\label{suppl-sec:consist}

The proof of Theorem~\ref{Theorem} closely follows arguments given in \citet{wager2018estimation}. Suppose the subsamples for building each tree in an ensemble forest are drawn from different subjects in the augmented site 1 data. Specifically, in one round of EF, we draw $m$ samples from the augmented data, where $m$ is less than the rows in the augmented data, i.e., $m < (n_1 \cdot K)$. By randomly picking $m$ unique subjects from site 1 and then randomly picking a site indicator $k$ out of $K$ sites for each of the $m$ subjects. The resulted $m$ subsamples should not be from the same subject and are hence independent and identically distributed. 
As long as $m < n_1$, we can ensure that all the subsamples are independent. 
In practice, when the ratio of $n_1 / K$ is relatively large, the probability of obtaining samples from the same subject is small. 

Assume that subject features $\bX_i$ and the site indicator $S_i$ are independent and have a density that is bounded away from 0 and infinity. Suppose moreover that the conditional mean function $\mathbb{E}[ \Tau|\bX=\bx,S=k]$ is Lipschitz continuous. We adopt the honesty definition in \citet{athey2016recursive} when building trees in a random forest. Honest approaches separate the training sample into two halves, one half for building the tree model, and another half for estimating treatment effects within the leaves \citep{athey2016recursive}. Following Definitions 1-5 and Theorem 3.1 in \citet{wager2018estimation}, the proposed estimator $\widehat \Tau_{\text{EF}}(\bx, 1)$ is a consistent estimator of the true treatment effect function $\tau_1(\bx)$ for site 1.

\section{Full Simulation Results}
\label{suppl-sec:sim}

\begin{figure}[!h]
\centering
 \begin{subfigure}{0.49\textwidth}
  \centerline{\includegraphics[width=\linewidth]{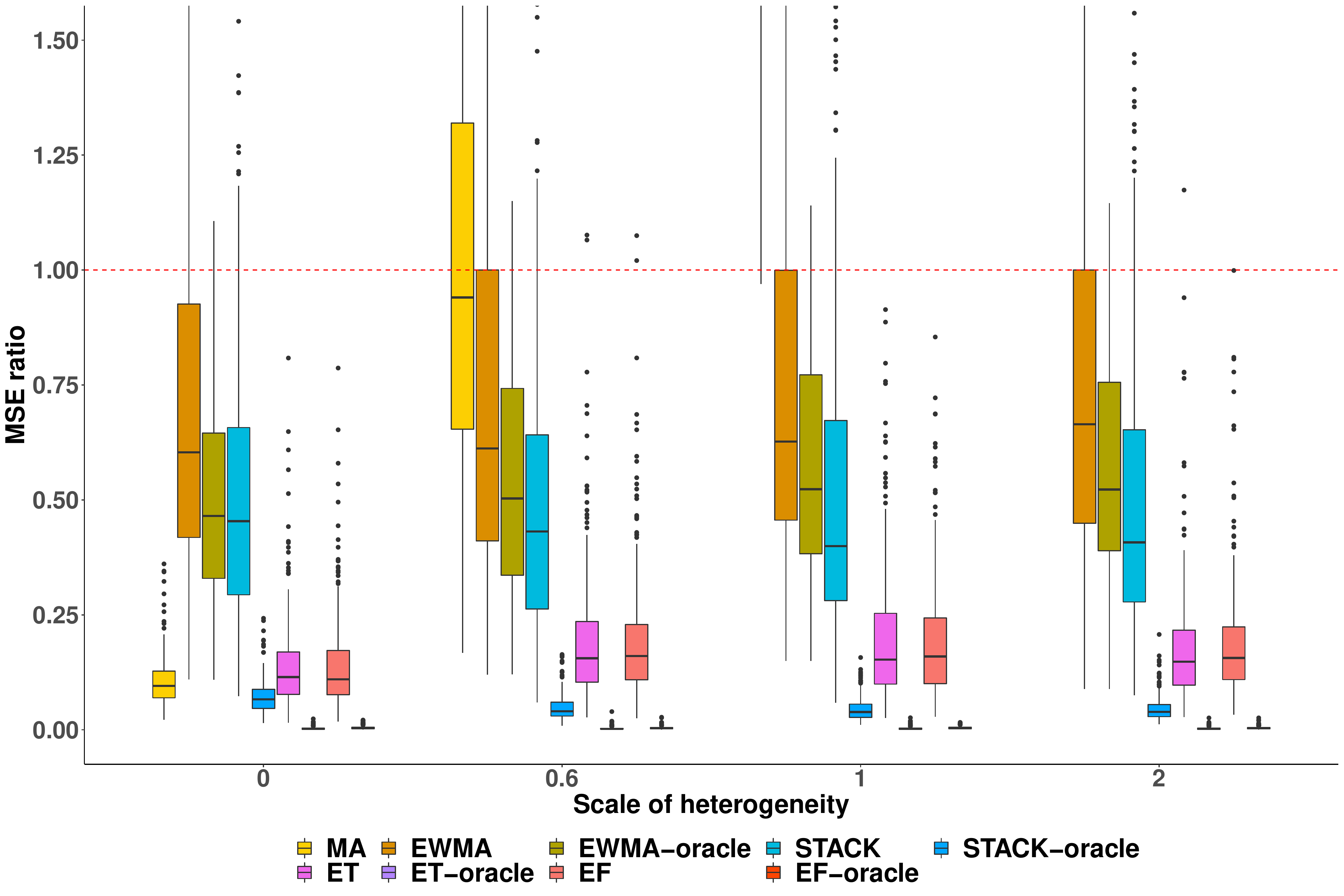}}
  \caption{}
 \end{subfigure}
 \begin{subfigure}{0.49\textwidth}
  \centerline{\includegraphics[width=\linewidth]{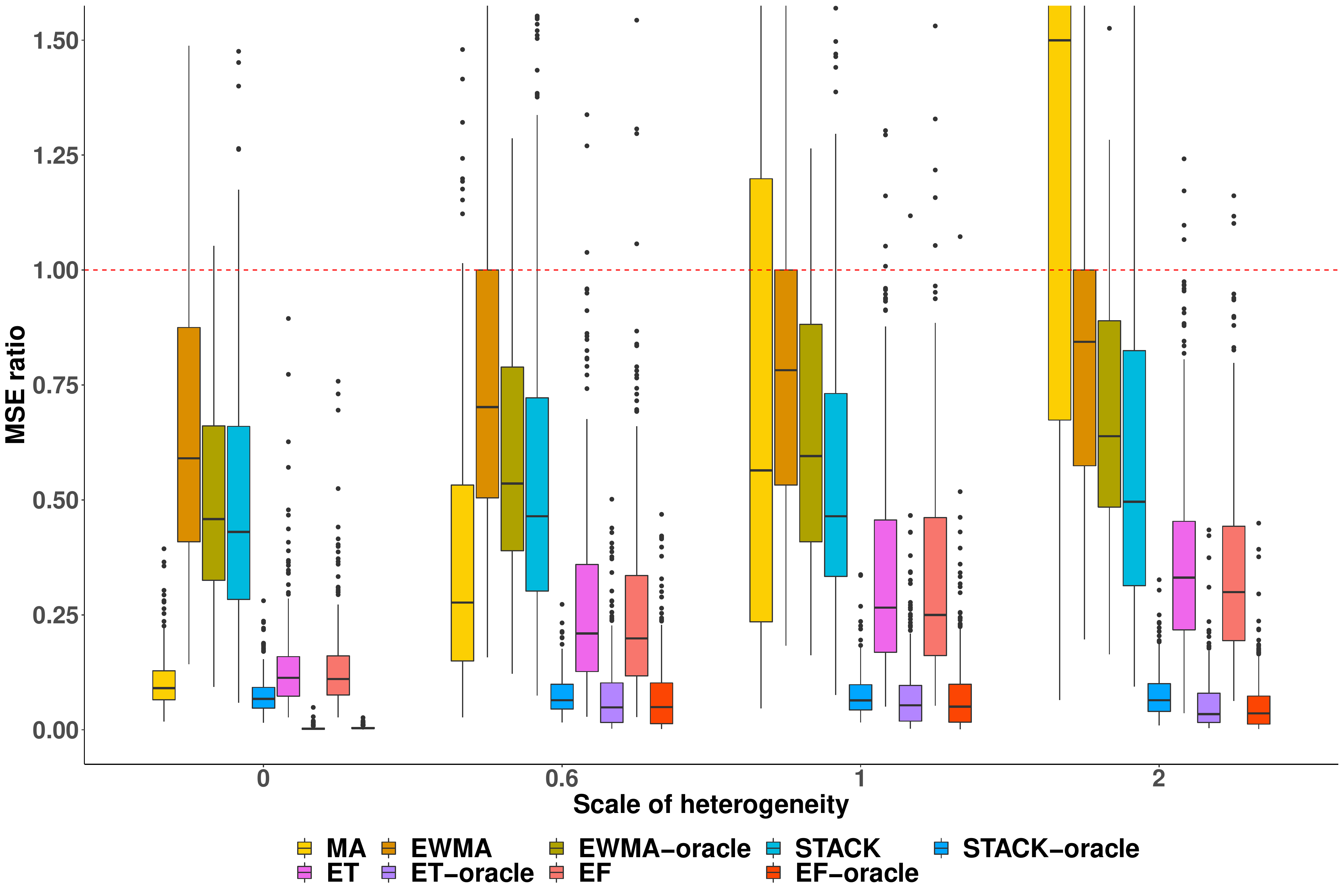}}
  \caption{}
 \end{subfigure}
 \caption{
Box plots of the MSE ratios of CATE estimators, respectively, over LOC (\textbf{CT}) and a sample size of \textbf{500} at each site for \textbf{(a) discrete grouping} and \textbf{(b) continuous grouping} across site, respectively, varying scale of global heterogeneity. 
Estimators ending with ``-oracle" makes use of ground truth treatment effects. 
Different colors imply different estimators, and x-axis, i.e., the value of $c$, differentiates the scale of global heterogeneity. The red dotted line denotes an MSE ratio of 1. 
MA performance is truncated due to large MSE ratios. 
The proposed ET and EF achieve competitive performance compared to standard model averaging or ensemble methods and are robust to heterogeneity across settings. 
Note that ET-oracle and EF-oracle achieve close-to-zero MSE ratios with very small spreads in some settings. 
}
 \label{web:sim_fig_ct500_full}
\end{figure}

\textbf{Connection to supervised learning.} 
Similar to ET-oracle and EF-oracle whose weights are built on the ground truth CATE functions $\tau_k$'s, we also consider for EWMA and STACK under a similar hypothetical setting.
Specifically, we assume the true $\tau_1$ is known and use it to compute the weights. 
This version of EWMA estimator is denoted as EWMA-oracle and its weight is given by $${\omega}_{k}^{\text{EWMA-oracle}} =\frac{\exp\{- \sum_{i \in \mathcal{I}_1^{(2)}}(\widehat\tau_k(\bx_i) - \tau_1(\bx_i))^2\} }{ \sum_{\ell=1}^{K} \exp\{- \sum_{i \in \mathcal{I}_1^{(2)}}(\widehat\tau_\ell(\bx_i) - \tau_1(\bx_i))^2\} }.$$
Similarly, the corresponding linear stacking approach, denoted as STACK-oracle, regresses the ground truth $\tau_1(\bx)$ on the predictions of the estimation set in site 1 from each local model, $\{\widehat\tau_1(\bx), \dots, \widehat\tau_k(\bx) \}$. 
We compare the proposed model averaging estimators with the local estimator, MA, two versions of modified EWMA, as well as two versions of the linear stacking approach. 
We present simulation results using CT as the local model and the sample size at local sites to be $n=500$. 
Figure~\ref{web:sim_fig_ct500_full} presents the performance of the proposed estimators along with other competing estimators. Each series of boxes corresponds to a different strength of global heterogeneity $c$. 
Table~\ref{tab:sim_res} reports the ratio between MSE of the estimator and MSE of the local model in terms of average and standard deviation of MSE, respectively, over 1000 replicates.
Our proposed estimators ET and EF shows the best performance overall in terms of the mean and variation of MSE among the estimators without using the information of ground truth $\tau_1(\bx)$. Comparing with ET, EF has a slightly smaller MSE when $c$ is large, which is expected because forest models tend to be more stable and accurate than a single tree. 
ET-oracle achieves minimal MSE for low and moderate degrees of heterogeneity while EF-oracle has the minimal MSE under all settings.
The local estimator (LOC) in general shows the largest MSE compared to other estimators, as it does not leverage information from other sites. By borrowing information from additional sites, variances are greatly reduced, resulting in a small MSE of ensemble estimators. 
MA that naively adopts the inverse of sample size as weights performs well under low levels of heterogeneity, but suffers from a huge MSE with large variation as $c$ increases. 
EWMA estimators perform slightly better and are more stable than LOC and MA. EWMA-oracle has better performance than EWMA in all settings as the information of true CATE is used for weight construction. STACK estimators performs better than EWMA estimators. 
Similarly, STACK-oracle performs better than STACK in all settings. STACK-oracle, with ground truth $\tau_1(\bx)$ available, outperforms ET and EF when there exists a moderate to high level of heterogeneity across sites.  

\begin{figure}[!h]
\centering
 \begin{subfigure}{0.49\textwidth}
  \centerline{\includegraphics[width=\linewidth]{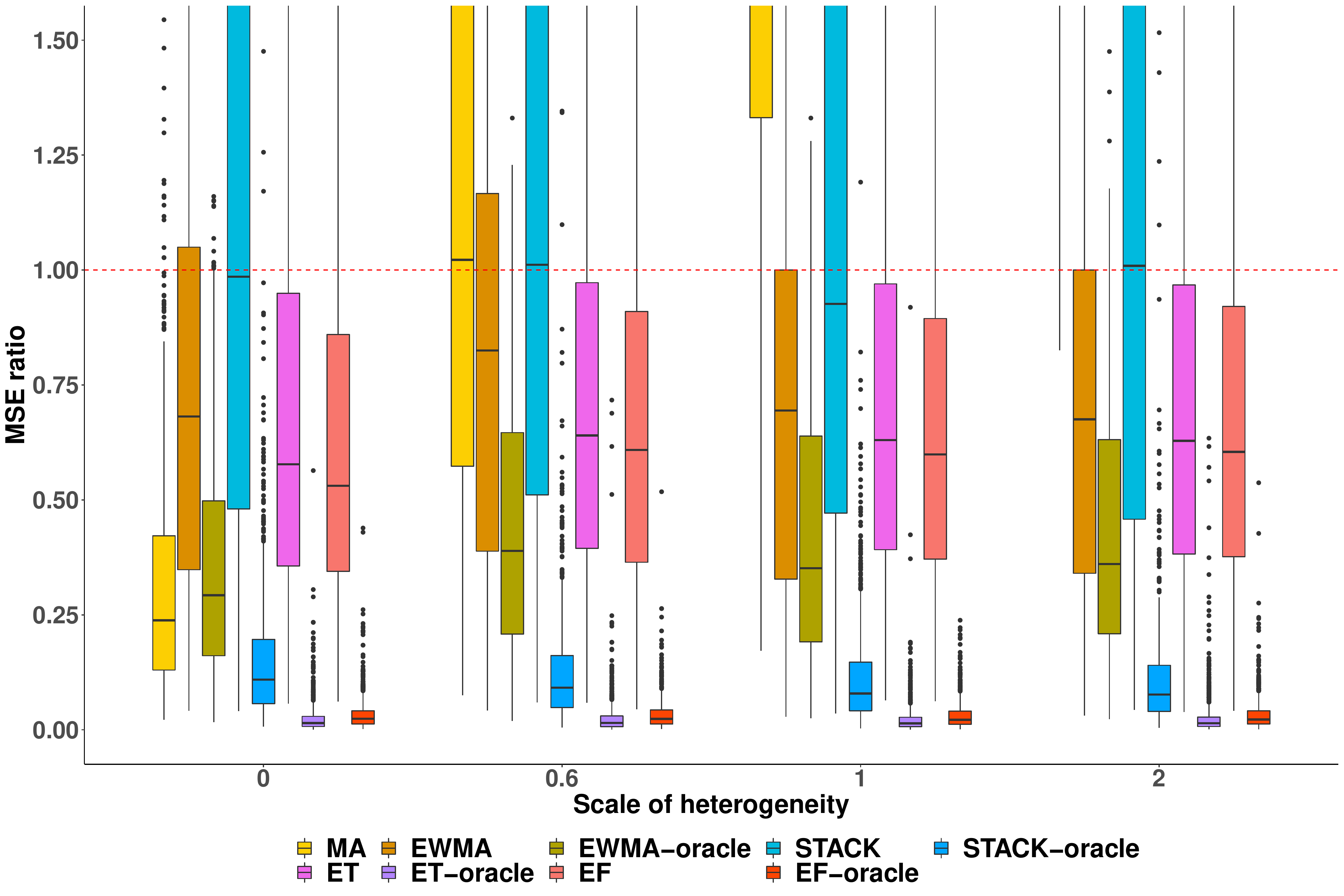}}
  \caption{}
 \end{subfigure}
 \begin{subfigure}{0.49\textwidth}
  \centerline{\includegraphics[width=\linewidth]{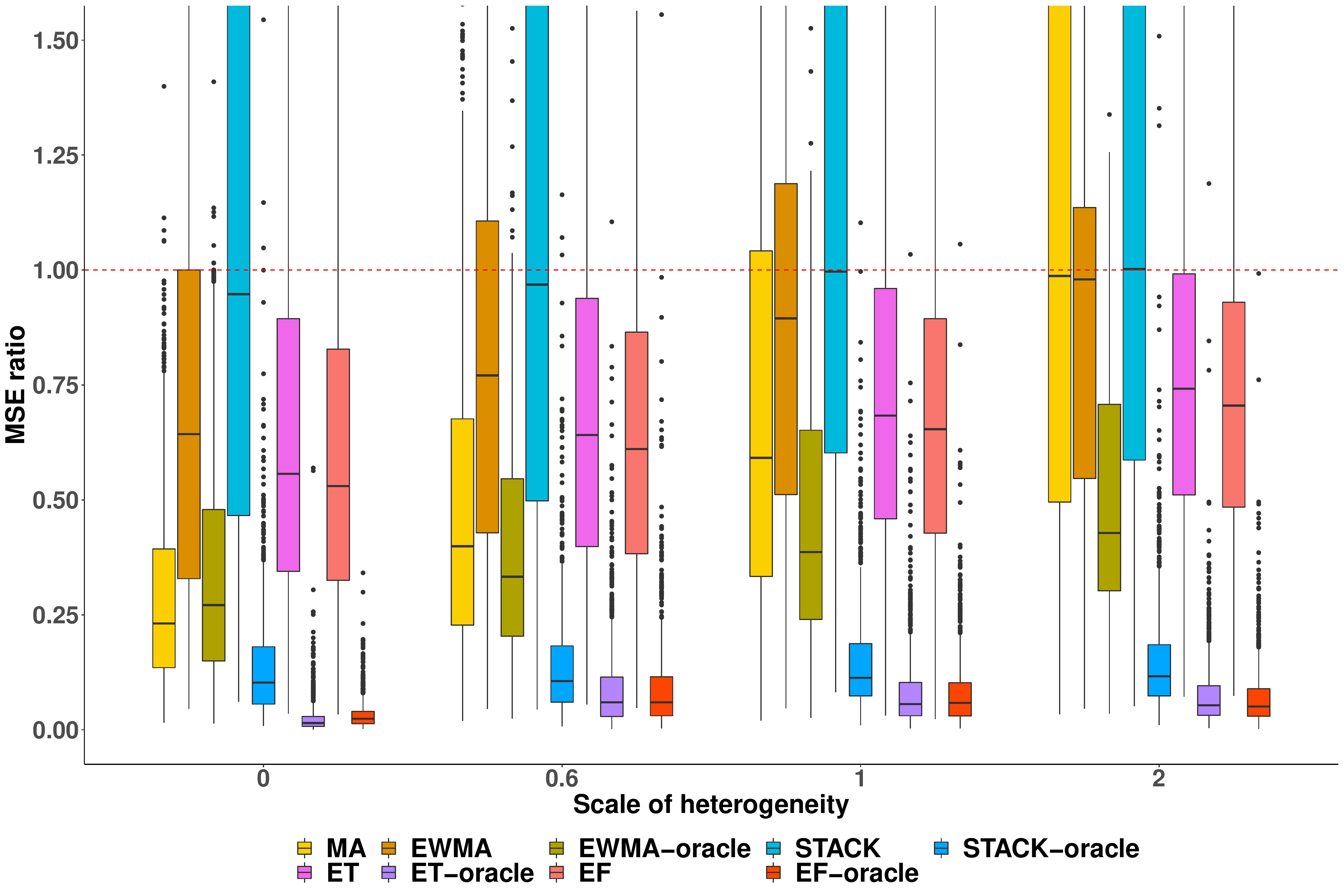}}
  \caption{}
 \end{subfigure}
 \caption{
Box plots of the MSE ratios of CATE estimators, respectively, over LOC (\textbf{CT}) and a sample size of \textbf{100} at each site for \textbf{(a) discrete grouping} and \textbf{(b) continuous grouping} across site, respectively, varying scale of global heterogeneity. 
Estimators ending with ``-oracle" makes use of ground truth treatment effects. 
Different colors imply different estimators, and x-axis, i.e., the value of $c$, differentiates the scale of global heterogeneity. The red dotted line denotes an MSE ratio of 1. 
MA performance is truncated due to large MSE ratios. 
The proposed ET and EF achieve competitive performance compared to standard model averaging or ensemble methods and are robust to heterogeneity across settings. 
Note that ET-oracle and EF-oracle achieve close-to-zero MSE ratios with very small spreads in some settings. 
 }
 \label{web:sim_fig_ct100}
\end{figure}

\begin{figure}[!h]
\centering
 \begin{subfigure}{0.49\textwidth}
  \centerline{\includegraphics[width=\linewidth]{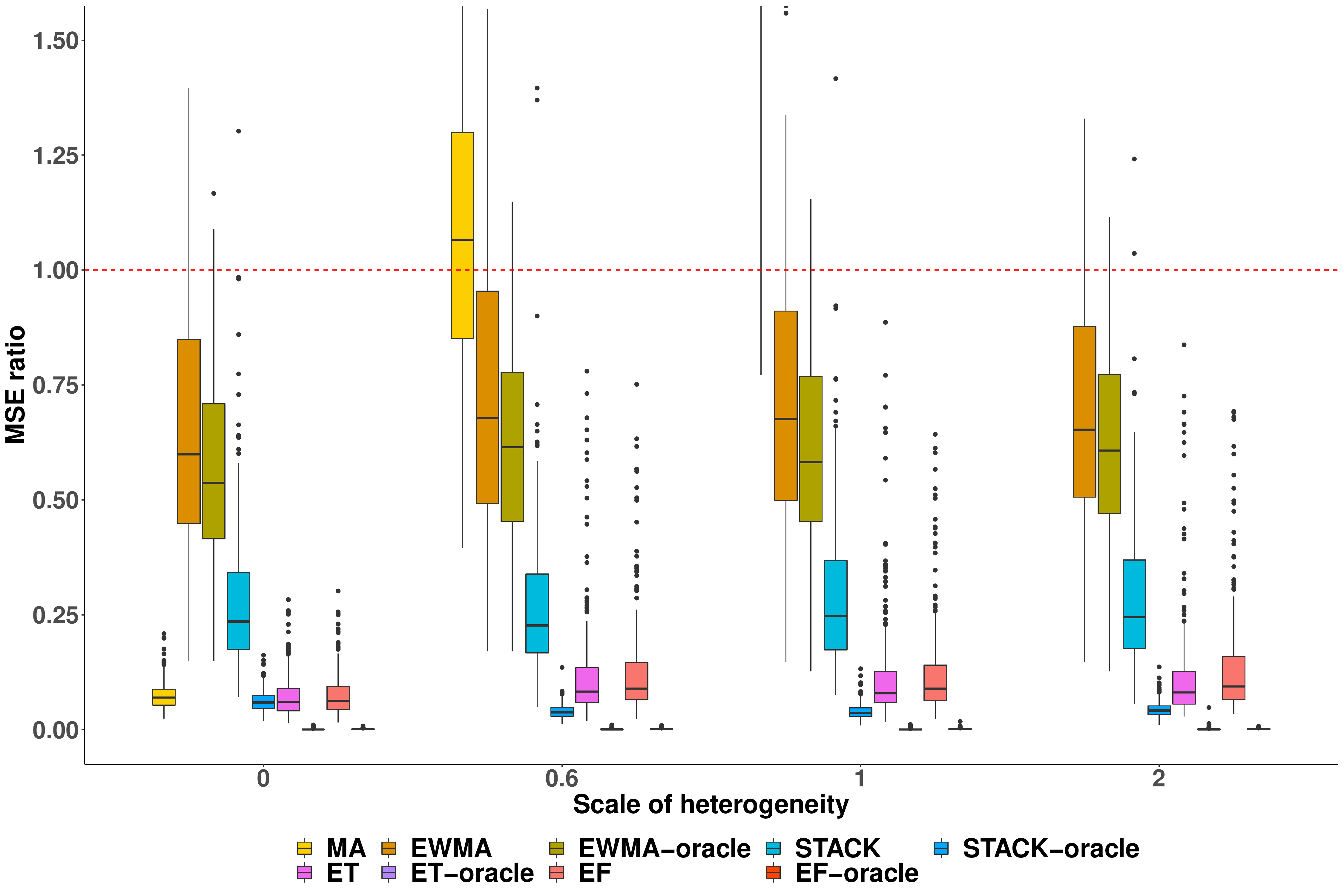}}
  \caption{}
 \end{subfigure}
 \begin{subfigure}{0.49\textwidth}
  \centerline{\includegraphics[width=\linewidth]{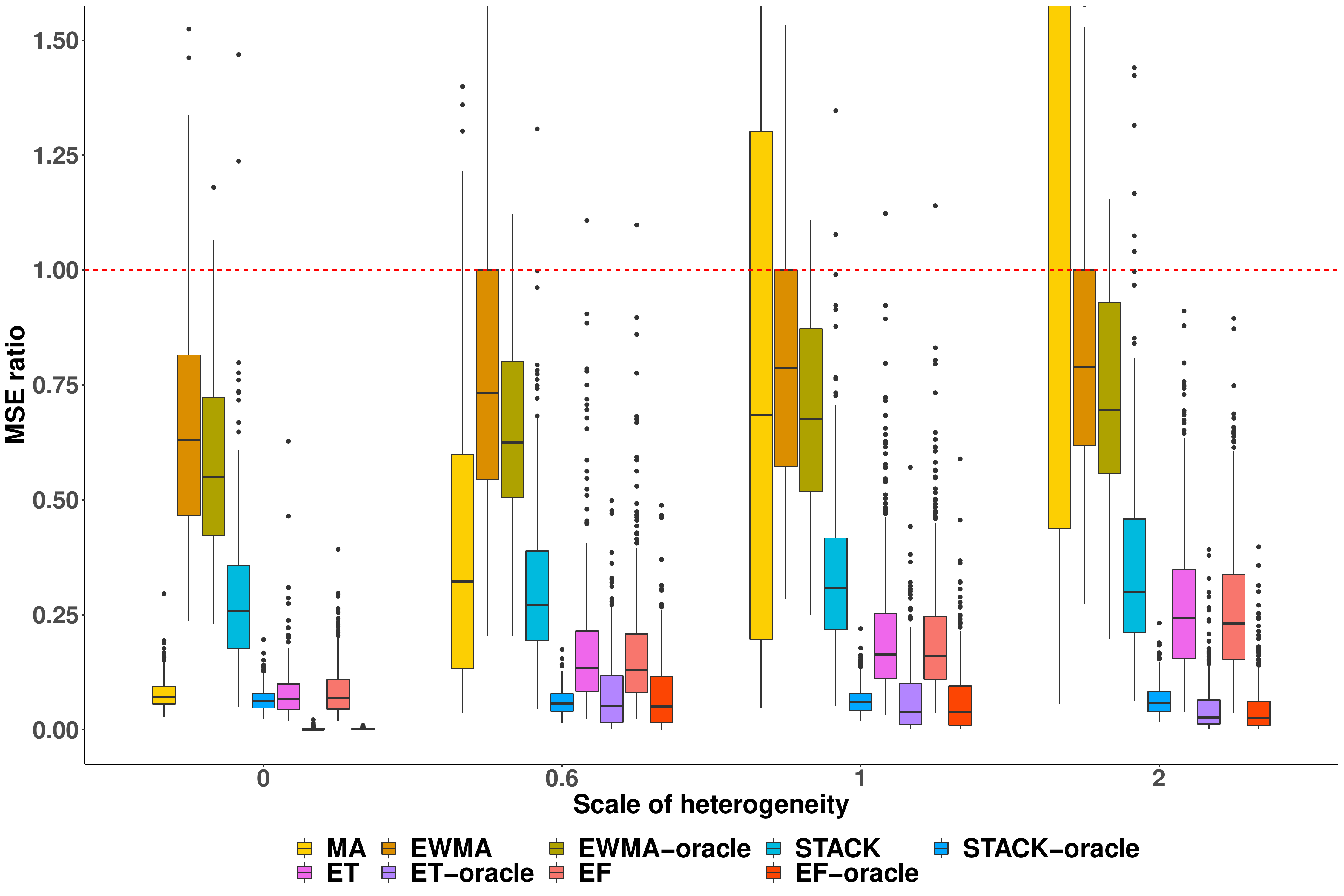}}
  \caption{}
 \end{subfigure}
 \caption{
Box plots of the MSE ratios of CATE estimators, respectively, over LOC (\textbf{CT}) and a sample size of \textbf{1000} at each site for \textbf{(a) discrete grouping} and \textbf{(b) continuous grouping} across site, respectively, varying scale of global heterogeneity. 
Estimators ending with ``-oracle" makes use of ground truth treatment effects. 
Different colors imply different estimators, and x-axis, i.e., the value of $c$, differentiates the scale of global heterogeneity. The red dotted line denotes an MSE ratio of 1. 
MA performance is truncated due to large MSE ratios. 
The proposed ET and EF achieve competitive performance compared to standard model averaging or ensemble methods and are robust to heterogeneity across settings. 
Note that ET-oracle and EF-oracle achieve close-to-zero MSE ratios with very small spreads in some settings. 
 }
 \label{web:sim_fig_ct1000}
\end{figure}

\begin{figure}[h]
\centering
 \begin{subfigure}{0.46\textwidth}
  \centerline{\includegraphics[width=0.9\linewidth]{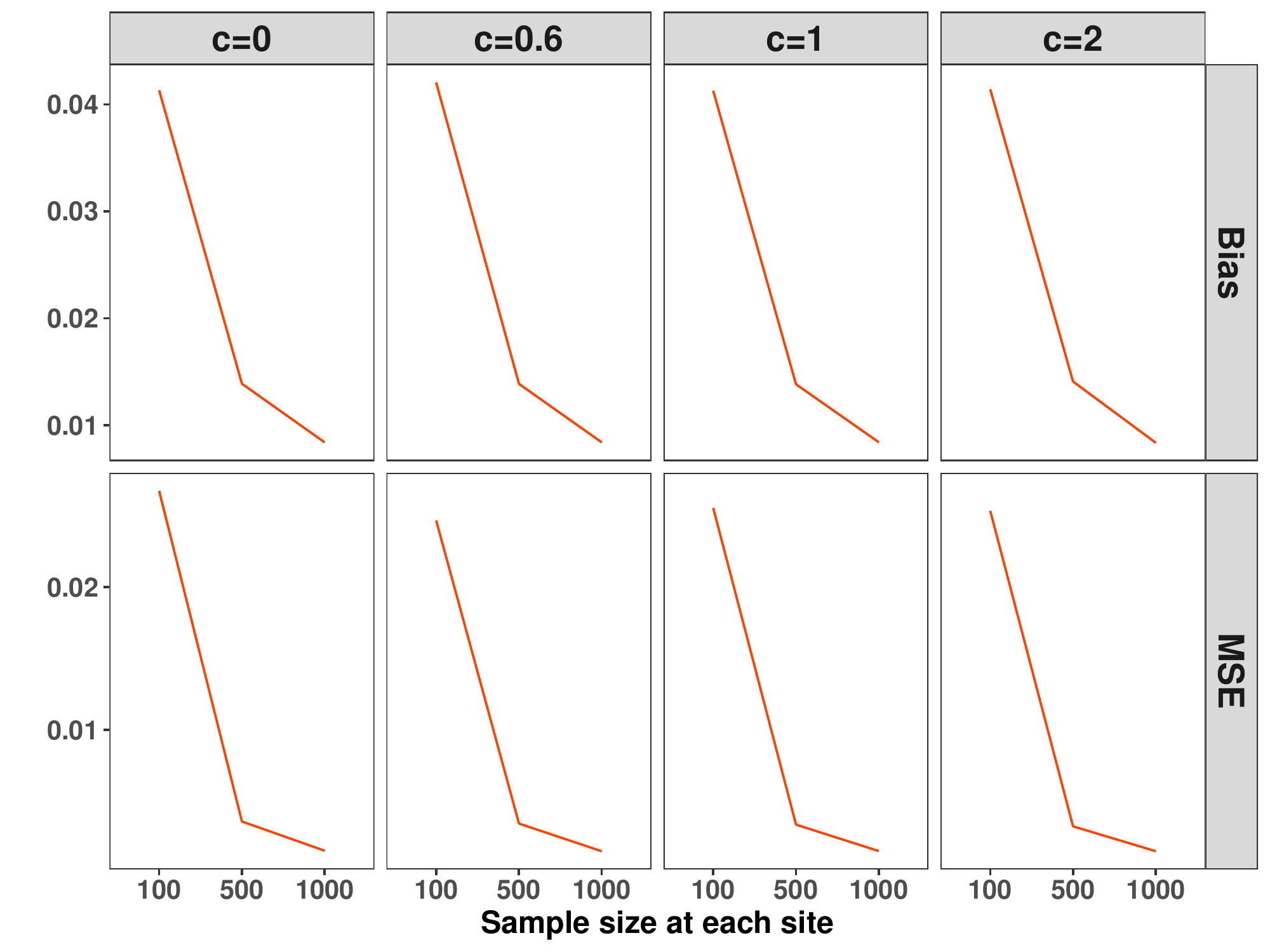}}
  \caption{}
 \end{subfigure}
 \begin{subfigure}{0.46\textwidth}
  \centerline{\includegraphics[width=0.9\linewidth]{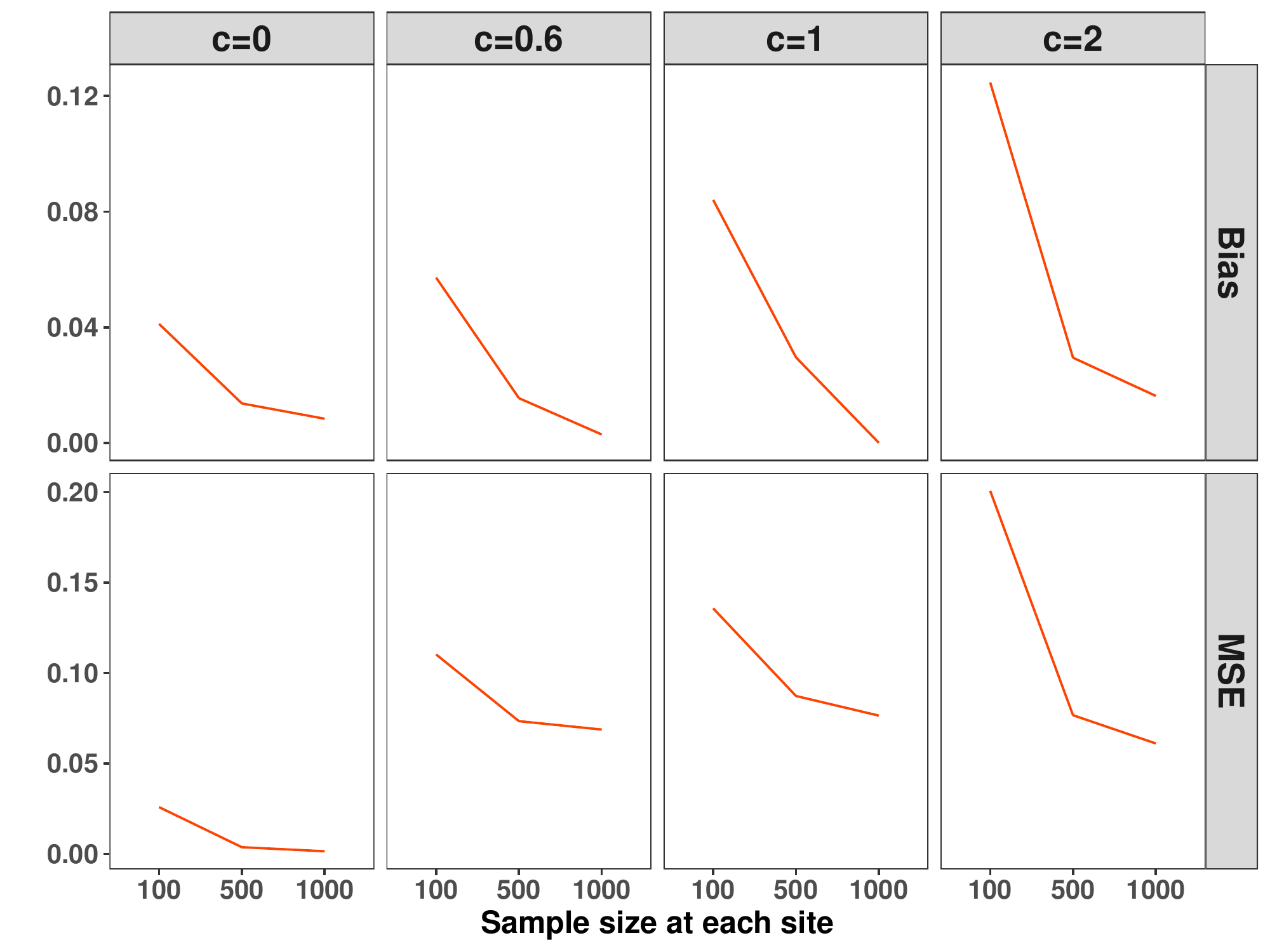}}
  \caption{}
 \end{subfigure}
 \caption{Plots of the bias and MSE of \textbf{EF-oracle} varying sample site at each site for \textbf{(a) discrete grouping} and \textbf{(b) continuous grouping} across site, varying scale of global heterogeneity. 
 Both bias and MSE reduces to zero as the sample size increases. 
 }
 \label{fig:sim_vary_n}
\end{figure}

\textbf{Various sample sizes in local sites.}
We provide detailed simulation results varying $n$ $(100, 500, 1000)$ with CT as the local model. 
Figure~\ref{web:sim_fig_ct100} and Figure~\ref{web:sim_fig_ct1000} show box plots of simulation results with a sample size of 100 and 1000, respectively, at each site. Our proposed methods ET and EF show robust performance in all settings. ET-oracle and EF-oracle achieve close-to-zero MSE with very small spreads in some settings.  Figure~\ref{fig:sim_vary_n} shows plots of the bias and MSE of EF-oracle varying sample size at each site ($n = 100, 500, 1000$). As the sample size increases, both bias and MSE of EF-oracle reduce to zero. Consistency of EF-oracle can be shown via simulation when perfect estimates are obtained from local models. 
Meanwhile, our proposed method greatly reduce MSE by selectively borrowing information from multiple sites. 

\begin{figure}[!h]
\centering
 \begin{subfigure}{0.49\textwidth}
  \centerline{\includegraphics[width=\linewidth]{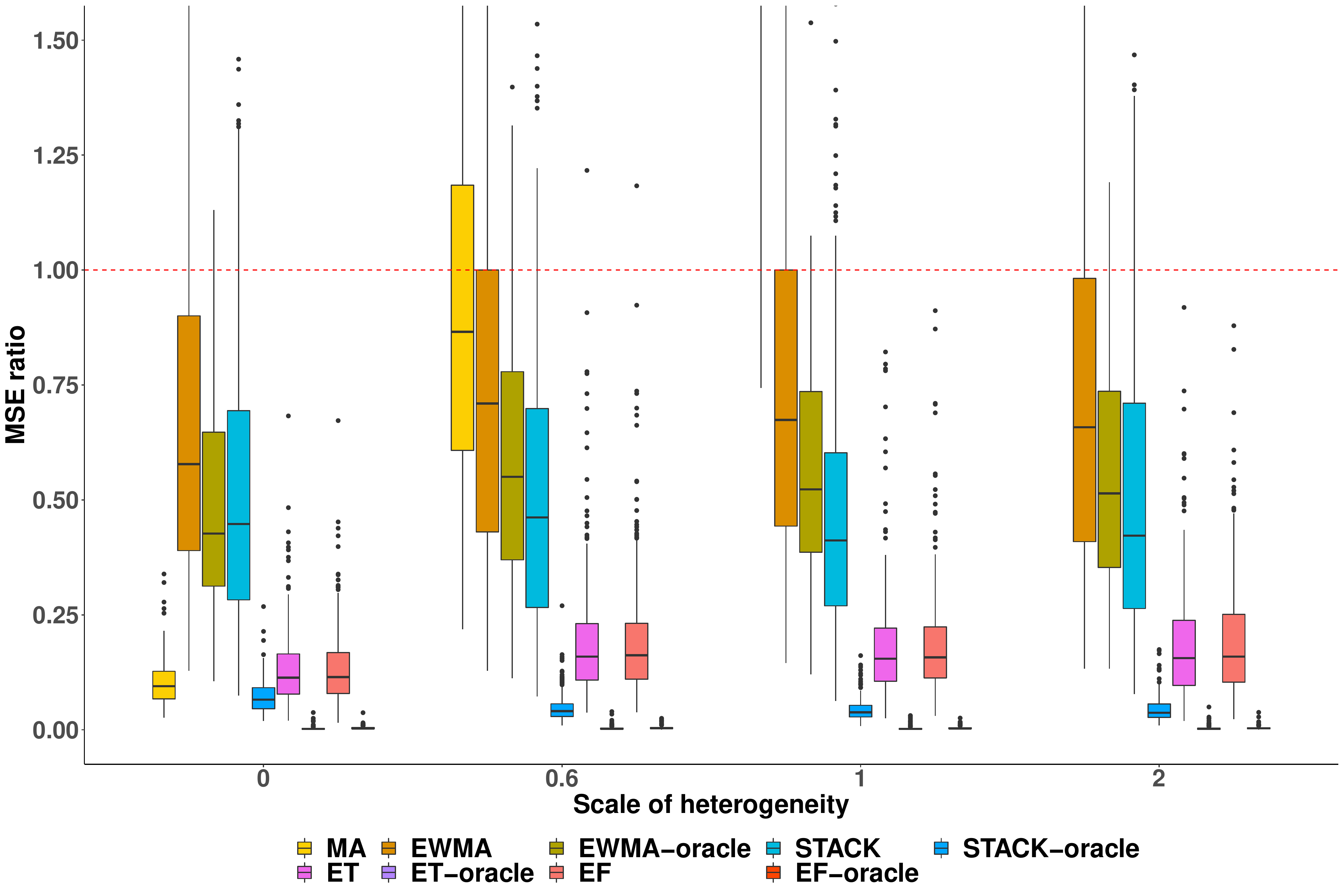}}
  \caption{}
 \end{subfigure}
 \begin{subfigure}{0.49\textwidth}
  \centerline{\includegraphics[width=\linewidth]{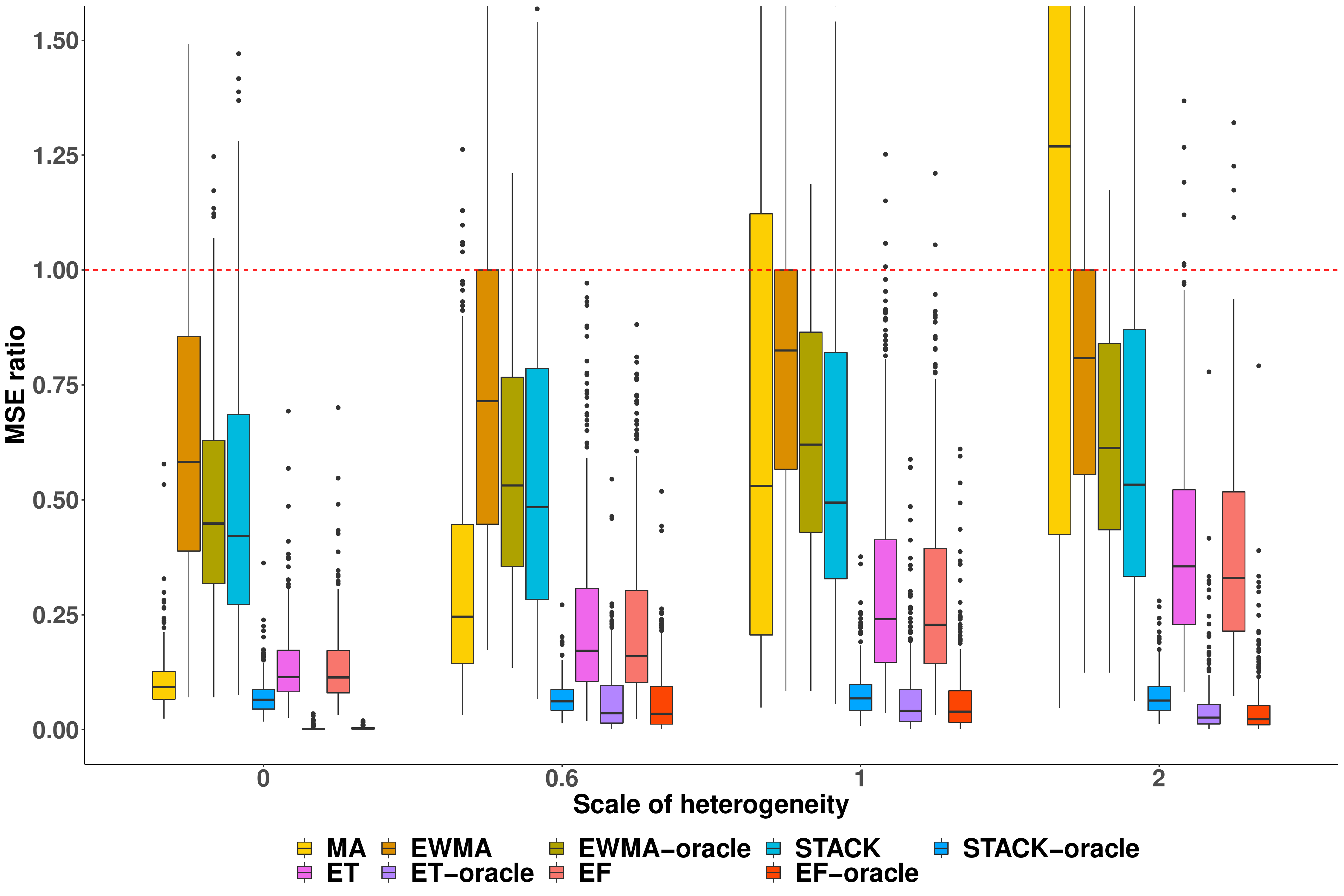}}
  \caption{}
 \end{subfigure}
 \caption{
Box plots of the MSE ratios of CATE estimators, respectively, over LOC (\textbf{CT}) and a sample size of \textbf{500} at each site under \textbf{observational design with a correctly specified propensity score model} for \textbf{(a) discrete grouping} and \textbf{(b) continuous grouping} across site, respectively, varying scale of global heterogeneity. 
Estimators ending with ``-oracle" makes use of ground truth treatment effects. 
Different colors imply different estimators, and x-axis, i.e., the value of $c$, differentiates the scale of global heterogeneity. The red dotted line denotes an MSE ratio of 1. 
MA performance is truncated due to large MSE ratios. 
The proposed ET and EF achieve competitive performance compared to standard model averaging or ensemble methods and are robust to heterogeneity across settings. 
Note that ET-oracle and EF-oracle achieve close-to-zero MSE ratios with very small spreads in some settings. 
 }
 \label{web:sim_obs_correct}
\end{figure}

\begin{figure}[!h]
\centering
 \begin{subfigure}{0.49\textwidth}
  \centerline{\includegraphics[width=\linewidth]{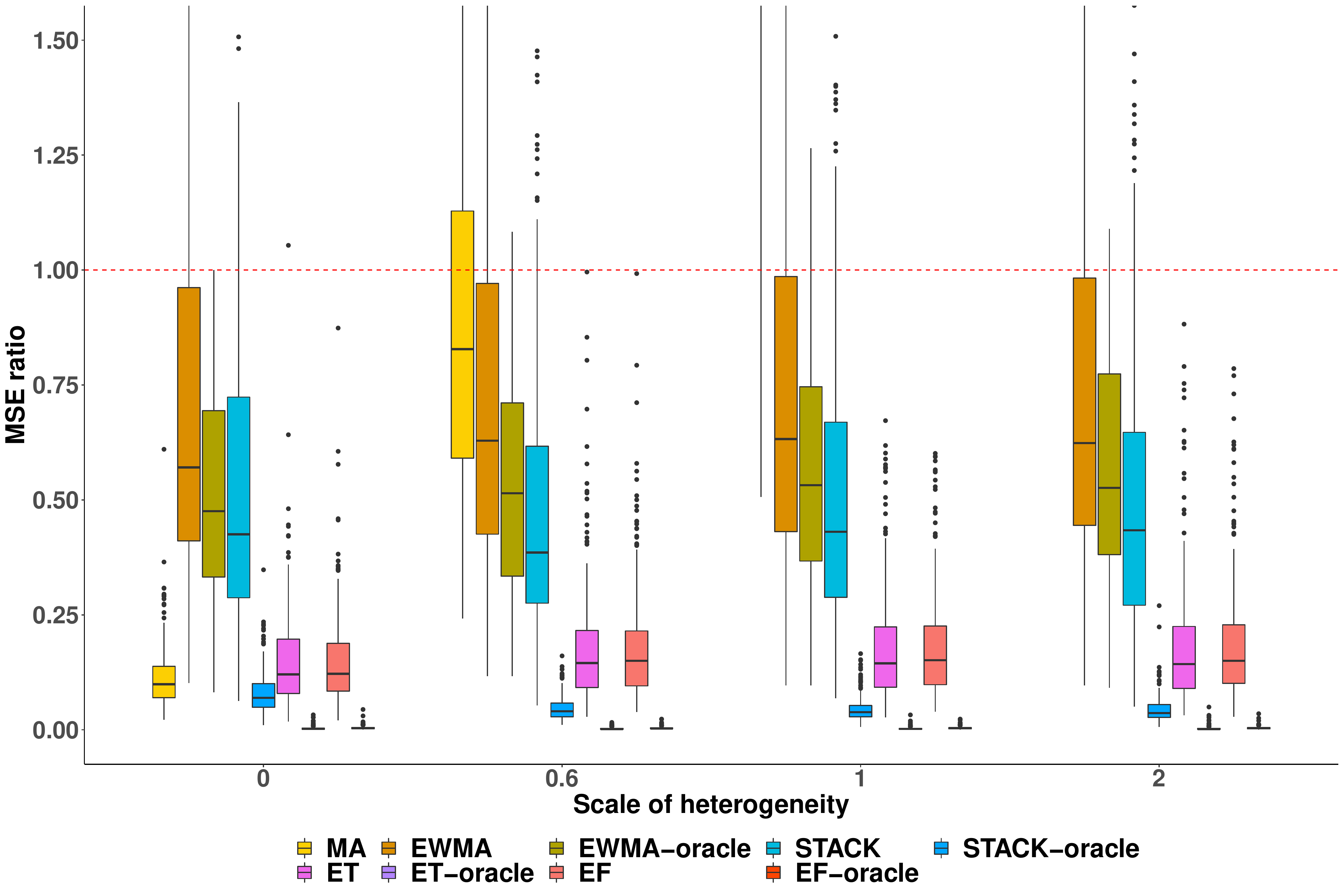}}
  \caption{}
 \end{subfigure}
 \begin{subfigure}{0.49\textwidth}
  \centerline{\includegraphics[width=\linewidth]{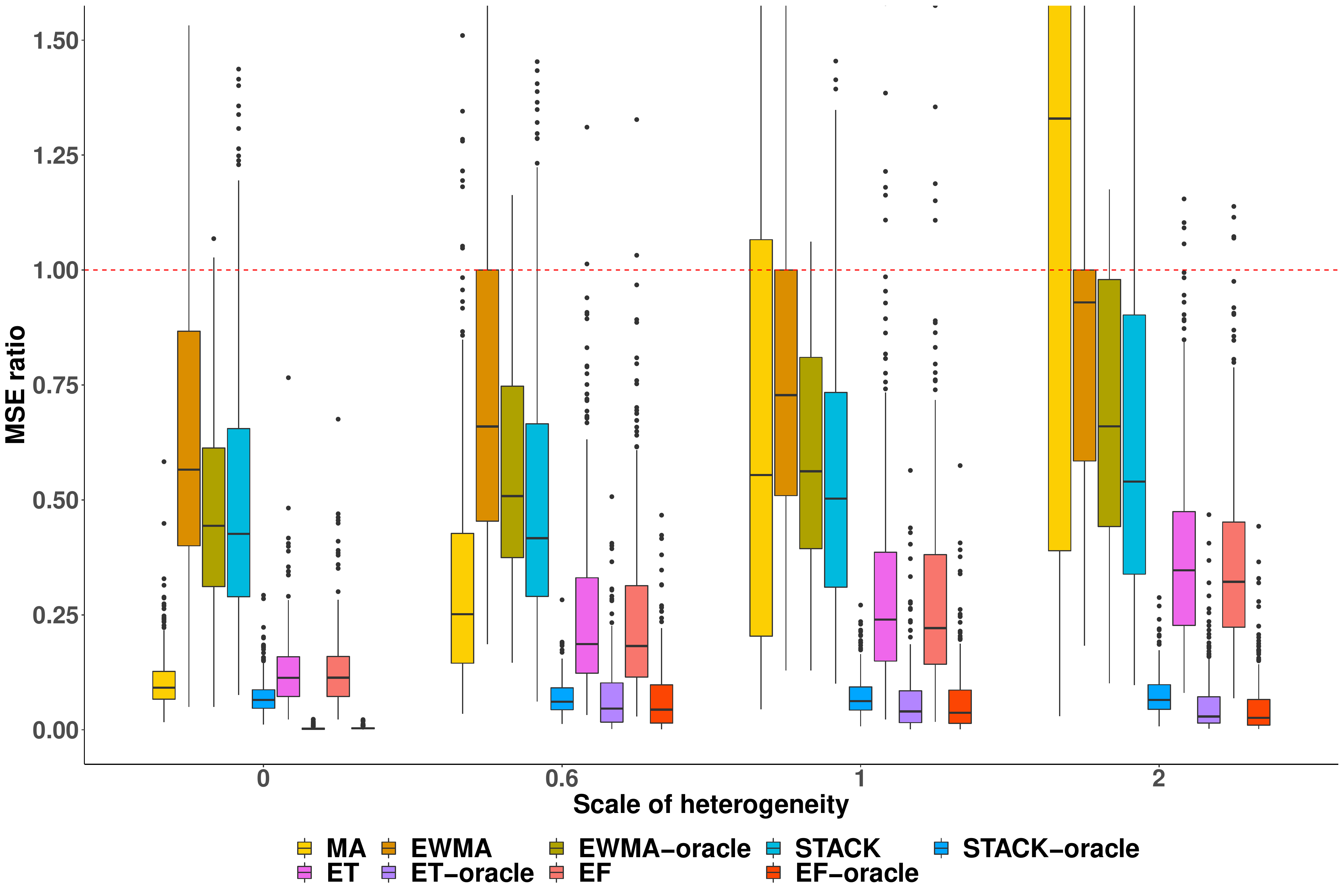}}
  \caption{}
 \end{subfigure}
 \caption{
Box plots of the MSE ratios of CATE estimators, respectively, over LOC (\textbf{CT}) and a sample size of \textbf{500} at each site under \textbf{observational design with a misspecified propensity score model} for \textbf{(a) discrete grouping} and \textbf{(b) continuous grouping} across site, respectively, varying scale of global heterogeneity. 
Estimators ending with ``-oracle" makes use of ground truth treatment effects. 
Different colors imply different estimators, and x-axis, i.e., the value of $c$, differentiates the scale of global heterogeneity. The red dotted line denotes an MSE ratio of 1. 
MA performance is truncated due to large MSE ratios. 
The proposed ET and EF achieve competitive performance compared to standard model averaging or ensemble methods and are robust to heterogeneity across settings. 
Note that ET-oracle and EF-oracle achieve close-to-zero MSE ratios with very small spreads in some settings. 
 }
 \label{web:sim_obs_misspecified}
\end{figure}

\textbf{Simulations under observational studies.}
We also consider the treatment generation mechanism under an observational design. 
Specifically, the propensity is given as $e(\bx) = \text{expit}(0.6x_1)$. We consider both a correctly specified propensity model using a logistic regression of $Z$ on $X_1$ and a misspecified propensity model with a logistic regression of $Z$ on all $\bX$. 
Figure~\ref{web:sim_obs_correct} and Figure~\ref{web:sim_obs_misspecified} show box plots of simulation results. 
In general, the proposed estimators obtain the best performance with similar results are obtained as in the Figure~\ref{fig:sim_box}. With the correctly specified propensity score model, the local estimator is consistent in estimating $\tau_k(\bx)$, the proposed framework is valid. When the propensity model misspecified, extra uncertainty is carried forward from the local estimates, but the proposed estimators can improve upon the local models. This is due to a bias-and-variance trade-off that guarantees small MSE in prediction, which remains smaller than those from local estimators. 

\begin{figure}[!h]
\centering
 \begin{subfigure}{0.49\textwidth}
  \centerline{\includegraphics[width=\linewidth]{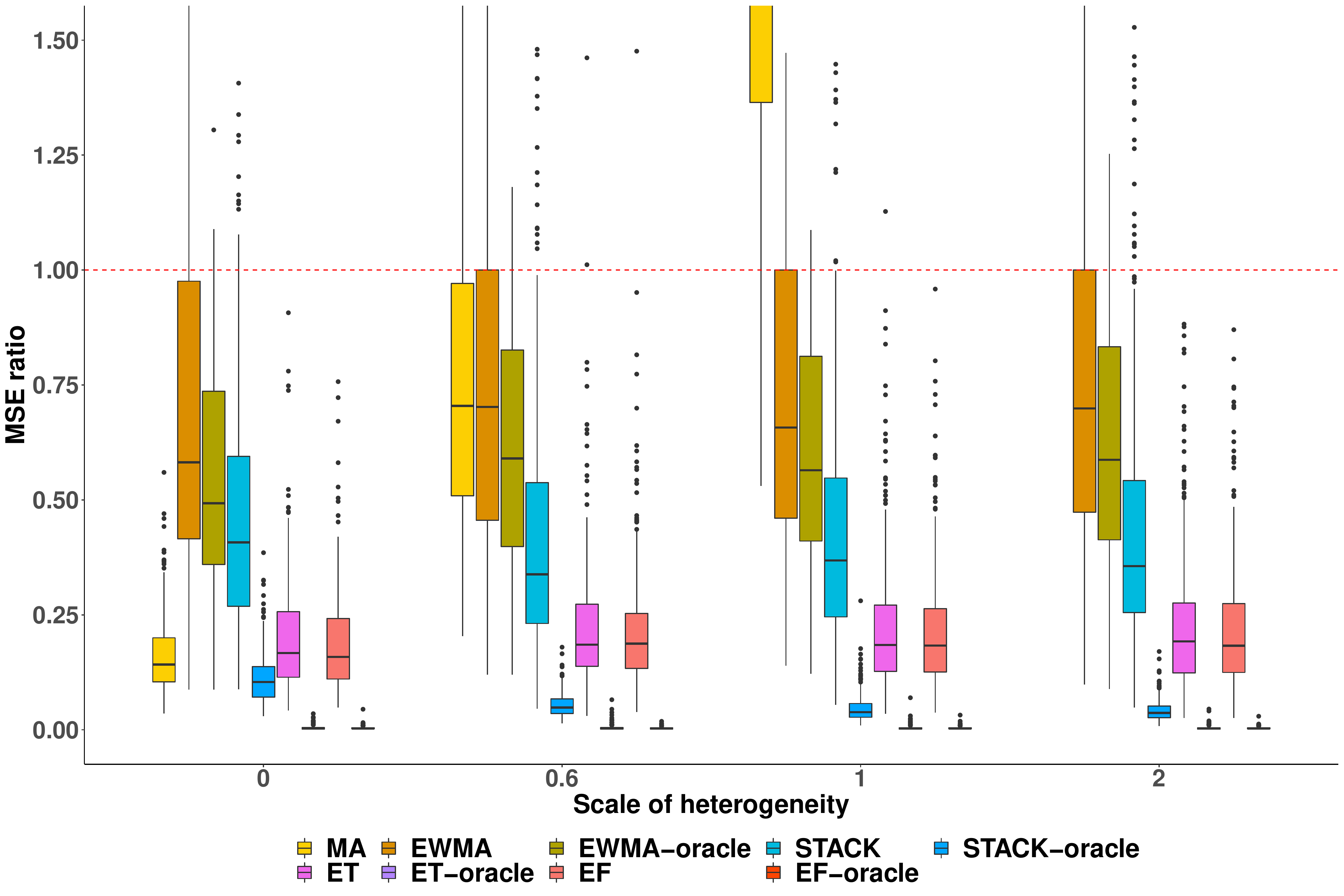}}
  \caption{}
 \end{subfigure}
 \begin{subfigure}{0.49\textwidth}
  \centerline{\includegraphics[width=\linewidth]{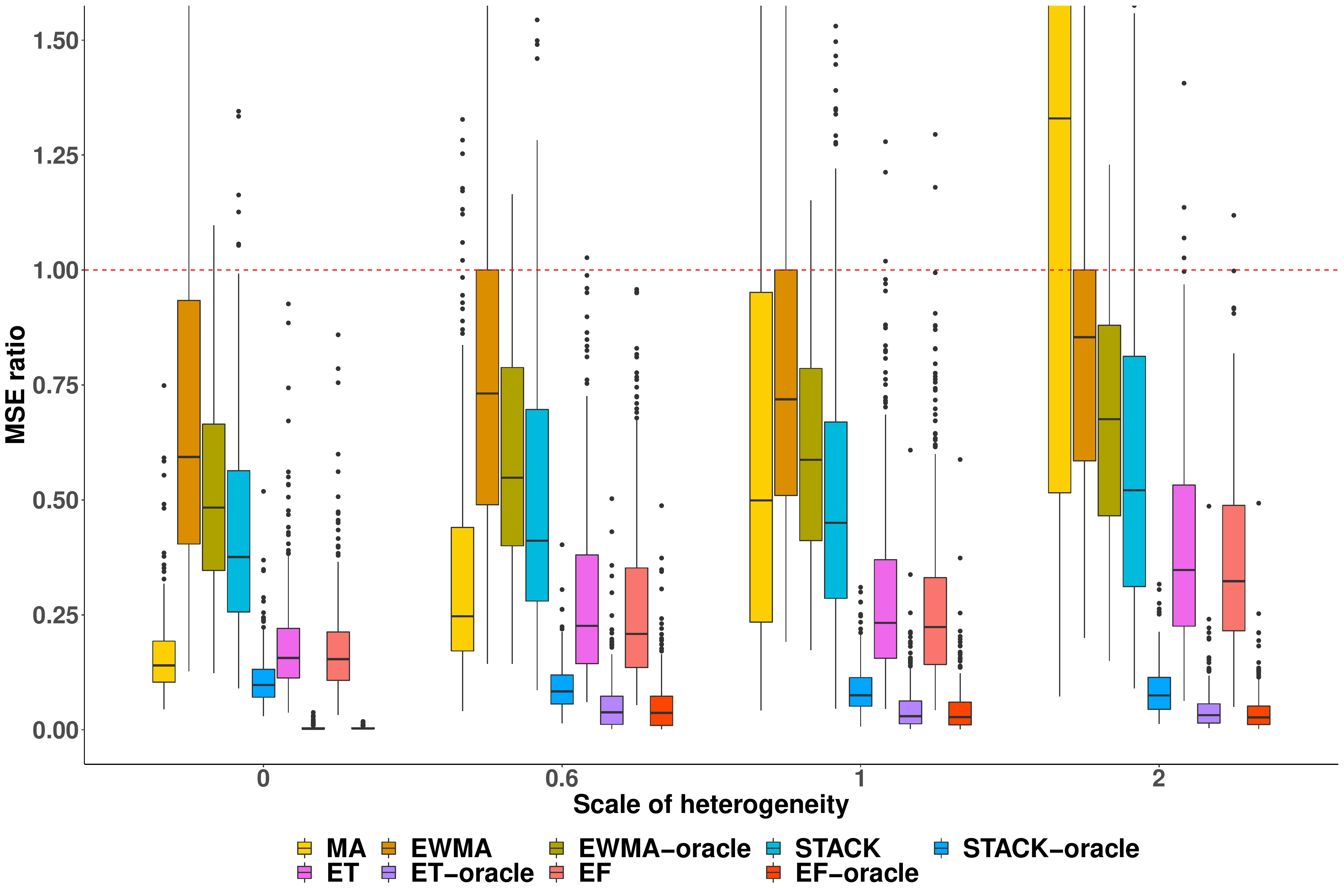}}
  \caption{}
 \end{subfigure}
 \caption{
Box plots of the MSE ratios of CATE estimators, respectively, over LOC (\textbf{CT}) and a sample size of \textbf{500} at each site, and covariate dimension of \textbf{20} for \textbf{(a) discrete grouping} and \textbf{(b) continuous grouping} across site, respectively, varying scale of global heterogeneity. 
Estimators ending with ``-oracle" makes use of ground truth treatment effects. 
Different colors imply different estimators, and x-axis, i.e., the value of $c$, differentiates the scale of global heterogeneity. The red dotted line denotes an MSE ratio of 1. 
MA performance is truncated due to large MSE ratios. 
The proposed ET and EF achieve competitive performance compared to standard model averaging or ensemble methods and are robust to heterogeneity across settings. 
Note that ET-oracle and EF-oracle achieve close-to-zero MSE ratios with very small spreads in some settings. 
 }
 \label{web:sim_p20}
\end{figure}

\begin{figure}[!h]
\centering
 \begin{subfigure}{0.49\textwidth}
  \centerline{\includegraphics[width=\linewidth]{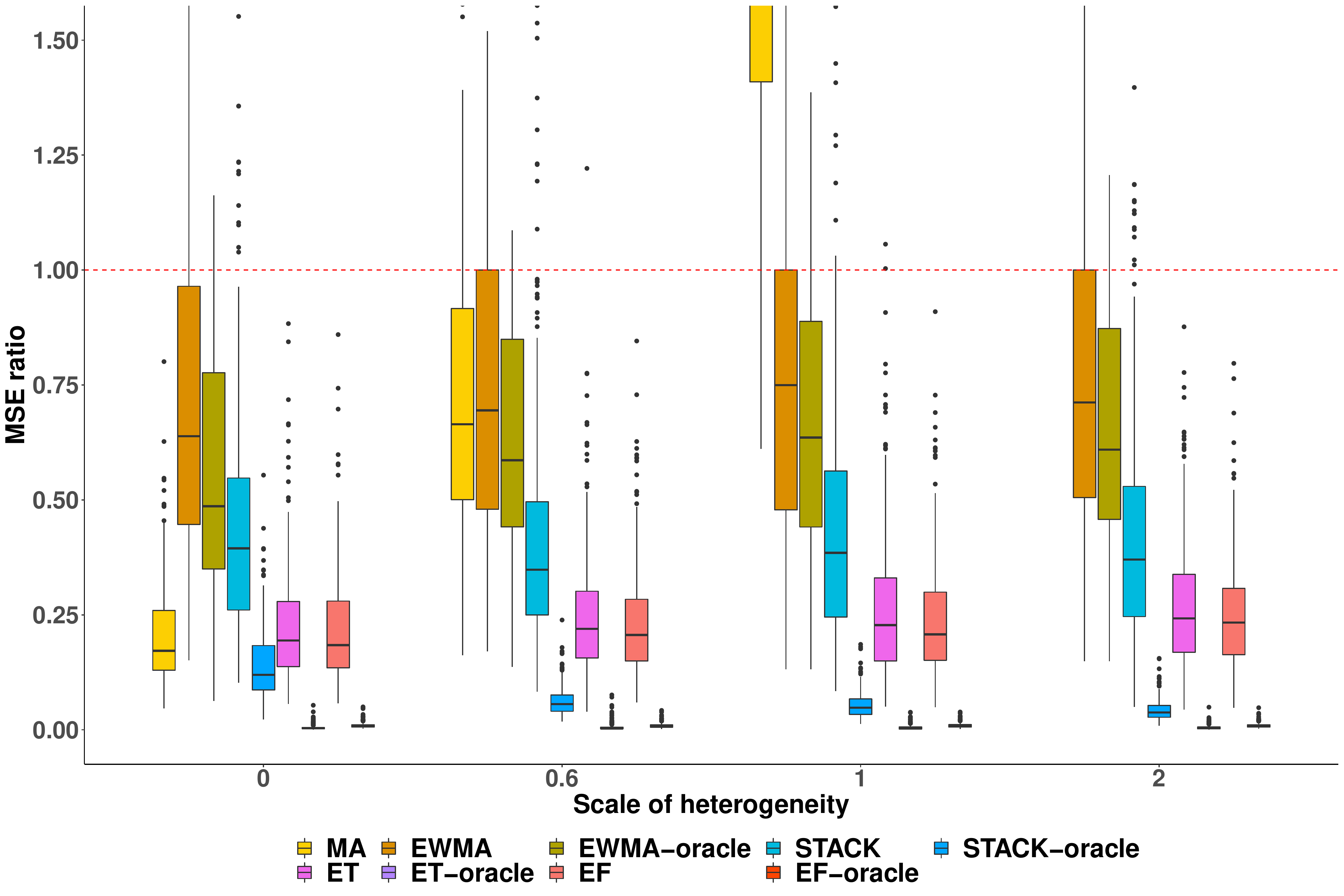}}
  \caption{}
 \end{subfigure}
 \begin{subfigure}{0.49\textwidth}
  \centerline{\includegraphics[width=\linewidth]{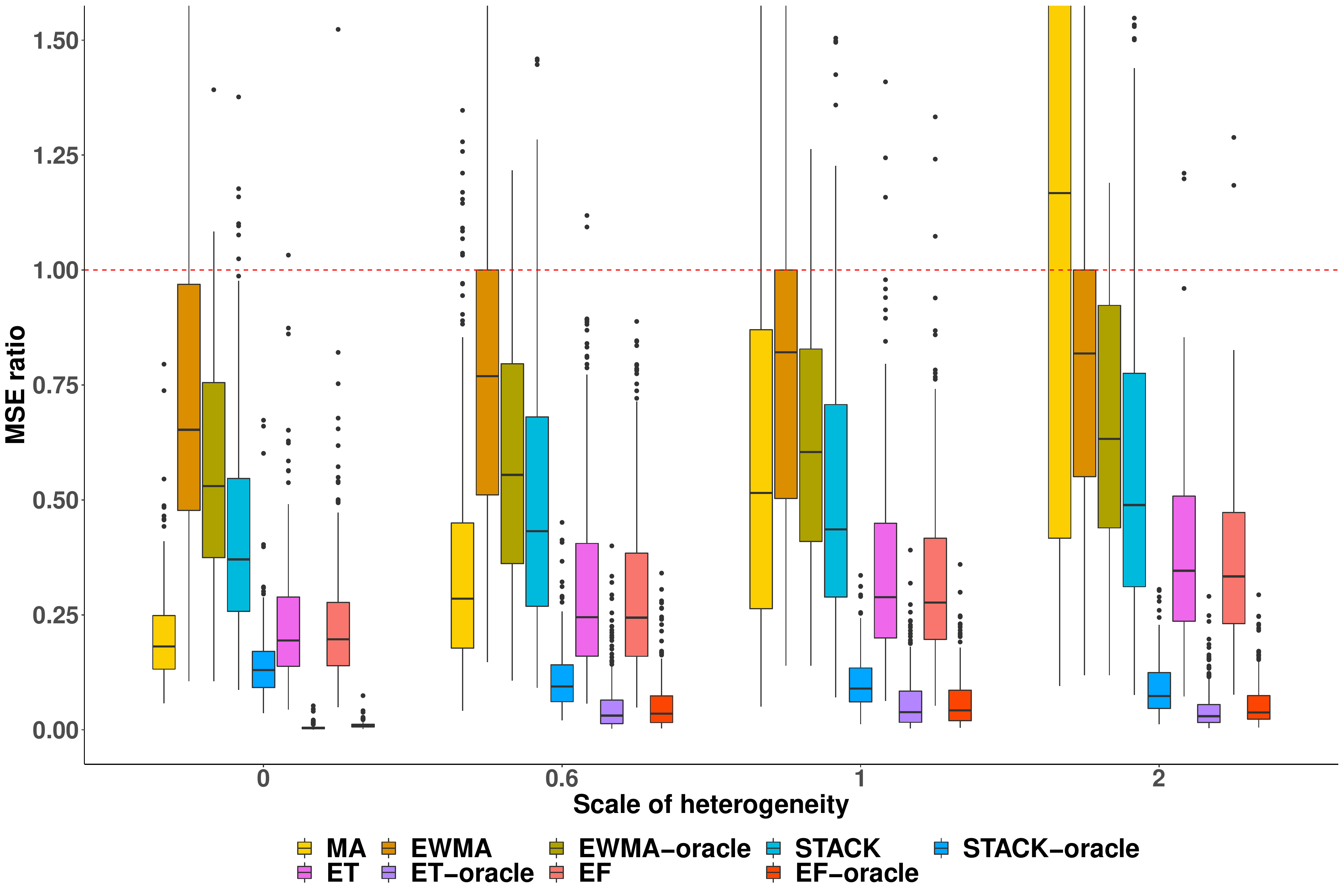}}
  \caption{}
 \end{subfigure}
 \caption{
Box plots of the MSE ratios of CATE estimators, respectively, over LOC (\textbf{CT}) and a sample size of \textbf{500} at each site, and covariate dimension of \textbf{50} for \textbf{(a) discrete grouping} and \textbf{(b) continuous grouping} across site, respectively, varying scale of global heterogeneity. 
Estimators ending with ``-oracle" makes use of ground truth treatment effects. 
Different colors imply different estimators, and x-axis, i.e., the value of $c$, differentiates the scale of global heterogeneity. The red dotted line denotes an MSE ratio of 1. 
MA performance is truncated due to large MSE ratios. 
The proposed ET and EF achieve competitive performance compared to standard model averaging or ensemble methods and are robust to heterogeneity across settings. 
Note that ET-oracle and EF-oracle achieve close-to-zero MSE ratios with very small spreads in some settings. 
 }
 \label{web:sim_p50}
\end{figure}

\textbf{Covariate dimensions}
We consider various choices of covariate dimensions besides $D=5$. Specifically, we also try $D=20$ and $D=50$. 
Figure~\ref{web:sim_p20} and Figure~\ref{web:sim_p50} show box plots of simulation results. 
With a higher dimension of variables, the MSE ratio between the proposed estimates and LOC estimates increases than that in the scenario with a small dimension.

\begin{figure}[!h]
\centering
 \begin{subfigure}{0.49\textwidth}
  \centerline{\includegraphics[width=\linewidth]{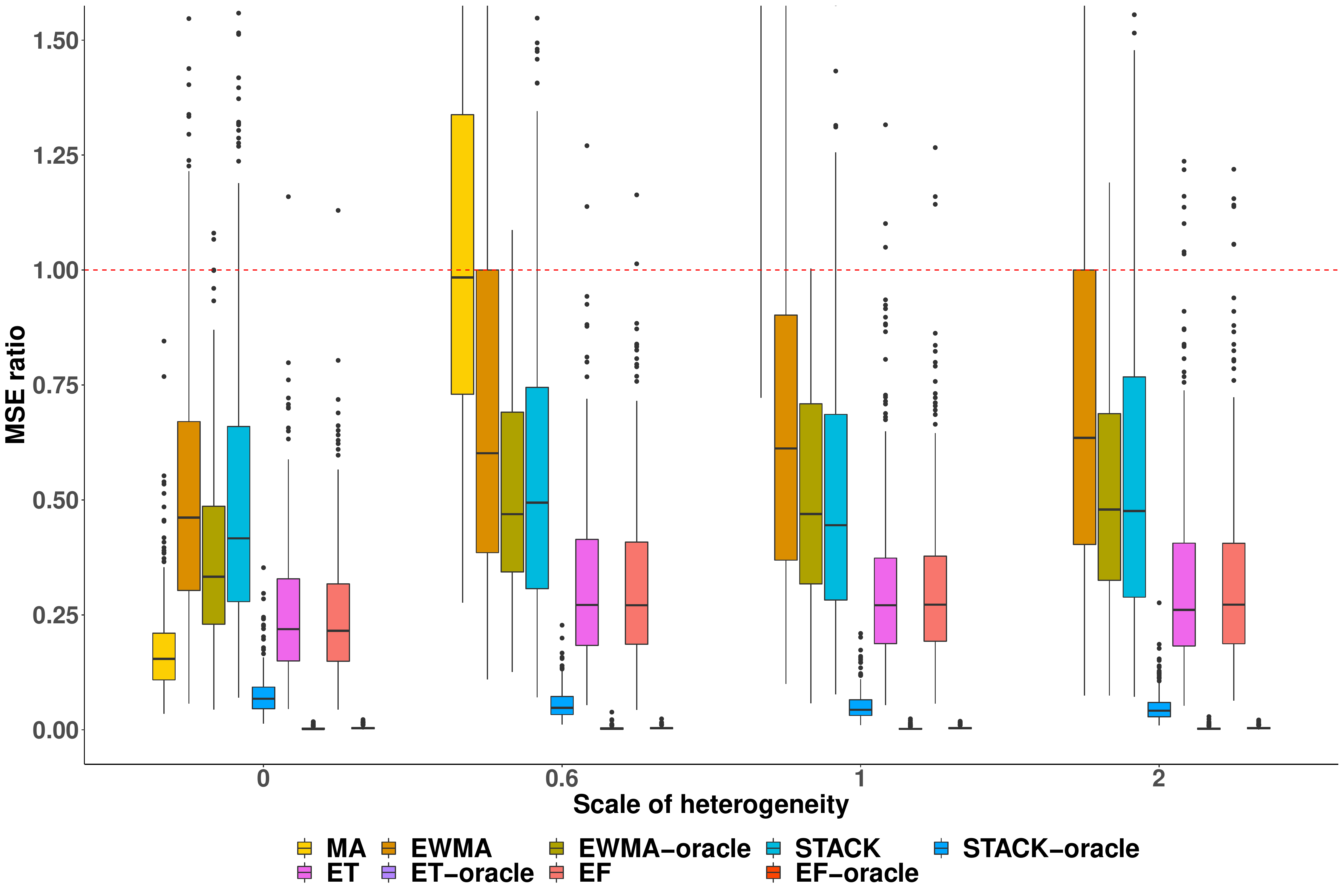}}
  \caption{}
 \end{subfigure}
 \begin{subfigure}{0.49\textwidth}
  \centerline{\includegraphics[width=\linewidth]{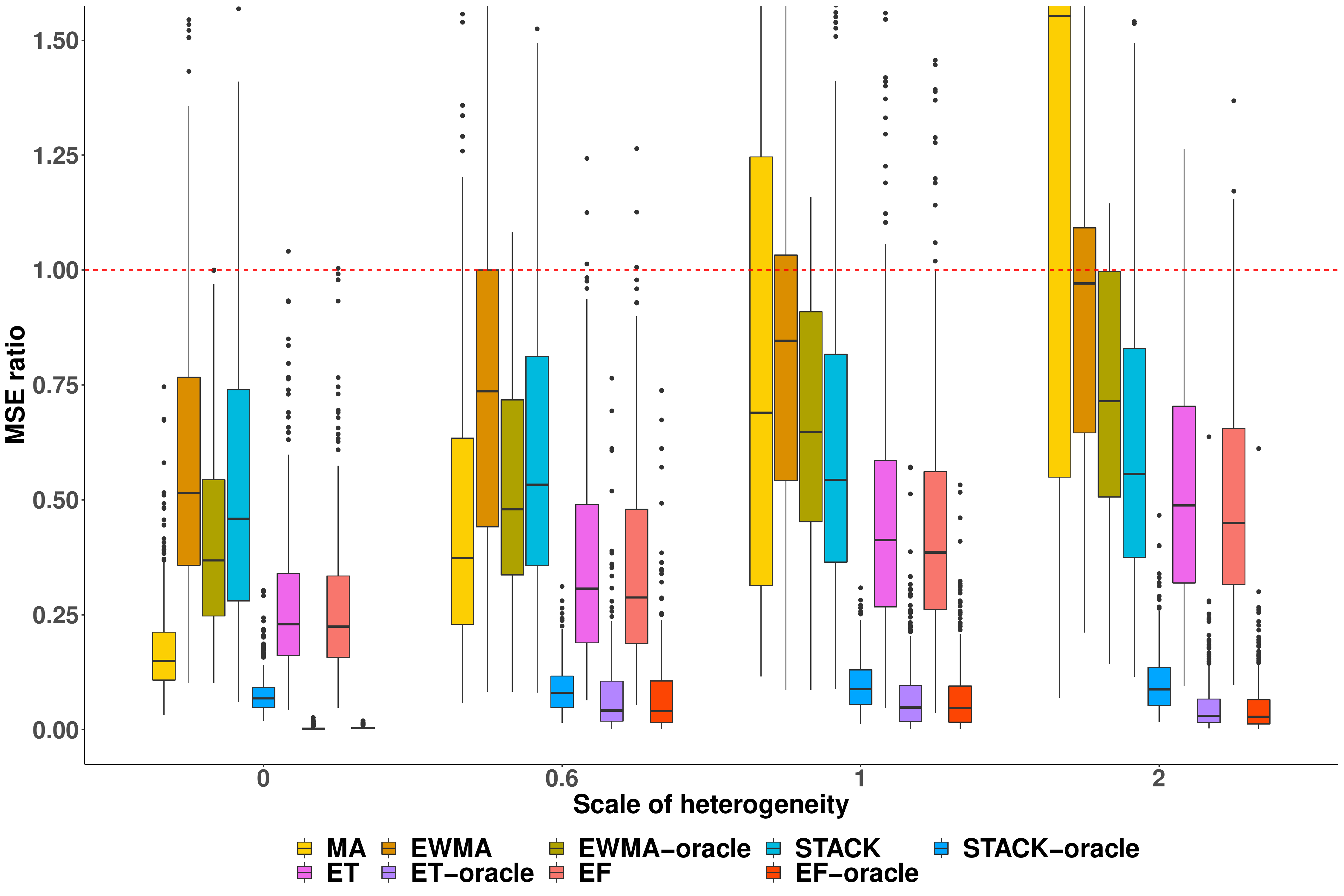}}
  \caption{}
 \end{subfigure}
 \caption{
Box plots of the MSE ratios of CATE estimators, respectively, over LOC (\textbf{CT}) and a sample size of \textbf{500} at site 1, and a sample size of \textbf{200} at other sites for \textbf{(a) discrete grouping} and \textbf{(b) continuous grouping} across site, respectively, varying scale of global heterogeneity. 
Estimators ending with ``-oracle" makes use of ground truth treatment effects. 
Different colors imply different estimators, and x-axis, i.e., the value of $c$, differentiates the scale of global heterogeneity. The red dotted line denotes an MSE ratio of 1. 
MA performance is truncated due to large MSE ratios. 
The proposed ET and EF achieve competitive performance compared to standard model averaging or ensemble methods and are robust to heterogeneity across settings. 
Note that ET-oracle and EF-oracle achieve close-to-zero MSE ratios with very small spreads in some settings. 
 }
 \label{web:sim_diffN}
\end{figure}

\textbf{Unequal sample size at each site.}
In the distributed date network, different sites may have a different sample size $n_k$. Those with a smaller sample size may not be representative of their population, leading to an uneven level of precision for local causal estimates. We consider a simulation setting where site 1 has a sample size of $n_1=500$ while other site $n_2,\ldots,n_K$ has a sample size of 200. 
Figure~\ref{web:sim_diffN} shows box plots of simulation results. 
Results show that the MSE ratio between the proposed estimates and LOC estimates increases compared to the scenario where the sample size in all sites are 500. However, the proposed estimators still enjoy the most robust performance via bias-and-variance trade-off. 
This also shows our method is robust to the existence of local uncertainty.

\begin{figure}[h]
\centering
 \begin{subfigure}{0.49\textwidth}
  \centerline{\includegraphics[width=\linewidth]{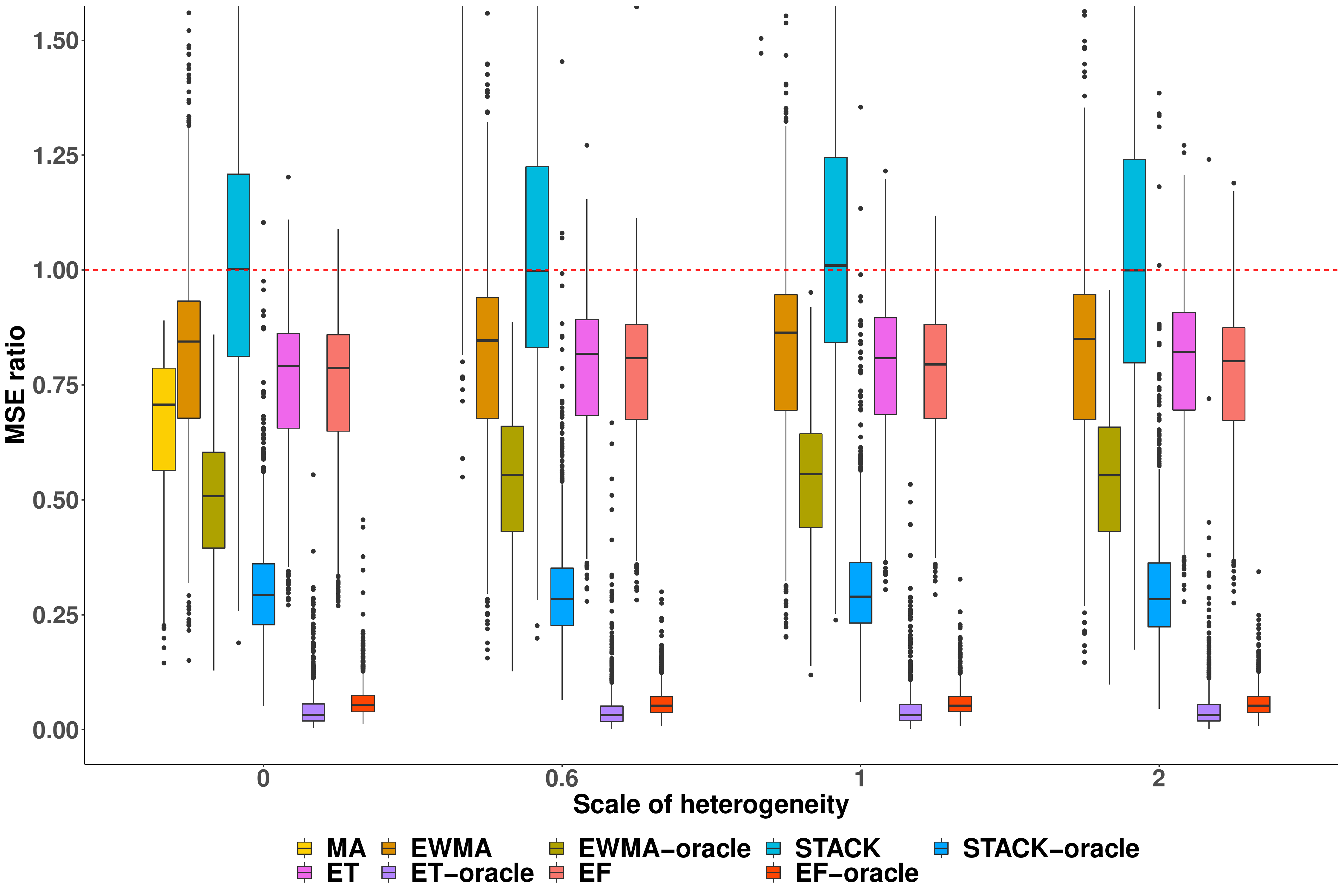}}
  \caption{}
 \end{subfigure}
 \begin{subfigure}{0.49\textwidth}
  \centerline{\includegraphics[width=\linewidth]{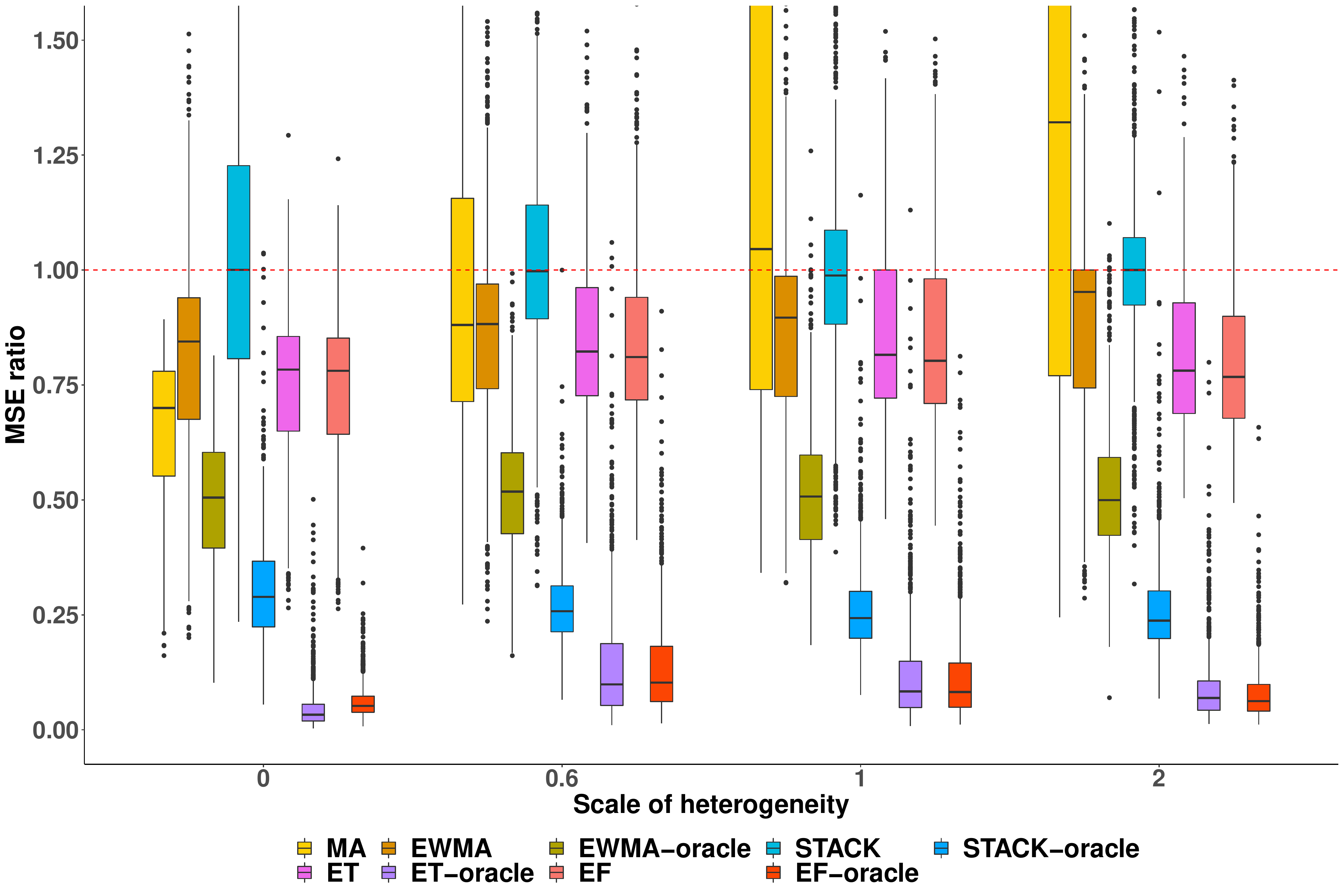}}
  \caption{}
 \end{subfigure}
 \caption{
Box plots of the MSE ratios of CATE estimators, respectively, over LOC (\textbf{CF}) and a sample size of \textbf{100} at each site for \textbf{(a) discrete grouping} and \textbf{(b) continuous grouping} across site, respectively, varying scale of global heterogeneity. 
Estimators ending with ``-oracle" makes use of ground truth treatment effects. 
Different colors imply different estimators, and x-axis, i.e., the value of $c$, differentiates the scale of global heterogeneity. The red dotted line denotes an MSE ratio of 1. 
MA performance is truncated due to large MSE ratios. 
The proposed ET and EF achieve competitive performance compared to standard model averaging or ensemble methods and are robust to heterogeneity across settings. 
Note that ET-oracle and EF-oracle achieve close-to-zero MSE ratios with very small spreads in some settings. 
 }
 \label{web:sim_fig_cf100}
\end{figure}

\begin{figure}[!h]
\centering
 \begin{subfigure}{0.49\textwidth}
  \centerline{\includegraphics[width=\linewidth]{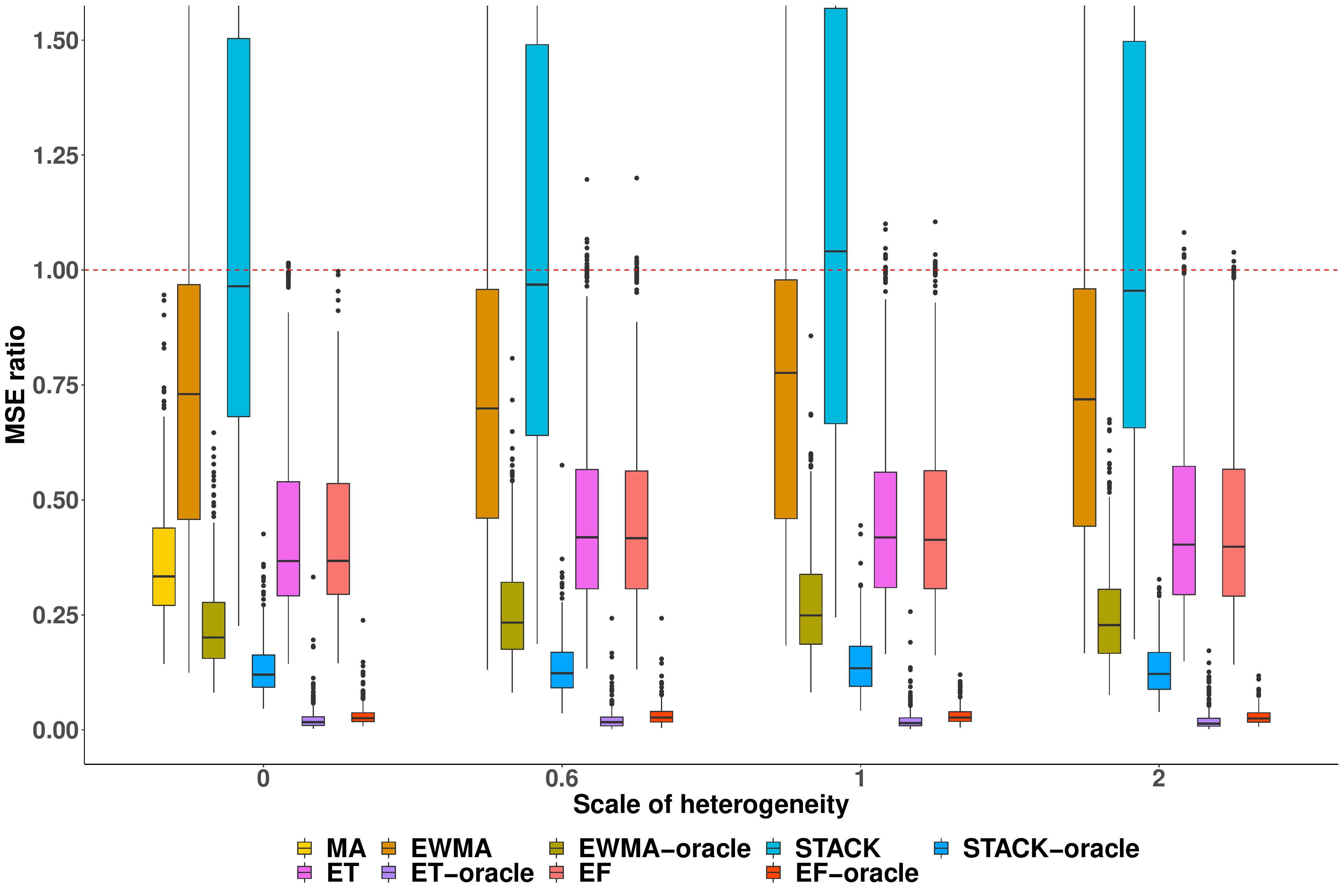}}
  \caption{}
 \end{subfigure}
 \begin{subfigure}{0.49\textwidth}
  \centerline{\includegraphics[width=\linewidth]{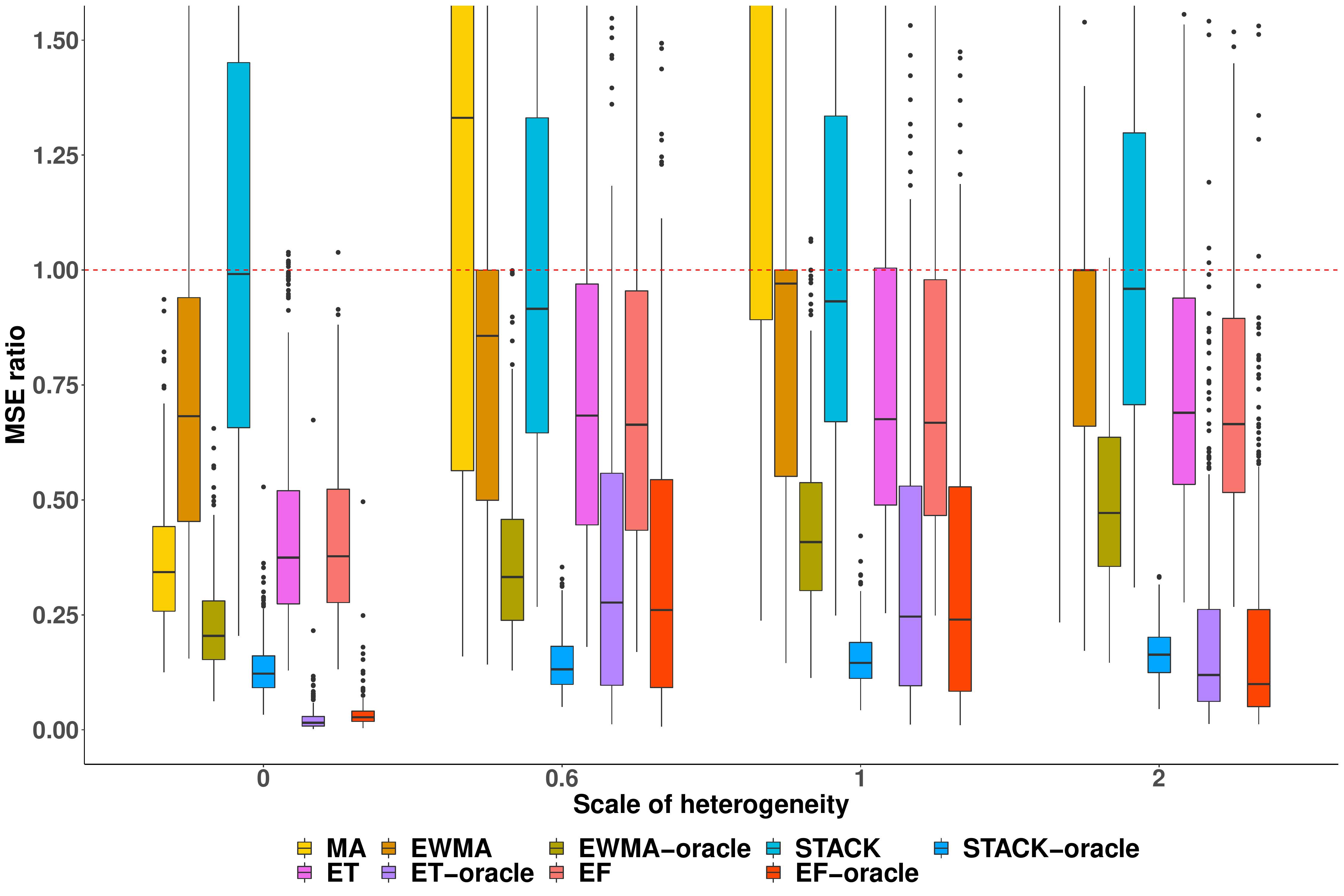}}
  \caption{}
 \end{subfigure}
 \caption{
Box plots of the MSE ratios of CATE estimators, respectively, over LOC (\textbf{CF}) and a sample size of \textbf{500} at each site for \textbf{(a) discrete grouping} and \textbf{(b) continuous grouping} across site, respectively, varying scale of global heterogeneity. 
Estimators ending with ``-oracle" makes use of ground truth treatment effects. 
Different colors imply different estimators, and x-axis, i.e., the value of $c$, differentiates the scale of global heterogeneity. The red dotted line denotes an MSE ratio of 1. 
MA performance is truncated due to large MSE ratios. 
The proposed ET and EF achieve competitive performance compared to standard model averaging or ensemble methods and are robust to heterogeneity across settings. 
Note that ET-oracle and EF-oracle achieve close-to-zero MSE ratios with very small spreads in some settings. 
 }
 \label{web:sim_fig_cf500}
\end{figure}

\begin{figure}[!h]
\centering
 \begin{subfigure}{0.49\textwidth}
  \centerline{\includegraphics[width=\linewidth]{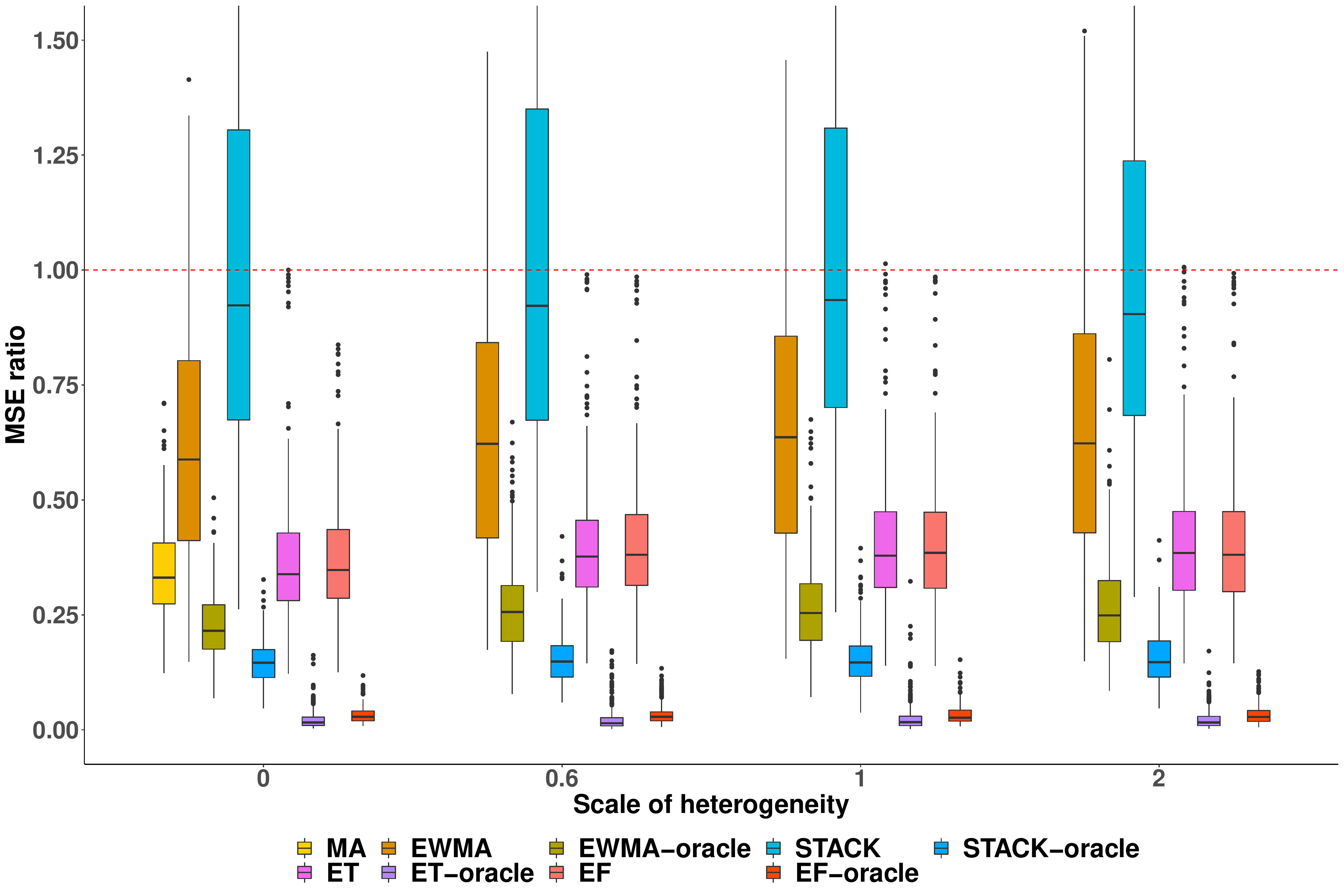}}
  \caption{}
 \end{subfigure}
 \begin{subfigure}{0.49\textwidth}
  \centerline{\includegraphics[width=\linewidth]{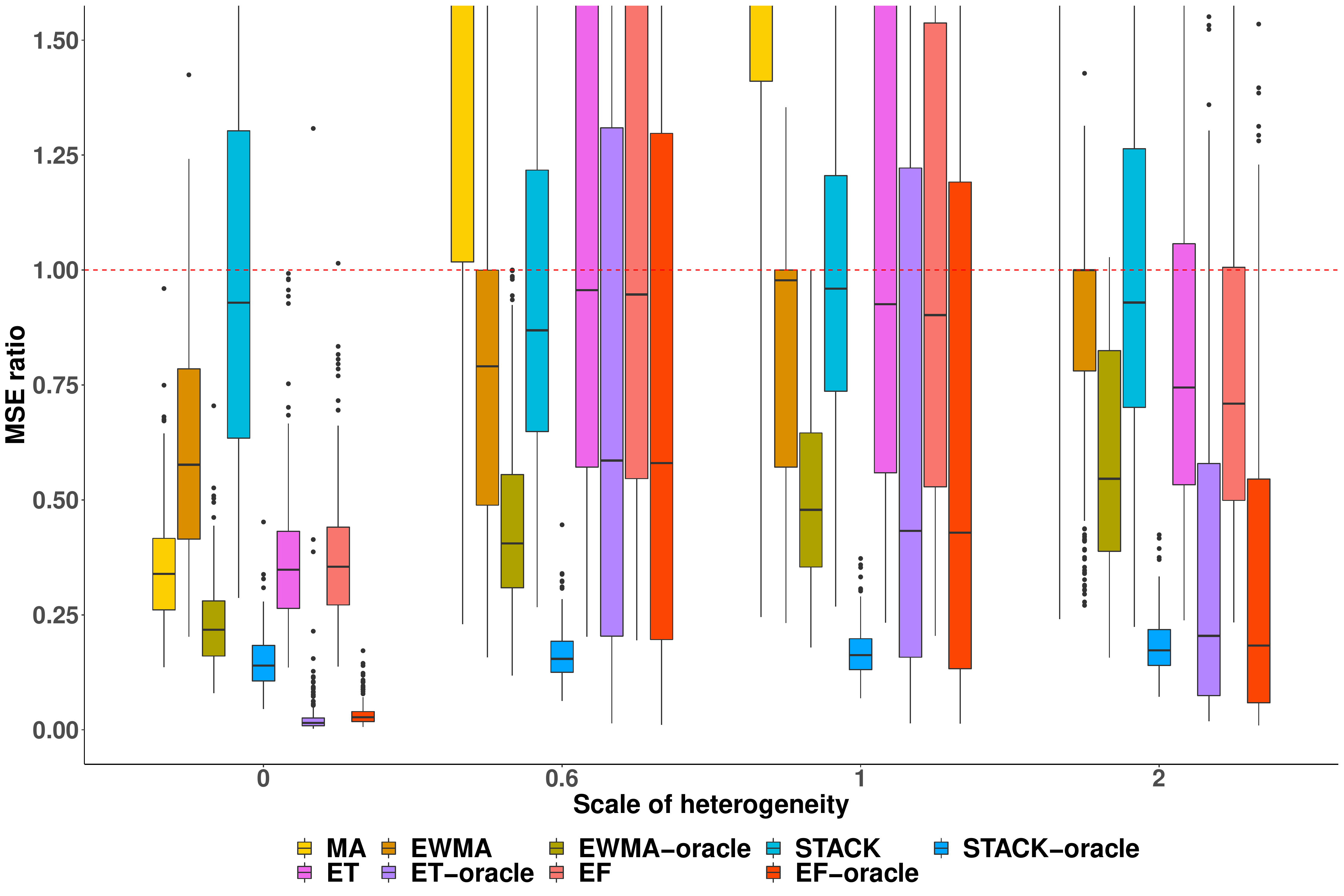}}
  \caption{}
 \end{subfigure}
 \caption{
Box plots of the MSE ratios of CATE estimators, respectively, over LOC (\textbf{CF}) and a sample size of \textbf{1000} at each site for \textbf{(a) discrete grouping} and \textbf{(b) continuous grouping} across site, respectively, varying scale of global heterogeneity. 
Estimators ending with ``-oracle" makes use of ground truth treatment effects. 
Different colors imply different estimators, and x-axis, i.e., the value of $c$, differentiates the scale of global heterogeneity. The red dotted line denotes an MSE ratio of 1. 
MA performance is truncated due to large MSE ratios. 
The proposed ET and EF achieve competitive performance compared to standard model averaging or ensemble methods and are robust to heterogeneity across settings. 
Note that ET-oracle and EF-oracle achieve close-to-zero MSE ratios with very small spreads in some settings. 
 }
 \label{web:sim_fig_cf1000}
\end{figure}

\textbf{Different local estimators.} 
We explore another option for the local model using the causal forest (CF) \citep{wager2018estimation} varying the sample size at local sites. 
A causal forest is a stochastic averaging of multiple causal trees \citep{athey2016recursive}, and hence is more powerful in estimating treatment effects. In each tree of the causal forest, MSE of treatment effect is used to select the feature and cutoff point in each split \citep{wager2018estimation}. 
CF is implemented in the R packages \verb|grf|.
Figure~\ref{web:sim_fig_cf100}, Figure~\ref{web:sim_fig_cf500}, and Figure~\ref{web:sim_fig_cf1000} show box plots of simulation results with a sample size of 100, 500, and 1000, respectively, at each site. Our proposed methods ET and EF show robust performance in all settings regardless of the use of information of the ground truth $\tau_1(\bx)$.

\textbf{Further comparisons to non-adaptive ensemble.} 
We provide simulation results to compare the proposed methods to the non-adaptive method STACK. 
Consider the following setting where the heterogeneity is continuous and nonlinear: $\tau(\bx, k) = \mathbbm{1}\{x_1 > 0\} \cdot x_1 +(x_1 - 3) \cdot {(U_k)}^c,$
with $U_k\sim Unif[0,3]$, $\bX_i\sim {N}(\bm{0},\bm{I}_5)$, and $c=(1,2,3,4)$. 
As $c$ increases, the heterogeneity across sites gets larger, reducing the influence of $x_1$ on heterogeneity, hence the weights become more non-adaptive. 
For $c=(1,2,3,4)$, 
the one-SD ranges of MSE ratios of EF over STACK are [0.73,0.82], [0.86,0.87], [0.99,1.04], [0.87,1.07], respectively. 
When $c$ is relatively small, the proposed EF has a smaller MSE compared to STACK. As $c$ increases, the performance of EF is similar to that of STACK, in the case of a large global heterogeneity. This further indicates the robustness of the proposed methods.

\section{Additional Results for Data Application}
\label{suppl-sec:real}



In real-life applications, hospitals may have different sample sizes $n_k$ that may affect the accuracy of the estimation of $\tau_k$. Table \ref{web:hosp_smry} shows hospital-level information for the 20 hospitals where the number of patients across sites varies. Information includes the region of the U.S. where the hospital is located, whether it is a teaching hospital, the bed capacity, and the number of patients within the hospital.

Hospitals with a smaller sample size may not be representative of the population, leading to an uneven level of precision for local causal estimates. To account for different sample sizes at each hospital, we consider a basic weighting strategy where we add weights to each observation $\widehat \tau_k(\bx)$ in the augmented site 1 data adjusting for the sample size of site $k$. The weights are defined as
$
    \eta_k(\bx) = K n_k \{\sum_{j=1}^K n_j\}^{-1}.
$

\begin{figure*}[htp]
\centering
 \begin{subfigure}{0.2\textwidth}
  \centerline{\includegraphics[width=\linewidth]{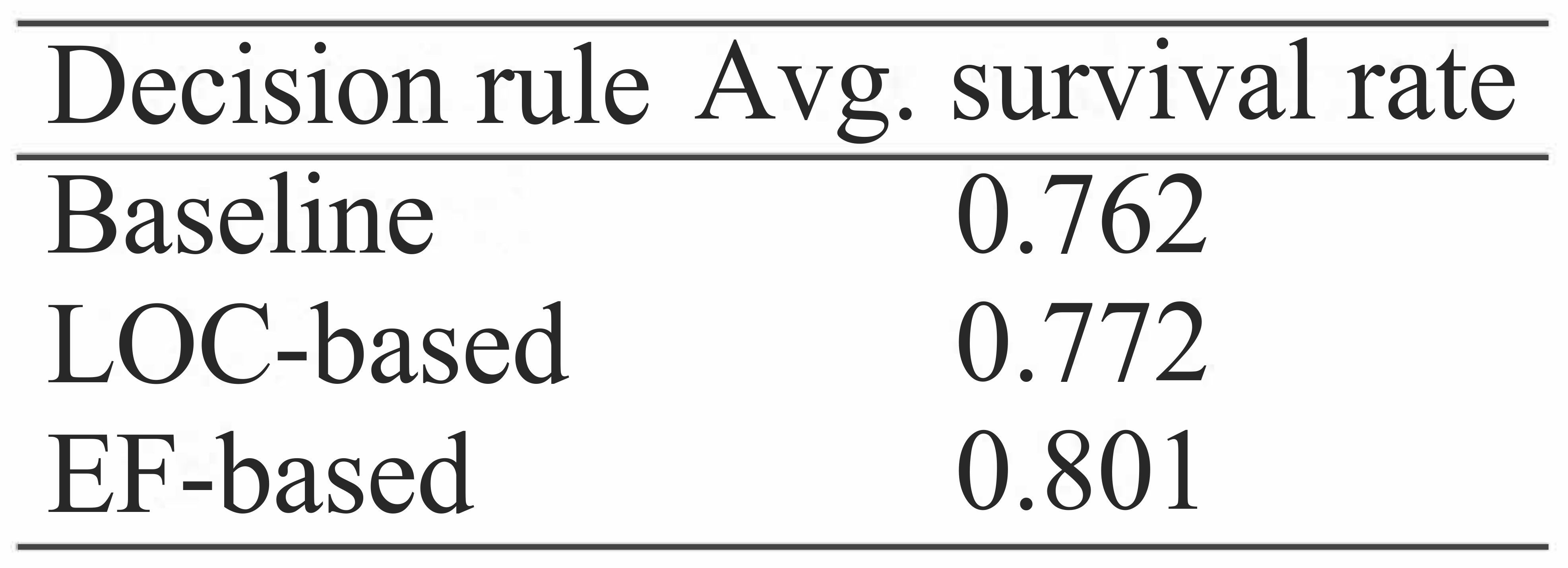}}
  \caption{}
 \end{subfigure}
 \begin{subfigure}{0.39\textwidth}
  \centerline{\includegraphics[width=\linewidth]{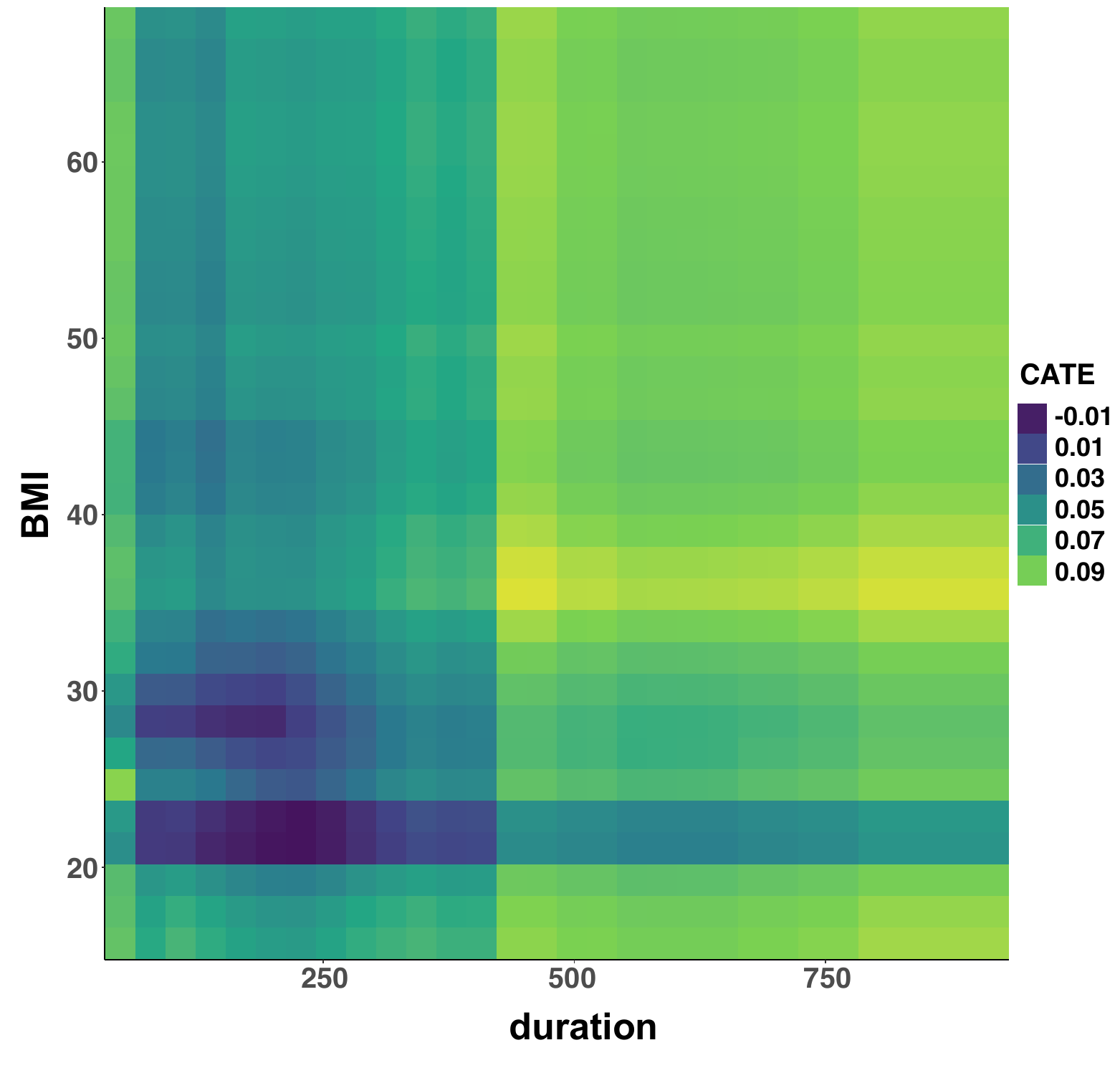}}
  \caption{}
 \end{subfigure}
 \begin{subfigure}{0.39\textwidth}
  \centerline{\includegraphics[width=\linewidth]{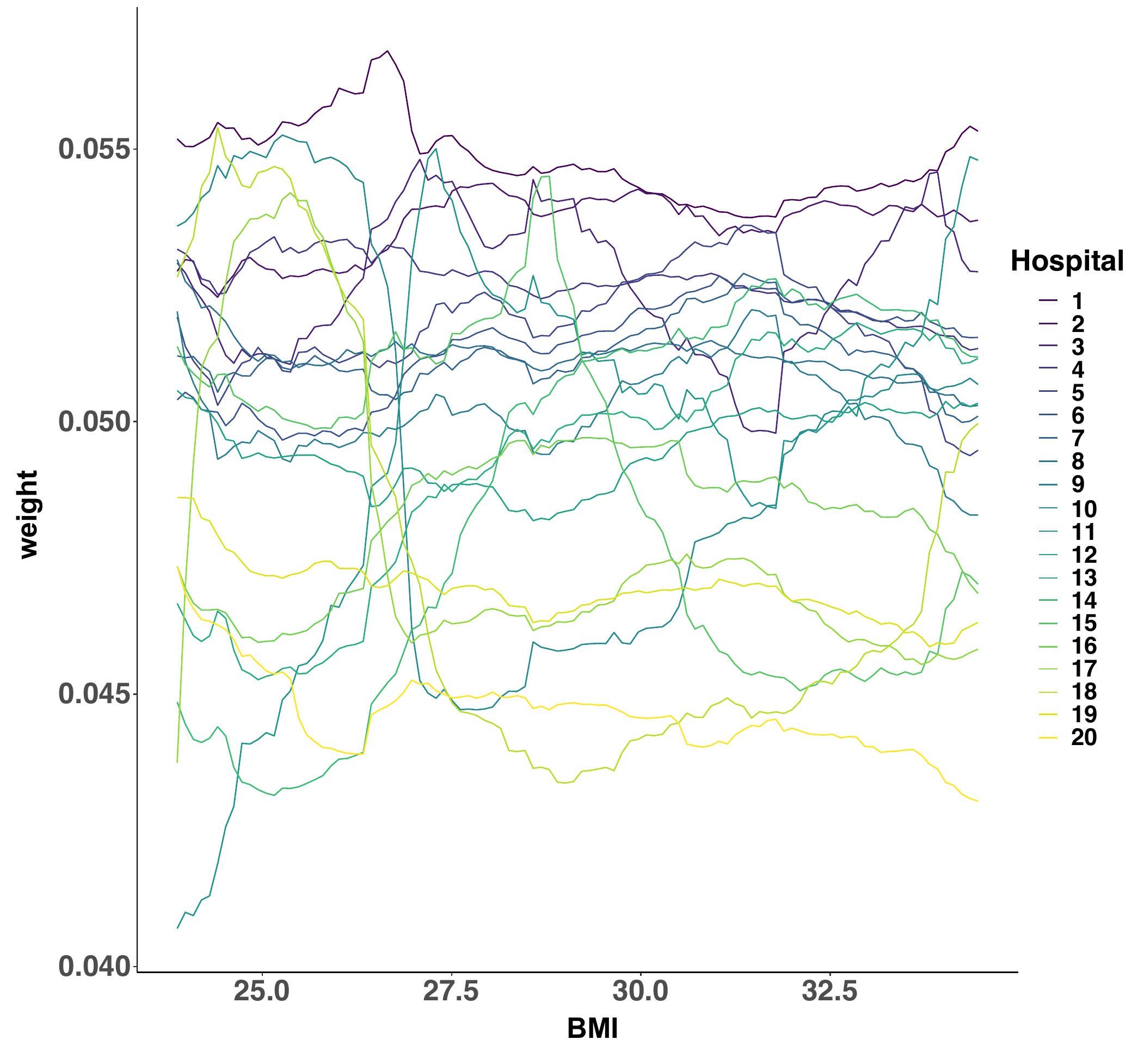}}
  \caption{}
 \end{subfigure}
 \vspace{-0.2cm}
 \caption{
 Application to estimating treatment effects of oxygen therapy on survival with a \textbf{sample size weighting strategy}. 
 (a) Expected survival of treatment decision following different estimators. 
 EF shows the largest gain in improving survival rate, more promising than the LOC and the baseline. 
 (b) Estimated treatment effects varying duration and BMI, two important features in the fitted EF. 
 Patients with a BMI around 35, and a duration above 400 benefited the most. 
 (c) Visualization of data-adaptive weights in EF varying BMI. 
 Hospitals with a larger bed capacity tend to contribute more, the data of which might be more similar to hospital 1.
}
\vspace{-0.3cm}
\label{web:real_wt}
\end{figure*}

Figure~\ref{web:real_wt} visualizes the performance of oxygen therapy on hospital survival with the weighting strategy adopted. 
CT is used as the local model with propensity score modeled by a logistic regression. 
Figure~\ref{web:real_wt}(a) shows the propensity score-weighted average survival for those whose received treatment is consistent with the estimated decision. 
Treatment rule based on our method can increase survival by 4\%, more promising than the EF estimates without the weighting strategy and the LOC and the baseline. The weighting strategy takes account into the unequal sample size among the hospital network, and assign weights based on precision of local estimates. 

In the fitted EF, 
the hospital indicator remains the most important, explaining about 48\% of the decrease in training error. 
Figure~\ref{web:real_wt}(b) shows the estimated CATEs varying two important features, BMI and oxygen therapy duration. 
Patients with BMI between 36 and 40 and duration above 400 show the most benefit from oxygen therapy in the target SpO$_2$ range. 
Patients with BMI between 20 and 30 and duration between 100 and 400 may not benefit from such alteration. 
The treatment estimates are similar to that in Figure~\ref{fig:real}(b)
Figure~\ref{web:real_wt}(c) visualizes 
the proposed model averaging scheme with
data-adaptive weights $\omega_{k}(\bx)$ in the fitted EF with respect to BMI for different models, while holding other variables constant. The weights of hospital 1 are quite stable while models from other sites may have different contribution to the weighted estimator for different values of BMI. 
Similar to Figure~\ref{fig:real}(c), hospitals with a larger bed capacity tend to be similar to hospital 1, and are shown to provide larger contributions. 
In general, the weighting strategy helps further improve the expected survival rate. The patterns in each subfigure are similar to Figure~\ref{fig:real}, which indicates the robustness of our proposed estimators.
We do stress that improvements to the weighting strategy for different sample sizes at each site are needed. 
A strategy considering both treatment proportion as well as covariate distributions across sites may further enhance the data-adaptive model averaging estimator.

\section{Real Data Access}
\label{suppl-sec:code}
Although the eICU data used in our application example cannot be shared subject to the data use agreement, access can be individually requested at \url{https://eicu-crd.mit.edu/gettingstarted/access/}. 

\newpage

\begin{table}[!h]
\centering
\caption{Simulation results for ratio between MSE of the estimator and MSE of LOC  (\textbf{CT}) with a sample size of \textbf{500} at each site. A smaller number indicates larger improvement over the local model. 
Estimators ending with ``-oracle" makes use of ground truth treatment effects.
Our proposed methods ET and EF shows robust performance in all settings whether or not using the information of ground truth $\tau_1(\bx)$.}
\label{tab:sim_res}
\begin{tabular}{@{}ccccccccc@{}}
\toprule
 & \multicolumn{4}{c}{Discrete grouping} & \multicolumn{4}{c}{Continuous grouping} \\ \cmidrule(l){2-9} 
Estimator & $c=0$ & $c=0.2$ & $c=0.6$ & $c=1$ & $c=0$ & $c=0.2$ & $c=0.6$ & $c=1$ \\ \midrule
\multicolumn{9}{l}{\textit{Ratio of average of MSEs over 1000 replicates}} \\ 
MA & 0.09 & 0.91 & 2.4 & 9.87 & 0.08 & 0.32 & 0.65 & 1.78 \\
EWMA & 0.57 & 0.62 & 0.61 & 0.62 & 0.56 & 0.65 & 0.7 & 0.77 \\
EWMA-oracle & 0.42 & 0.5 & 0.49 & 0.5 & 0.42 & 0.49 & 0.53 & 0.59 \\
STACK & 0.44 & 0.45 & 0.44 & 0.45 & 0.45 & 0.45 & 0.48 & 0.54 \\
STACK-oracle & 0.06 & 0.04 & 0.04 & 0.04 & 0.06 & 0.06 & 0.06 & 0.07 \\
ET & 0.12 & 0.17 & 0.16 & 0.16 & 0.13 & 0.24 & 0.29 & 0.37 \\
ET-oracle & $<$0.01 & $<$0.01 & $<$0.01 & $<$0.01 & $<$0.01 & 0.08 & 0.1 & 0.07 \\
EF & 0.1 & 0.13 & 0.13 & 0.13 & 0.1 & 0.19 & 0.25 & 0.3 \\
EF-oracle & $<$0.01 & $<$0.01 & $<$0.01 & $<$0.01 & $<$0.01 & 0.06 & 0.06 & 0.05 \\
\midrule
\multicolumn{9}{l}{\textit{Ratio of standard deviation of MSEs over 1000 replicates}} \\
MA & 0.15 & 0.35 & 0.76 & 3.05 & 0.14 & 0.24 & 0.38 & 0.81 \\
EWMA & 0.61 & 0.65 & 0.67 & 0.66 & 0.58 & 0.65 & 0.69 & 0.75 \\
EWMA-oracle & 0.46 & 0.52 & 0.54 & 0.54 & 0.44 & 0.52 & 0.55 & 0.6 \\
STACK & 0.47 & 0.46 & 0.47 & 0.47 & 0.45 & 0.49 & 0.52 & 0.6 \\
STACK-oracle & 0.1 & 0.08 & 0.08 & 0.08 & 0.09 & 0.11 & 0.12 & 0.14 \\
ET & 0.18 & 0.23 & 0.22 & 0.22 & 0.18 & 0.26 & 0.32 & 0.43 \\
ET-oracle & 0.02 & 0.03 & 0.02 & 0.02 & 0.02 & 0.06 & 0.07 & 0.07 \\
EF & 0.17 & 0.19 & 0.19 & 0.2 & 0.17 & 0.23 & 0.29 & 0.39 \\
EF-oracle & 0.03 & 0.03 & 0.03 & 0.03 & 0.03 & 0.06 & 0.07 & 0.08\\ \bottomrule
\end{tabular}
\vspace{1cm}
\end{table}

\begin{table}[h]
\centering
\caption{Hospital-level information of our analysis cohort in eICU database. Hospitals are relabeled according to their average contribution to the estimation task at hospital 1, the target site.}
\label{web:hosp_smry}
\begin{tabular}{@{}rrrrlll@{}}
\toprule
\multicolumn{1}{l}{Hospital} &
  \multicolumn{1}{l}{Number of} &
  \multicolumn{1}{l}{Number of} &
  \multicolumn{1}{l}{Number of} &
  Bed &
  Teaching &
  \multirow{2}{*}{Region} \\
\multicolumn{1}{l}{site} &
  \multicolumn{1}{l}{patients} &
  \multicolumn{1}{l}{control} &
  \multicolumn{1}{l}{treated} &
  capacity &
  status &
   \\ \midrule
1  & 477 & 205 & 272 & $\geq$ 500 & False & South     \\
2  & 297 & 109 & 188 & $\geq$ 500 & True  & West      \\
3 & 163 & 58  & 105 & $\geq$ 500 & True  & Midwest   \\
4 & 222 & 58  & 164 & $\geq$ 500 & False & South     \\ 
5 & 659 & 165 & 494 & $\geq$ 500 & True  & Midwest   \\
6  & 305 & 174 & 131 & $\geq$ 500 & False & South     \\
7 & 347 & 109 & 238 & $\geq$ 500 & True  & Midwest   \\
8  & 523 & 162 & 361 & $\geq$ 500 & False & South     \\
9  & 210 & 78  & 132 & Unknown             & False & Unknown   \\
10 & 379 & 161 & 218 & $\geq$ 500 & True  & Midwest   \\
11 & 234 & 70  & 164 & $\geq$ 500 & True  & Midwest   \\
12 & 747 & 185 & 562 & $\geq$ 500 & True  & Northeast \\
13  & 464 & 129 & 335 & $\geq$ 500 & True  & South     \\
14 & 474 & 229 & 245 & $\geq$ 500 & False & South     \\
15 & 166 & 64  & 102 & 100 - 249           & False & Midwest   \\
16  & 388 & 94  & 294 & $\geq$ 500 & False & Midwest   \\
17 & 435 & 240 & 195 & $\geq$ 500 & True  & South     \\
18 & 200 & 55  & 145 & 250 - 499           & False & South     \\
19  & 183 & 52  & 131 & 250 - 499           & False & West      \\
20  & 149 & 71  & 78  & 250 - 499           & False & South     \\
\bottomrule
\end{tabular}
\end{table}



\end{document}